\documentclass[arxiv,article,pdftex,moreauthors]{Definitions/mdpi} 
\firstpage{1} 
\pubvolume{1}
\issuenum{1}
\articlenumber{0}
\pubyear{2024}
\copyrightyear{2024}
\datereceived{ } 
\daterevised{ } 
\dateaccepted{ } 
\datepublished{ } 
\hreflink{https://doi.org/} 

\usepackage{algorithm}
\usepackage{algpseudocode}
\usepackage{amsmath}
\usepackage{amsthm}  
 \usepackage{enumitem} 

\usepackage{graphicx} 
\usepackage[justification=centering]{caption}
\usepackage{array}
\usepackage{booktabs}
\usepackage{caption}
\usepackage{ragged2e}
\usepackage{calc}
\newlength\oversetwidth
\newlength\underwidth

\usepackage{comment}
\makeatletter
\let\c@lemma\relax
\let\c@example\relax
\makeatother

\Title{{Selective} 
 {Reviews} 
 of Bandit Problems in AI via a Statistical~View}

\TitleCitation{Selective Reviews of Bandit Problems in AI via a Statistical View}

\Author{{Pengjie Zhou} 
 $^{1,\dagger}$, Haoyu Wei $^{2}$ and Huiming Zhang $^{1,}$*}

\AuthorNames{Pengjie Zhou, Haoyu Wei and Huiming Zhang}

\AuthorCitation{{Zhou, P.}
; Wei, H.; Zhang, H.}

\address{%
$^{1}$ \quad Institute of Artificial Intelligence, Beihang University, Beijing {100191}
, China; sy2442115@buaa.edu.cn\\
$^{2}$ \quad Department of Economics, {University of California San Diego}
, {La Jolla, CA 92093, USA}
; h8wei@ucsd.edu}

\corres{Correspondence: zhanghuiming@buaa.edu.cn}

\abstract{
Reinforcement Learning (RL) is a widely researched area in artificial intelligence that focuses on teaching agents decision-making through interactions with their environment. A key subset includes stochastic multi-armed bandit (MAB) and continuum-armed bandit (SCAB) problems, which model sequential decision-making under uncertainty. This review outlines the foundational models and assumptions of bandit problems, explores non-asymptotic theoretical tools like concentration inequalities and minimax regret bounds, and compares frequentist and Bayesian algorithms for managing exploration–exploitation trade-offs. 
Additionally, we explore $K$-armed contextual bandits and SCAB, focusing on their methodologies and regret analyses. We also examine the connections between SCAB problems and functional data analysis.
Finally, we highlight recent advances and ongoing challenges in the field.
}

\begin{document}


\section{Introduction}

Reinforcement Learning (RL) is one of the most prominent and widely discussed methods in artificial intelligence, primarily focusing on how an agent learns to make decisions by interacting with an environment to maximize cumulative rewards \citep{richard2012barto,sugiyama2015statistical}. RL has been extensively applied in various domains, including autonomous driving \citep{maurer2016autonomous}, recommendation systems \citep{zhou2017large}, unmanned aerial vehicles \citep{del2021unmanned}, large language models (DeepSeek-R1,~\citep{guo2025deepseek}; RL with human feedback \citep{ouyang2022training}), financial trading \citep{shen2015portfolio}, causal inference~\cite{wager2024causal}, digital twin (DT)~\cite{cronrath2019enhancing,xia2021digital}, and~precision medicine \citep{durand2018contextual,lu2021bandit}.

Among the various models in RL, one of the most fundamental and widely studied problems is the stochastic bandit problem.
It exemplifies the {exploration-exploitation trade-off} dilemma, 
where an agent must choose between exploring new options to gather more information and exploiting known options to maximize rewards; see~\cite{bouneffouf2019survey,slivkins2019introduction}.

The current review literature on stochastic bandit algorithms highlights applications in areas such as recommendation systems \citep{elena2021survey,liu2021map,letard2024bandit}, experimental design \citep{burtini2015survey}, precision medicine \citep{lu2021bandit}, and causal inference \citep{shah2020survey}.

However, these topics remain underexplored in existing reviews, particularly from the perspective of rigorous theoretical analysis and statistical foundations.
This paper aims to address this gap by focusing on the probabilistic and statistical foundations of stochastic algorithms, with~particular emphasis on concentration inequalities, minimax rate of regret upper bounds, small-sample statistical inferences, linear models, Bayesian optimization, statistical learning theory, design of experiments, the~Neyman-Rubin causal model, functional data analysis, robust statistics, information theory, and~so~on.

\subsection{Stochastic Bandit in~RL}\label{se:RL}
A stochastic bandit, from~a statistician's perspective, can be represented as a collection of possible distributions of populations for the reward random variable (r.v.) $Y_a \sim P_a$, where $a$ is the action and $P_a$ comes from
$$\nu = \left\{P_a: a \in \mathcal{A}\right\},$$
where $\mathcal{A}$ is the set or space of available actions.

The agent and the environment interact sequentially over $T$ rounds:
        \begin{itemize}
            \item In each round $t \in \{1, \ldots, T\} =: [T]$, the~agent selects an action $A_t \in \mathcal{A}$, which is communicated to the environment. Here $T\in\mathbb{N}$ is the \textit{{horizon}
} (total number of steps).
            \item Given the action $A_t$, the~environment generates a reward $X_t \in \mathbb{R}$, drawn from the distribution $P_{A_t}$, and~discloses the reward $X_t$ to the agent.
            \item This interaction between the agent and the environment induces a probability distribution over the sequence of outcomes $(A_1, X_1, A_2, X_2, \ldots, A_T, X_T)$.
        \end{itemize} 
        
The horizon $T$ is finite due to budgetary constraints (non-asymptotic theory) in some cases, but~we may assume an infinite horizon $T = \infty$ (asymptotic theory) in theoretical settings. The~sequence of outcomes typically satisfies assumptions~\cite{lattimore2020bandit}:
    \begin{itemize}
        \item The conditional distribution of $X_t$ given $A_1, X_1, \ldots, A_{t-1}, X_{t-1}, A_t$ is $P_{A_t}$, i.e.,
        $$ {P}(X_t|A_1,X_1,\cdots,X_{t-1},A_t)= {{P}(X_t|A_t)}\sim P_{A_t}$$
        indicating that the environment samples $X_t$ from $P_{A_t}$ in round $t$. 
         \item Here $A_t$ is determined by the \textit{{history}} defined by
    	$H_{t-1}:=(A_1,X_1,\cdots,A_{t-1},X_{t-1}).$
        \item  The conditional distribution of action $A_t$ given $H_{t-1}$ is
        \[
            \pi_t\left(\cdot \mid A_1, X_1, \ldots, A_{t-1}, X_{t-1}\right), 
        \] 
    \end{itemize}
where $\pi_1, \pi_2, \ldots$ is a sequence of \textit{{probability kernels}} (\textit{policies}) characterizing the~agent.

The \textit{policy} is the action by a learner to interact with an environment. Let $\mathcal{H}=\{H_{t-1} \mid t=1,\cdots,T\}$, we denote the policy at round $t$ by $\pi_t$:
    $$\pi_t:\mathcal{H}\rightarrow\mathcal{A},~~A_t=\pi_t(H_{t-1}),~~t=1,\cdots,T.$$
    
Key assumptions above are { {that: the selected actions  do not affect the reward distribution of the arms and the agent's current decision cannot depend on future observations}}. 

The agent's objective is to maximize the total reward by designing a policy $\pi := (\pi_1, \pi_2, \ldots, \pi_T)$ to maximize the sum of rewards
$$\sum_{t=1}^T X_t,$$
which is a random quantity dependent on the agent’s actions and the rewards sampled by the environment. However, this maximization is not a classical optimization problem due to the fact that the reward $X_t$ is~random.

For a stochastic bandit $v=\left(P_a: a \in \mathcal{A}\right)$, we define
$\mu_a(v)=\int_{-\infty}^{\infty} x d P_a(x)$ if the mean exists. To~earn more reward, we prefer to choose

\clearpage

~\vspace{-12pt}
$$\mu^*(v)=\max _{a \in \mathcal{A}} \mu_a(v)$$
as the optimal mean reward among all possible actions.
One standard approach is to compare the policy's cumulative rewards to the best-action benchmark $\mu^*(v)$ : the summation of expected rewards if the agent always played the optimal action up to round $T$, which is the best possible total expected reward for a particular problem. Formally, we define the following quantity, called regret at round $T$:
\begin{equation}\label{def_reg}
    \operatorname{Reg}_T(\pi,v):=T \mu^*(v)- E [ \sum_{t = 1}^T X_t ].
\end{equation}

\noindent {Here}
, \(\operatorname{Reg}_T(\pi,v)\) is revealed to the policy \( \pi\) and the distribution $v$ of the rewards. A~desirable asymptotic property of an algorithm is termed \textit{{no-regret}} \cite{srinivas2010gaussian} if the average regret converges to 0 as \( T \) approaches infinity:
$
\lim_{T \rightarrow \infty} \frac{\operatorname{Reg}_T(\pi,v)}{T} = 0.
$

Regret quantifies the loss from not selecting the optimal action from the start. The~goal is to minimize regret by balancing exploration (testing different actions) and exploitation (choosing the best-known action).
Often, we write $\mu_a := \mu_a(v)$ when $v$ is specified. 
This framework parallels the bias-variance trade-off in statistics and machine learning: exploration introduces potential bias by prioritizing learning over immediate gains, while exploitation can amplify variance by relying on potentially incomplete~information.

\subsection{Structured and Unstructured~Bandits}

In numerous practical applications, it is often unrealistic to assume that the bandit instance, denoted by $\nu$, is fully specified or follows a parametric distribution. Instead, we often possess only partial information regarding its distribution.
To capture this uncertainty, we define a set of bandit instances $\mathcal{E}$, which encompasses all possible distributions to which $\nu$ could belong. This set $\mathcal{E}$ is referred to as the \textit{{environment class}} \cite{lattimore2020bandit}. The~classification of bandits into structured and unstructured environments is crucial in statistical inference, ranging from mean estimation to regression prediction. Structured bandits incorporate additional information or dependencies between actions, which can be exploited to improve decision-making. 
In contrast, unstructured bandits correspond to the classical formulation of the bandit problems, where each arm operates independently, and~no further relationships or information between actions are available. This distinction has a profound impact on the design and efficiency of the policy \( \pi\).

\begin{Definition}\label{defsubG}
An environment class $\mathcal{E}$ is unstructured if $\mathcal{A}$ is finite and there exist sets of distributions $\mathcal{M}_a$ for each $a \in \mathcal{A}$ such that
    $$
       \mathcal{E}=\left\{\nu=\left(P_a: a \in \mathcal{A}\right): P_a \in \mathcal{M}_a \text { for all } a \in \mathcal{A}\right\}.
    $$
\end{Definition}
\noindent {This} definition highlights the simplicity and generality of unstructured environments in bandit problems. Such environments are characterized by the following key~properties:
\begin{itemize}
        \item \textit{{Independence}}: each arm $a$ yields rewards from an unknown probability distribution $P_a$ independently of other arms.
    \item \textit{{No side information}}: there are no features or context associated with the arms.
\end{itemize}

Environment classes commonly play a pivotal role in determining the performance of learning algorithms \citep{lattimore2020bandit}. Parametric environments, such as Bernoulli and Gaussian bandits, assume specific density functions. For~non-parametric classes, like sub-Gaussian and sub-exponential bandits, do not rely on a density function assumption but are instead characterized by conditions on their moment-generating functions (see Section~\ref{se:CI}). The~correct specification of the environment is critical; failure to do so, or~relying on an incorrect model, can significantly degrade the algorithm's efficacy \citep{lattimore2020bandit,wei2023zero}. 
Depending on the underlying data-generating mechanism, the~choice of environment spans a range of distributions—from bounded distributions to those that are light-tailed or heavy-tailed. 
This flexibility underpins many problems in reinforcement learning and decision~theory. 

In this review, we will focus on bounded, sub-Gaussian, and~sub-exponential bandits, which are characterized by finite moment-generating functions \citep{zhang2022sharper}. A~detailed treatment of heavy-tailed bandits, such as those with sub-Weibull distributions or distributions with finite moments (or even infinite variance), typically requires additional techniques (see Section~2.2 in~\cite{xu2023non} and~\cite{wei2023zero}) and is left for future~work.

%



\vspace{3pt}
\noindent \textbf{{Multi-Armed Bandit Problems}
} 
\vspace{3pt}

\begin{figure}[H] 
    
    \includegraphics[width=0.53\textwidth]{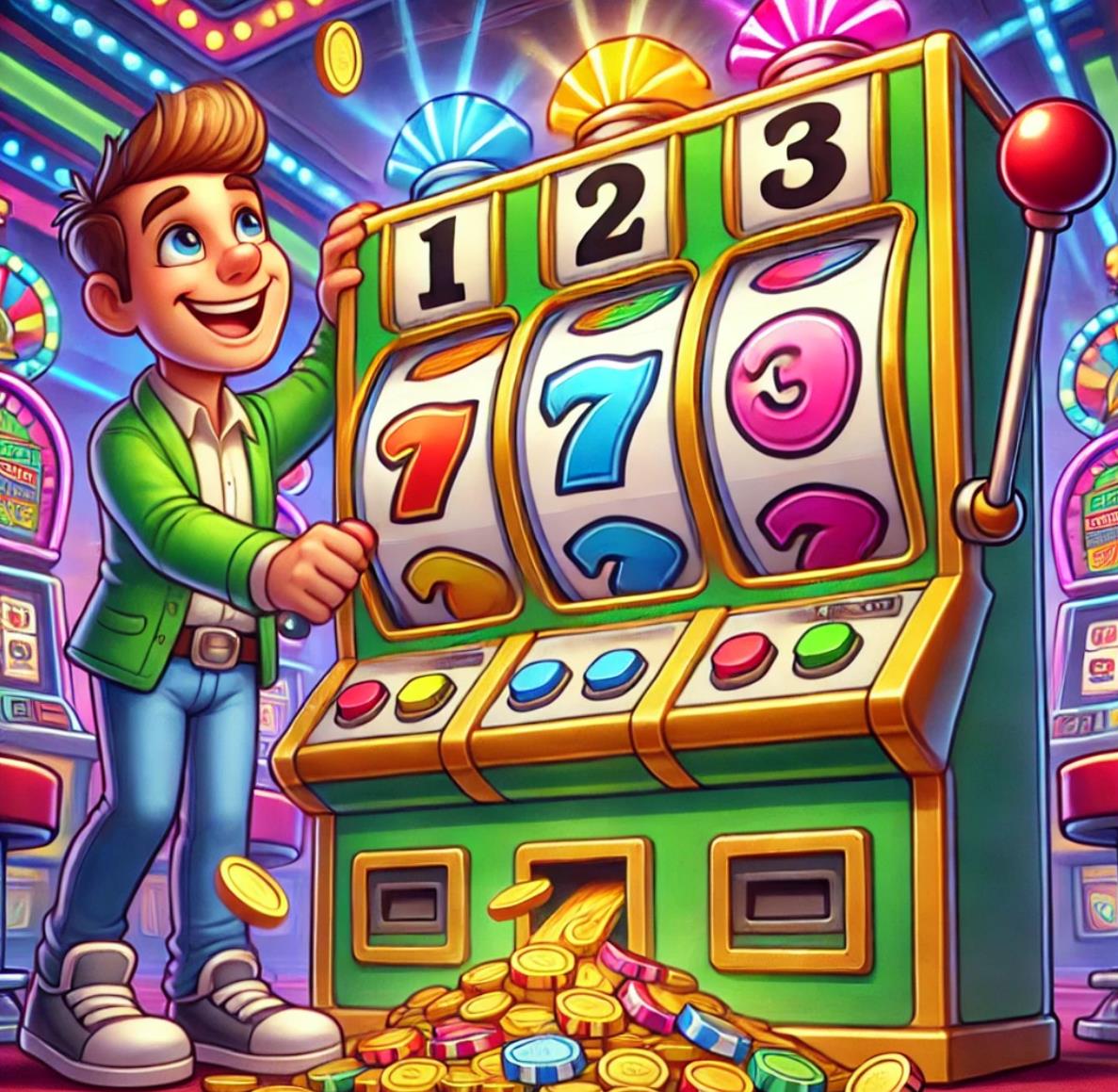} 
    \caption{{A} 
 player plays at a three-armed bandit machine in a casino.} 
    \label{fig:mab} 
\end{figure}

When $\mathcal{A} = [K]$ in stochastic bandit models with $K \in \mathbb{N}$, the~problem reduces to a multi-armed bandit (MAB) problem (Figure \ref{fig:mab}).
\begin{Example} [$K$-armed bandits, MAB]
        $\mathcal{A} = [K]$ is finite, and $\mathcal{M}_a$ only contains one probability measure for a fixed $a \in \mathcal{A}$.
\end{Example}

\vspace{-9pt}

In the field of statistics, the~time-uniform confidence sequence problem \citep{howard2021time} is often framed as an MAB problem, first introduced by~\cite{robbins1952some} in the context of sequential experimental design. 
The topic has been extensively studied in the machine learning literature, with~significant contributions documented in major journals. For~a comprehensive review, see Section~1 in~\cite{lattimore2020bandit}.
There are several compelling reasons to begin the study of bandit problems with MAB problems. First, their simplicity makes them relatively straightforward to analyze, providing a deep understanding of the fundamental trade-off between exploration and exploitation. Second, many algorithms designed for finite-armed bandits, along with the underlying principles, can be generalized to more complex settings. Lastly, finite-armed bandits have practical applications, particularly as an alternative to A/B testing as two-arm bandits, which involves random assignment of experimental units to treatment groups (A and B). For~example, in~a drug comparison experiment, patients are randomly assigned to either the new drug or standard drug control group, ensuring unbiased allocation for valid comparisons (see Example \ref{ex:N-R} below).

In MAB problems, a~typical scenario
involves an agent choosing between $K$ slot machines (a $K$-armed bandit), each with an unknown reward distribution \mbox{$\{Y_k\}_{k=1}^K \in \mathbb{R}$}. These rewards may be unbounded, non-Gaussian, or~even negative. 
Without prior knowledge of the reward distributions, it is commonly assumed that the moment-generating functions (MGFs) exist and that the distributions belong to the sub-$F$ family, denoted as $\operatorname{subF}(\mu, \sigma^2)$ {\cite{bubeck2012regret}}
. Here, $F$ encompasses distributions such as Gaussian, exponential, or~Weibull, with~$\mu$ representing the mean and $\sigma^2 > 0$ serving as a variance-dependent parameter (not necessarily the variance itself).
Given this setup, the~agent's primary goal is to maximize the cumulative rewards by identifying the optimal action, which in this context refers to selecting the \textit{arm} with the highest expected reward.
To formalize, the following is typically assumed:
 \begin{itemize}
  \item
  Each reward $Y_k$ follows the sub-$F$ family:
\begin{equation}\label{eq:REGsubg}
Y_{k}\sim \operatorname{subF}(\mu_{k},\sigma^{2}_{k}),~k \in [K].
 \end{equation}
 \item 
 The index ${k^{\star}}$ corresponding to the unique arm with the maximal mean is defined as
 $$\mu_{k^{\star}}=\max_{k \in [K]}\mu_k.$$

\end{itemize}

The objective is to identify the optimal arm while minimizing the selection of sub-optimal arms (i.e., those with rewards $\{Y_k\}_{k \neq k^{\star}}$). This creates the classic exploration–exploitation trade-off: the agent must balance exploring sub-optimal arms to gather new information (exploration) with selecting the best-known arm based on existing knowledge (exploitation).
Designing an exploration procedure for MAB with both valid theoretical guarantees and sound practical performance is a long-established and challenging problem. 
Classical works~\cite{robbins1952some,lai1985asymptotically} dealt with asymptotic results, while we focus more on non-asymptotic results via the non-parametric distribution family \eqref{eq:REGsubg}.

{
\vspace{3pt}
\noindent \textbf{{Linear Bandit Problems}} 
\vspace{3pt}
 
 Another important case of structured bandits is stochastic linear bandits, which models the expected reward of an arm as a linear function of known features.}

\begin{Example}[Stochastic linear bandit, SLB]\label{ex:SLB}
        Let $\mathcal{A} \subset \mathbb{R}^p$ and model parameter $\eta_* \in \mathbb{R}^p$ and
$$
\nu_{\eta_*}=\{N(\langle a, \eta_*\rangle, 1): a \in \mathcal{A}\}~\text { and } \mathcal{E}=
\{\nu_{\eta_*}:{\eta_*} \in \mathbb{R}^p\}.
$$
{{Various choices of $\mathcal{A}$ lead to many settings}}: 
\begin{enumerate}
\item[(1)] For unit vectors $\{e_i\}_{i=1}^K$ and  $\mathcal{A} = \{e_1, \dots, e_K\}$, if~${\eta_*} = (a_1, \dots, a_K)^{T}$ represents the mean vector, then the SLB model reduces to the Gaussian MAB  $\nu_{\eta_*} = \{N(a_i, 1)\}_{i=1}^K$ without specific structures.
\item[(2)] Given a shared feature $C$, if~$x_{k}=:\psi(C, k)\in \mathbb{R}^d$ for $ k \in[K]$ and ${\eta_*}=(\theta_1,\cdots,\theta_K)^{T}$ with $\theta_k\in \mathbb{R}^d$, where $\psi(\cdot, \cdot)$ is a link function. Then, $\mathcal{A}=\{(\mathbf{0}_{d(i-1)}, x_k^T, \mathbf{0}_{d(K-i)})\}_{i=1}^K \subset \mathbb{R}^{d K}$ becomes a $K$-arm contextual linear bandit with $k$ disjoint linear models $\{N(x_{k}^T \theta_k,1)\}_{k=1}^K$, and~$p=dK$ [see Section~\ref{sec:LinUCB} and Section~2.2.3 of \citep{lu2021bandit} for details].
\end{enumerate}
\end{Example}
\noindent Key characteristics of~SLB are as follows:
\begin{itemize}
\item \textit{{Arms are related}}: arms are related through a shared feature, $C$, or known relationships.
\item \textit{{Side information}}: each arm may have an associated $d$-dimensional feature (context vector), which provides additional information.
\end{itemize}

We aim to estimate the unknown $\eta_*$ in Example \ref{ex:SLB} to select arms with the highest expected reward.
In general SLB problems, the~reward of an action is typically modeled as a sub-Gaussian or sub-F r.v. with a mean that is the inner product of the action vector and an unknown parameter vector $\eta$.
Even if $\mathcal{A}$ or $\mathcal{C}$ is large, the~agent can determine the environment by a scalar product of \textit{feature map} $\psi: \mathcal{C}\times\mathcal{A} \rightarrow \mathbb{R}^d$ [see Section~19.1 in \citep{lattimore2020bandit}] and an unknown parameter vector $\eta_* \in \mathbb{R}^d$

\clearpage
~\vspace{-33pt}

$$
r(c, a)=\left\langle\eta_*, \psi(c, a)\right\rangle~\text{and}~\nu_{\eta_*}=\{\operatorname{subF}(r(c, a), 1): a \in \mathcal{A}\}\quad \text { for all }(c, a) \in \mathcal{C} \times \mathcal{A}.
$$
{{Such} an SLB framework becomes particularly interesting when $\eta_*$ is sparse  in high-dimensional settings, as~studied in Chapter 23 of \citep{lattimore2020bandit} and recent works like~\cite{wang2020nearly, hao2020high, wang2023efficient, fan2023provably}.}

For a general reward function $r(c, a)$, if~the variance-dependent parameter $\sigma^2$ is unknown, the~problem extends naturally to stochastic contextual bandits (SCB), which is detailed in the following~example.

\begin{Example}[Stochastic Contextual Bandits, SCB]\label{ex:SCB} 
Given a sub-$F$ family and a reward functional class $\mathcal{R}$, the~environment class for SCB is defined as
$$
\nu_{\mathcal{R},F}=\{\operatorname{subF}(r(c, a),\sigma^{2}): (c, a) \in \mathcal{C} \times \mathcal{A}\}~\text { and } \mathcal{E}=
\{\nu_{\mathcal{R},F}\}.
$$
\end{Example}

The unknown variance parameter $\sigma^2$ introduces additional challenges in algorithm design and theoretical analysis, as~recently highlighted by~\cite{zhang2023tight}. Contextual bandits have vital applications like recommendations, advertising, and~web search, where rewards might be user clicks, video views, or~revenue~\cite{li2019perspective}.

\subsection{Stochastic Continuum-Armed~Bandits}\label{se:SCB}

{Stochastic Continuum-Armed Bandits (SCAB,~\cite{agrawal1995continuum}) extend the classical $K$-armed bandit problem by allowing the set of possible actions (or arms) to lie in a continuous space $\mathcal{A}$ rather than in a discrete set $\mathcal{A}=[K]$ or $\mathcal{A} = \mathbb{N}$. 
To distinguish this continuous action space, we denote it as $D$. In~this setting, the~agent seeks to identify the optimal reward point from a continuous domain, typically modeled as an interval or subset of $\mathbb{R}^d$.}

In statistics, the~optimal design in SCAB problems belongs to {{optimal design of experiments}} \cite{pukelsheim2006optimal} and Bayesian optimization \citep{garnett2023bayesian}. Statistical analysis on the data with a continuous domain $\mathcal{A}$ is closely related to functional data analysis. The~reward function in SCAB is typically assumed to possess a degree of smoothness, meaning that it can be represented as a stochastic process that exhibits regularity over space, such as a Lipschitz-continuous function or a Gaussian process (GP,~\cite{williams1995gaussian}).

The fundamental objective in SCAB, as~in the traditional bandit setting, is to effectively balance \textit{{exploration}}—sampling from various points in the continuous action space to gather information about the reward function—and \textit{{exploitation}}—leveraging current knowledge to select actions believed to yield the highest expected reward. The~continuous nature of the action space introduces additional complexity, as~the agent must navigate an infinite number of potential actions, requiring more advanced algorithms for efficient exploration and optimization. These methods often rely on the smoothness of the reward function to guide the search for optimal actions while minimizing cumulative~regret.

Formally, we focus on the {{sequential optimization}} of an unknown reward function $f: D \rightarrow \mathbb{R}$ as follows:

\begin{enumerate}
\item \textbf{{Objective}}: The primary goal is to identify the maximum point:
\begin{equation}\label{eq:mp}
        x^{\star} = \arg \max_{x \in D} f(x),
\end{equation}
as a {{black-box optimization}} without any geometric structure of $f$.

 \item \textbf{{Action selection}}: in each round $t$, the~agent chooses a fixed point $x_t \in D$ and receives an observed value perturbed by noise $\epsilon_t$:
    $$
    y_t = y_t(x_t) = f(x_t) + \epsilon_t,~\text{where}~E[y_t] = f(x_t).
    $$

 \item \textbf{{Decision-making}}: given the uncertainty of the maximum of $f$, we aim {for} 
$$maximizing~the~expected~total~reward~\sum\nolimits_{t=1}^{T} f(x_t)~over~a~finite~time~horizon~T.$$ 

\item \textbf{{Performance metric}}: given $f\in \mathcal{F}$ (fixed or random functional class) and $\epsilon_t \sim $sub-$F$ family, to~quantify the loss of reward, the~cumulative regret after rounds $T$ is 
    $$
    \operatorname{Reg}_T(\pi;\mathcal{F},v):= \sum_{t=1}^{T} (f(x^{\star}) - f(x_t)),
    $$
    where the instantaneous regret is given by $f(x^{\star}) - f(x_t)$.
    
\end{enumerate}

Despite its added complexity, the~essence of designing algorithms for MAB, SLB, or~even SCAB remains fundamentally the same: accurately estimating the reward function while balancing the exploration–exploitation trade-off. This requires technical tools to handle the uncertainty in the observed rewards and ensure reliable decision-making under limited information. One such tool is concentration inequalities, providing confidence intervals to quantify uncertainty in reward estimates, which we will introduce in the next~section.

\section{Concentration~Inequalities}\label{se:CI}

Building on the need for technical tools to handle uncertainty in observed rewards and ensure reliable decision-making, concentration inequalities play a pivotal role in deriving robust guarantees. 
 These tools are particularly important when drawing conclusions with minimal data assumptions, a~fundamental challenge in machine learning.
Typically, inference relies on confidence intervals under specific distributional assumptions \citep{fisher1922mathematical}. However, exact distributions are often unavailable or too complex. Instead, we may assume the data belong to sub-classes like sub-Gaussian \citep{kahane1960} or sub-exponential \citep{cramer1938nouveau} distributions. These assumptions are widely used in non-asymptotic inference and machine learning to derive concentration inequalities with exponential~decay.

\subsection{Basic Concentration~Inequalities}\label{se:baCI}

Concentration inequalities are a commonly used method to quantify the degree of concentration of a measure. 
Specifically, concentration inequalities quantify the extent to which a random variable $X$ deviates from its mean ${E}X = \mu$ by expressing the measure of concentration of $X - \mu$ through one-sided or two-sided tail probabilities (denoted by $t > 0$ for deviation):
\begin{equation}\label{eq:two-sided tail}
P(X - \mu > t) \le \delta_t \quad \text{or} \quad P(|X - \mu| > t) \le \delta_t,
\end{equation}
where $\delta_t$ represents the tail probability. For~suitably large $t$, $\delta_t$ can be made arbitrarily small, providing strong guarantees for controlling~deviations.

Tail probabilities and expectations are closely linked. From~the relationship between tail probabilities and expectation ({see Theorem 12.1(1) in} 
 \cite{gut2013probability}), the~expectation can be bounded as
\begin{equation}\label{eq:tail probabilities}
E|X - \mu| = \int_0^\infty P(|X - \mu| > t) \, dt \le \int_0^\infty \delta_t \, dt.
\end{equation}
This shows that expectation bounds can be viewed as concentration inequalities via integral transforms. Conversely, tail probabilities can be bounded using expectations, as~illustrated by the widely used Markov's inequality:
\begin{Lemma}[Markov's Inequality]
Let $\varphi(x) : \mathbb{R} \to \mathbb{R}^{+}$ be a non-decreasing function. For~r.v. $X$ with $E[\varphi(X)] < \infty$,
\begin{equation}\label{eq:Markov's Inequality}
P(X \geq a) \leq \frac{E[\varphi(X)]}{\varphi(a)}, \ \forall a \in \mathbb{R}.
\end{equation}
\end{Lemma}
\begin{proof}[Proof]
By the positivity and the non-decreasing property of $\varphi$, we obtain
$$P(X \geq a) = E[I\{X \geq a\}] \leq E\left[ \frac{\varphi(X)}{\varphi(a)} I\{X \geq a\} \right] \leq E[\varphi(X)] \frac{1}{\varphi(a)}.$$
\end{proof}
Using Markov's inequality, we derive Chebyshev's inequality in the following~lemma.

\begin{Lemma}[Chebyshev's inequality]
Let $X$ be a zero-mean r.v. with finite $\operatorname{Var} \, X$; then,
\begin{equation}\label{eq:Chebyshev's Inequality}
P(|X| \geq a) \leq \frac{\operatorname{Var} \, X}{a^2},~~ \forall  a \in \mathbb{R}^{+}.
\end{equation}
\end{Lemma}
\begin{proof}[Proof]
Chebyshev's inequality follows from Markov's inequality with $\varphi(x) = x^2$.
\end{proof}
Chebyshev's inequality is specific to deviations from the mean and depends on the variance.
While it is widely used in probability limit theory for its simplicity, it provides a tail inequality with a polynomial decay rate $O(a^{-2})$.
However, in~many scenarios within statistics and machine learning, such polynomial decay rates are insufficient, especially when dealing with high-dimensional data or rare events. In~these cases, we require tail probabilities with exponential decay for sharper bounds.
For Gaussian random variables, Mill’s inequality (Lemma A.2.1 in~\cite{gut2013probability}) provides an exponential tail bound:
$$\left( \frac{x}{x^2+1} \right) \cdot \frac{e^{-x^2/2}}{\sqrt{2\pi}} \leq P(X \geq x) \leq \frac{1}{x} \cdot \frac{e^{-x^2/2}}{\sqrt{2\pi}} \text{if}~X \sim {N}(0,1), \text{for}~x > 0.$$
This inequality is widely used in probability theory and stochastic processes to bound the tail probabilities of Gaussian random variables, demonstrating their exponential decay.
A more refined version of Mill’s inequality (Lemma B.4 in~\cite{giraud2021introduction}) is as follows:
\begin{Lemma}[A refined Mill's Inequality]\label{Mill}
If $X \sim {N}(0, \sigma^2)$, for~$x > 0$, we have:
\begin{equation}
P(|X| \geq x) \leq e^{-x^2/(2\sigma^2)} \quad.
\end{equation}
\end{Lemma}
\noindent Lemma \ref{Mill}
removes the factor $x^{-1}$ from the original Mill's inequality, providing a cleaner bound with an exponential decay rate of $O(e^{-a^2})$.
To illustrate the limitations of polynomial decay rates, we present an example from high-dimensional statistics~below.

\begin{Example}[$O(a^{-2})\text{-decay tail inequality is not enough}$]
Consider r.v.s.
$\{X_{ij}\}\overset{i.i.d.}{\sim} {N}(0, \sigma^2)$ for $i = 1, \dots , n \text{ and } j = 1, \dots , p_n$, where $p_n \gg n \to \infty.$
Then
$\sum_{i=1}^n X_{ij} \overset{i.i.d.}{\sim} {N}(0, n\sigma^2)$, we have
\[
T_n := P\left( \max_{1 \leq j \leq p} \left| \sum_{i=1}^n X_{ij} \right| \geq t \sqrt{n} \right) \leq \sum_{j=1}^p P\left( \left| \sum_{i=1}^n X_{ij} \right| \geq t \sqrt{n} \right) \leq \frac{p \sigma^2}{t^2}.
\]
by Chebyshev's inequality. Refined Mill's inequality gives
\[
T_n \leq \sum_{j=1}^p P \left( \left| \sum_{i=1}^n X_{ij} \right| \geq t \sqrt{n} \right) = p e^{-(t\sqrt{n})^2 / (2n\sigma^2)} = p e^{-t^2 / (2\sigma^2)}.
\]

Setting $t = \sqrt{p}$ and letting $p = p_n \to \infty$, Chebyshev's inequality gives
$T_n \leq p \sigma^2 / t^2 = \sigma^2 \neq 0,$ whereas refined Mill's inequality ensures
$T_n \leq p e^{-p^2 / (2\sigma^2)} \to 0.$
\end{Example}

This example demonstrates that Chebyshev's inequality does not guarantee tail probabilities approaching zero as the dimensionality increases, while exponential decay guarantees (like Mill’s inequality) can ensure this property. Such exponentially decaying tail inequalities are indispensable in high-dimensional statistics and machine learning to achieve desired~results.

Another more general and powerful inequality for obtaining exponentially decaying bounds is Chernoff's inequality. It applies to random variables with MGFs and is particularly useful for sums of independent random variables.
\begin{Lemma}[Chernoff's inequality, or~exponential Markov inequality]
For a r.v. $X$ with $E[e^{tX}] < \infty$ for all $t > 0$, we have
\begin{equation}    
    P(X \geq a) \leq \inf_{t > 0} \left\{ e^{-ta} E e^{tX} \right\}.
\end{equation}
\end{Lemma}
\begin{proof}[Proof]
Applying Markov's inequality with $\varphi(x) = e^{tx}$ gives
$P(X \ge a) \le {e^{ - ta}}{E}{e^{tX}}$, and~by minimizing over $t > 0$ yields the bound.
\end{proof}

The advantage of Chernoff's inequality lies in its ability to provide exponentially decaying tail probabilities, making it considerably sharper as deviations increase. 
This property is particularly valuable in scenarios involving rare events, such as analyzing sums of independent random variables. Its exponential decay rate makes Chernoff's inequality a powerful tool in applications like bandit algorithms.
As a concrete example, in~1963, Hoeffding introduced a concentration inequality for sums of independent bounded r.v.s., achieving \(O(e^{-a^2})\)-decay.
\begin{Lemma}[Hoeffding's inequality, Theorem 2 in~\cite{doi:10.1080/01621459.1963.10500830}]
Let $\{X_i\}_{i=1}^{n}$ be independent r.v.s satisfying the bounded condition $a_i \leq X_i \leq b_i$. Then,
\begin{equation}
P\left( \left| \sum_{i=1}^{n} (X_i - EX_i) \right| \geq t \right) \leq 2\exp \left( \frac{-2t^2}{\sum_{i=1}^{n} (b_i - a_i)^2} \right).
\end{equation}
\end{Lemma}
For applications, Hoeffding's inequality can be used to construct confidence intervals (CIs) for the probability of heads in coin-tossing~problems.

\begin{Example}[Confidence intervals by Hoeffding's inequality]
Suppose $X_1, \ldots, X_n$ are i.i.d. r.v.s with $X_i \sim \text{Bernoulli}(p)$, where $p$ represents the probability of heads. Let $\overline{X}_n := \frac{1}{n} \sum_{i=1}^n X_i$. For~any $\epsilon > 0$, Hoeffding's inequality gives
${P}\left(\left|\overline{X}_n - p\right| > \epsilon\right) \leq 2 e^{-2 n \epsilon^2} = \alpha$. By~setting $\epsilon_{n, \alpha} = \sqrt{\frac{1}{2 n} \log \frac{2}{\alpha}}$,
 we can construct a $(1 - \alpha)100\%$ confidence interval for $p$ as
\[
   \left[\overline{X}_n - \epsilon_{n, \alpha}, \, \overline{X}_n + \epsilon_{n, \alpha}\right].
\]
This interval ensures that the $p$ lies within the bounds with at least probability $1 - \alpha$.
\end{Example}
Hoeffding's inequality also finds application in bounding the deviation of the empirical distribution function $\mathbb{F}_n(x)$ from the true distribution $F(x)$.
\begin{Example}[Empirical distribution function, EDF]\label{eg:edf}
Let $\{ {X_i}\}_{i = 1}^n \stackrel{\rm i.i.d.}{\sim} F(x)$ for a distribution $F(x)$. Let $\mathbb{F}_{n}(x):={\frac  1n}\sum _{{i=1}}^{n}{\rm{1}}_{{\{X_{i}\leq x\}}},~x\in {\mathbb{R}}$ be the empirical distribution. By~Hoeffding's inequality (using $a_i-b_i=1/n$), we have
$$P(|\mathbb{F}_{n}(x)-F(x)|>\varepsilon  )\leq 2e^{-2n\varepsilon ^{2}},~\forall \varepsilon >0.$$
\end{Example}
This result in Example \ref{eg:edf} ensures that, for~a fixed $x$, the~probability of deviation decreases exponentially as the sample size n increases.
To extend this pointwise result to a uniform bound across all $x \in \mathbb{R}$, we rely on a stronger concentration inequality known as the Dvoretzky-Kiefer-Wolfowitz (DKW) inequality \citep{dvoretzky1956asymptotic}.
\begin{equation}\label{eq:DKW}
    P{\Bigl (}\sup _{x\in \mathbb{R} }|\mathbb{F}_{n}(x)-F(x)|>\varepsilon {\Bigr )}\leq 2e^{-2n\varepsilon ^{2}}\quad \forall \varepsilon >0,
\end{equation}
which provides a uniform version of Hoeffding's inequality.

\subsection{Sub-Gaussian and Sub-Exponential Concentration~Inequalities}\label{se:baCI}

When the r.v.s are unbound, such as Gaussian variables, the~classical Hoeffding's inequality fails for non-asymptotic analysis.  To~address this, the~concept of sub-Gaussian r.v.s is introduced, allowing Hoeffding-type concentration results to be extended to sums of unbounded random variables. Sub-Gaussianity is widely used in statistical machine learning research, where data often exhibit Gaussian-like tail behavior. Specifically, a~r.v. $X$ is considered sub-Gaussian if it satisfies a Gaussian-like moment-generating function $E{e^{sX}} \approx {e^{{\rm{Var}}(X){s^2}/2}}$ or tail probability $P(|X| \ge x) \mathbin{\lower.3ex\hbox{$\buildrel<\over{\smash{\scriptstyle\sim}\vphantom{_x}}$}} {e^{ - {x^2}/[2{\rm{Var}}(X)]}}$; see~\cite{zhang2020concentration}. For~a rigorous definition, the~sub-Gaussian class is characterized by the MGF bound.
\begin{Definition}\label{defsubG}
A {zero-mean} r.v. $X \in \mathbb{R}$  is sub-Gaussian with \textit{variance proxy} ${\sigma^2}$  if its MGF satisfies
\begin{equation}
E[e^{sX}] \leq e^{{\sigma^2}{s^2}/2},\; \forall s \in \mathbb{R}.
\end{equation}
We denote $X$ by $X \sim {\mathrm{subG}}({\sigma^2})$.
\end{Definition}
Such definition allows us to derive sub-Gaussian concentration inequalities using Chernoff's inequality.
\begin{equation}
P(X \ge t) \leq \mathop{\inf}\limits_{s > 0} {e^{-st}} E[e^{sX}] \leq \mathop{\inf}\limits_{s > 0} {e^{-st + \frac{{\sigma^2 s^2}}{2}}} = {e^{-\frac{t^2}{2\sigma^2}}}, s = t/\sigma^2.
\end{equation}
Similarly, we have $P(-X \ge t) \leq {e^{-{t^2}/(2\sigma^2)}}$, and~thus, 
$P(|X| \ge t) \leq 2{e^{-{t^2}/(2\sigma^2)}}.$

By leveraging independence, this concentration property extends to sums of independent sub-Gaussian~r.v.s.

\begin{Theorem}[Concentration inequalities for sub-Gaussian sums]\label{prop:Sub-Gaussian}
 Assume $\{ {X_i}\} _{i = 1}^n$ are independent zero-mean r.v.s, where ${X_i} \sim {\rm{subG}}(\sigma _i^2)$.~Then,
 \begin{enumerate}
\item[(a)] The sum $\sum\nolimits_{i=1}^n{X_i} \sim {\rm{subG}}(\sum_{i=1}^n\sigma _i^2)$, and~for any  $t \geq 0$
\begin{equation}\label{eq:Sub-Gaussian}
P\left( {\frac{1}{n}\left| {\sum\limits_{i = 1}^n {{X_i}} } \right| \ge t} \right) \leq 2\exp \left\{ { - {n t^2}/\left( {\frac{2}{n}\sum\limits_{i = 1}^n {\sigma _i^2} } \right)} \right\}.
\end{equation}
\item[(b)] Finite mixture sub-Gaussian: 
$$\sum_{i=1}^{m} p_i\operatorname{subG}(\sigma_i^{2})\sim \operatorname{subG}\left(\max_{i\in[m]} \sigma_{i}^{2}\right)~\text{for}~\sum_{i=1}^{m} p_i=1, p_i\ge 0, m<\infty,$$
where $Z \sim \sum_{i=1}^{m} p_i\operatorname{subG}(\sigma_i^{2})$ means $Z \sim \operatorname{subG}(\sigma_i^{2})$ with the probability $p_i>0$.
\item[(c)] If $X \sim \operatorname{subG}(\sigma^{2})$, then
$${E}|X|^{k} \leq(2 \sigma^{2})^{k / 2} k \Gamma(\frac{k}{2})~and~\|X\|_k:=[{E}(|X|^{k})]^{1 / k} \le \sigma e^{1/ e}{k^{  1 / 2}},~k \geq 2.$$
\item[(d)] If $X \sim \operatorname{subG}(\sigma^{2})$, then
$\sigma^2 \ge \operatorname{Var}X$.
\end{enumerate}
\end{Theorem}
\begin{proof}\leavevmode
\begin{enumerate}
\item[{(a)}] {By} 
 the independence, 
$$E e^{t(\sum_{i=1}^{n}X_{i})}=\prod_{i=1}^{n} E e^{t X_{i}}\leq \prod_{i=1}^{n}  e^{\sigma_i^2 \frac{t^2}{2}}=e^{\sum_{i=1}^{n} \sigma_i^2\frac{t^2}{2}}, ~\forall \, t \in \mathbb{R}.$$
\item[{(b)}]  Let $Z\sim \sum_{i=1}^{m} p_i\operatorname{subG}\left(\sigma_i^{2}\right)$ and $Z_i \sim \operatorname{subG}\left(\sigma_i^{2}\right)$, we have
$$E e^{tZ}=\sum\nolimits_{i=1}^{m}p_i E e^{t Z_{i}}\le \sum\nolimits_{i=1}^{m}p_i  e^{\sigma_{i}^{2} t^2 / 2}\le \sum\nolimits_{i=1}^{m}p_i  e^{\max_{i} \sigma_{i}^{2} t^2 / 2}=e^{ \max_{i} \sigma_{i}^{2} t^2 / 2}, ~\forall \, t \in \mathbb{R}.$$
\item[{(c)}]  It relies on transforming
tail bound to moment bound \eqref{eq:tail probabilities}:
\end{enumerate}\vspace{-15pt}

\begin{align*}
{E}|X|^{k}&=\int_{0}^{\infty} {P}\left(|X|^{k}>t\right) \mathrm{d} t =\int_{0}^{\infty} {P}\left(|X|>t^{1 / k}\right) \mathrm{d} t\leq 2 \int_{0}^{\infty} e^{-\frac{t^{2 / k}}{2 \sigma^{2}}} dt \\
[\text{Put}~u=\frac{t^{2 / k}}{2 \sigma^{2}}]~&{=}\left(2 \sigma^{2}\right)^{k / 2} k \int_{0}^{\infty} e^{-u} u^{k / 2-1} \mathrm{~d} u=\left(2 \sigma^{2}\right)^{k / 2} k \Gamma(k / 2).
\end{align*}
The second statement follows from
$$\Gamma(k / 2) \leq(k / 2)^{k / 2}~\text{and}~k^{1 / k} \leq e^{1 / e}~\text{for any}~k \geq 2.$$
It yields
$$
\|X\|_k=\Big[\big(2 \sigma^{2}\big)^{k / 2} k \Gamma(k / 2)\Big]^{1 / k} \leq k^{1 / k} \sqrt{\frac{2 \sigma^{2} k}{2}} \leq \sigma e^{1/ e}{k^{  1 / 2}}.
$$
\begin{enumerate}
\item[{(d)}]  By Taylor's expansion of MGF,
\end{enumerate}\vspace{-12pt}

$${\frac{\sigma^2 s^{2}}{2}}+o(s^2)={e^{\frac{\sigma^2 s^{2}}{2}} -1 \ge {E}e^{s X}-1}= s{E} X +\frac{s^{2}}{2}{E} X^2+\cdots=  \frac{s^{2}}{2}\cdot\operatorname{Var}X+o\left(s^{2}\right)$$
which implies $\sigma^2 \ge \operatorname{Var}X$ by dividing $s^{2}$ on both sides and taking $s \rightarrow 0$.
\end{proof}

\begin{Example}
Consider $\{X_i-\mu\}_{i = 1}^n \stackrel{\rm{i.i.d.}} \sim \operatorname{subG}(\sigma^2)$.
By applying Theorem \ref{prop:Sub-Gaussian}(a), we can construct a non-asymptotic $100(1-\alpha)\%$ CI for $\mu$ as
\begin{equation}\label{eq:ci}
    \mu \in [{\overline X_n} \pm \sigma\sqrt {{2}{n^{-1}}\log ({2}/{\alpha})}],
\end{equation}
If an estimate of the sub-Gaussian parameter $\widehat{\sigma}^2$ is available, the~CI can be refined as (see~\cite{zhang2023tight})
$$
\mu \in\left[\overline{X}_n-\sqrt{2 \widehat{\sigma}^2 n^{-1} \log (2 / \alpha)}, \overline{X}_n+\sqrt{2 \widehat{\sigma}^2 n^{-1} \log (2 / \alpha)}\right] .
$$
\end{Example}


The growth moment condition presented in Theorem \ref{prop:Sub-Gaussian}(c) establishes that the normalized $k$-th moment, $\|X\|_k$, is bounded above by $O(k^{1/2})$.
This result serves as a practical tool for assessing whether an unbounded random variable exhibits sub-Gaussian behavior. Furthermore, Theorem \ref{prop:Sub-Gaussian}(d) highlights that the variance proxy not only quantifies the rate of tail probability decay but also provides an upper bound for $\operatorname{Var} X$.
Importantly, many common distributions, including the normal distribution, mixtures of Gaussian, and~all bounded distributions, belong to the sub-Gaussian class.
Additionally, Theorem \ref{prop:Sub-Gaussian}(a) and (b) demonstrate that the sum or mixture of independent sub-Gaussian r.v.s retains the sub-Gaussian property, preserving this key property under these~operations. \\
 
As a practical application of Theorem \ref{prop:Sub-Gaussian}, consider estimating the causal effect of a specific treatment on a disease outcome using the Neyman-Rubin causal~model.

\begin{Example}[Neyman-Rubin causal model as two-armed bandit, Chapter 18 in~\cite{duchi2024lecture}]\label{ex:N-R}
Each individual $i \in [n]$ has potential outcomes \((Y_0(i), Y_1(i)) \sim (\operatorname{subG}(\sigma^2), \operatorname{subG}(\sigma^2))\), where the following is true:
\begin{itemize}
    \item $Y_0(i)$: the individual's response under the control (no treatment);
    \item $Y_1(i)$: the individual's response under treatment.
\end{itemize}

\noindent {These} 
 potential outcomes are unobservable simultaneously, as~an individual can only receive one/no treatment. Two-armed bandit model, assuming that $A_i$ does not depend on the individual $i$,~implying the following:
\begin{itemize}
    \item ${E}\left[Y_a(i) \mid A_i=a\right]=\mu_a={E}\left[Y_a(i)\right]$ for $a \in \{ 0, 1\}$;
    \item the (unobservable) treatment effect is given by ${E}[Y_1(i)-Y_0(i)]$.
\end{itemize}

\noindent \textls[-5]{{Assign} $n / 2$ individuals to treatment and $n / 2$ to control, uniformly at random.
By \mbox{Theorem \ref{prop:Sub-Gaussian}(a)},} the~estimator
\[
    \begin{aligned}
        \widehat{\tau} & :=\frac{1}{n / 2} \sum_{i: A_i=1} Y_i (A_i)-\frac{1}{n / 2} \sum_{i: A_i=0} Y_i (A_i) \\
        & \sim  \operatorname{subG}\left(\frac{2\sigma^2}{n}\right) + \operatorname{subG}\left(\frac{2\sigma^2}{n}\right) \stackrel{d}{=} \operatorname{subG}\left(\frac{4\sigma^2}{n}\right).
    \end{aligned}
\]
is unbiased for $\tau=\sum_{i=1}^n{E}[Y_1(i)-Y_0(i)]/n$, with~
a confidence interval derived as
\[
    P \Big( |\widehat{\tau}-\tau| \le {2\sigma}\sqrt {2{n^{ - 1}}\log (2/\alpha )} \Big) \geq 1 - \alpha.
\]
\end{Example}

In the definition of a sub-Gaussian r.v., the~MGF satisfies \(E[e^{s X}] \leq \exp\left(\frac{\sigma^2 s^2}{2}\right)\) for all \(s \in \mathbb{R}\). 
However, such stringent conditions may exclude certain r.v.s that exhibit sub-Gaussian-like behavior. For~instance, consider the following~example.

\begin{Example} [MGF of exponential distributions]\label{eg:Exponential}
Let $X \sim {\rm{Exp}}(\mu)$ with density function\linebreak $f(x) = {\mu ^{ - 1}}{e^{ - x/\mu }} \cdot I(x > 0)$ and ${E}X=\mu>0$. For~$X-\mu$, the~MGF satisfies
$$ {{E}}{e^{s(X - \mu )}} = {e^{ - s\mu }}{\left( {1 - {\rm{s}}\mu } \right)^{ - 1}} = {\left( {\frac{{{e^{ - s\mu /2}}}}{{\sqrt {1 - s\mu } }}} \right)^2} \le {e^{2{{(s\mu /2)}^2}}} ={e^{{s^2}\mu^2/2}} , \quad
\forall ~ |s| \le (2\mu)^{-1},$$
where the last inequality is by ${e^{-2t}}/({1-2 t}) \leq e^{4 t^{2}}$ for $|t|\le 1/4$ ({By} 
 the property of $f(t): = (1 - 2t){e^{4{t^2} + 2t}}$ with $f(0)=1$: (a). $f'(t) > 0,~0 < t < 1/4$; (b). $f(t) \ge 1, ~- 1/4 < t < 0$).
\end{Example}

\begin{Definition}[Sub-exponential distribution,~\cite{wainwright2019high}]\label{def:Sub-exponential}
 A r.v. $X \in \mathbb{R}$ with mean zero is sub-exponential with two non-negative parameters $(\lambda, \alpha)$ (denoted by $X \sim \operatorname{subE}(\lambda, \alpha)$):
$$E [e^{s X}] \leq e^{\frac{s^{2}\lambda ^{2}}{2}}~\text { for all }|s|<\frac{1}{\alpha}.$$
\end{Definition}

This definition asserts a \textit{locally} sub-Gaussian property for sub-exponential r.v.s, where the MGF is bounded within a neighborhood of zero. However, the~bound does not hold for large $s$.
The following theorem provides equivalent characterizations of~sub-exponentiality.

\begin{Theorem}[Characterizations of sub-exponentiality, Lemma 2.2 in~\cite{petrov1995limit}] Let $X$ be a r.v.
The following are equivalent:
\begin{enumerate}
    \item There exists a positive constant $h$ such that
${Ee}^{t X}<\infty \text { for }|t|<h.$

\item There exists a positive constant $a$ such that
${Ee}^{a|X|}<\infty$.

\item There exist positive constants $b$ and $c$ such that
$$
P(|X| \ge x) \le b {e}^{-c x}~\text{ for all } x>0
$$
If ${E} X=0$, the~above statements are each equivalent to the assertion:
\item There exist positive constants $g$ and $r$ such that
$Ee^{t X} \le {e}^{g t^2}~\text { for }|t| \le r$. 
\end{enumerate}
\end{Theorem}

The first characterization, known as \emph{{Cramér's condition}
}, serves as a fundamental criterion: \emph{{A random variable is sub-exponential if its MGF exists in a neighborhood around zero.}} This criterion encompasses a broad class of light-tailed distributions with exponential decay in their tail probabilities. 
It is particularly valuable in machine learning applications, where most real-world data tend to follow light-tailed distributions rather than heavy-tailed~ones.

\begin{Theorem}[Concentration inequalities for sub-exponential sums; see Corollary 4.2 in~\cite{zhang2020concentration}]\label{sub-exponentialConcentration}
Let $\{ X_{i}\} _{i = 1}^n $ be independent zero-mean r.v.s. with $X_i \sim\operatorname{subE}(\lambda_i,\alpha_i )$. Define
$$\alpha:=\max_{1 \le i \le n} \alpha_i>0,~\| \boldsymbol\lambda \|_2:= (\sum\nolimits_{i = 1}^n {\lambda _i^2} )^{1/2}~and~\overline \lambda : = (\frac{1}{n}\sum\nolimits_{i = 1}^n {\lambda _i^2} )^{1/2}.$$
\begin{enumerate}
\item[\emph{(1).}] {\rm Closed under summation} {{$\sum_{i = 1}^n {{X_i}}  \sim {\rm{subE}}(\| \boldsymbol \lambda \|_2,\alpha)$}};
\item[\emph{(2).}] {\rm SubG+SubE decay}
\end{enumerate}

$$P\left( {\frac{1}{n}\left| {\sum\limits_{i = 1}^n {{X_i}} } \right| \ge t} \right) \le 2{e^{ - \frac{1}{2}(\frac{{n{t^2}}}{{{{\overline \lambda }^2}}} \wedge \frac{{nt}}{\alpha })}}= \left\{ {\begin{array}{*{20}{c}}
2{e^{ - \frac{{n{t^2}}}{{2{{\overline \lambda }^2}}}}},~0 \le t \le \frac{{{{\overline \lambda }^2}}}{\alpha }\\
2{e^{ - \frac{{nt}}{{2\alpha }}}},~~~~t > \frac{{{{\overline \lambda }^2}}}{\alpha }
\end{array}} \right..$$
\end{Theorem}
Specially, in~Theorem \ref{sub-exponentialConcentration}(2) we have 
$P \left(\frac{1}{n} \left| \sum_{i = 1}^n {{X_i}} \right| \ge  \overline \lambda \sqrt {\frac{{2s}}{n}}  + \alpha  \cdot \frac{{2s}}{n} \right) \le 2e^{-s},~\forall~s\ge 0$ by considering two rates in $(\frac{{n{t^2}}}{{{{\overline \lambda }^2}}} \wedge \frac{{nt}}{\alpha })$ separately.

A comprehensive review of concentration inequalities for machine learning applications can be found in~\cite{zhang2020concentration}. For~contextual bandits, however, relying solely on concentration inequalities for summation is insufficient \citep{li2017provably,lattimore2020bandit,wei2023zero}. These problems often require inequalities that provide tight control over deviations of empirical processes or bounds on the expectation of maxima. Notable tools include the Dvoretzky–Kiefer–Wolfowitz (DKW) inequality \eqref{eq:DKW}, which provides uniform bounds for empirical distribution~functions.

Alternative bounds can sometimes outperform traditional inequalities. For~example, Anderson's bound offers a tighter alternative to Hoeffding's inequality in~\cite{phan2021towards}. More recently, Waudby-Smith and Ramdas
~\cite{waudby2024estimating} introduced a betting-based method for deriving confidence intervals, enhancing the accuracy of mean estimation for bounded random variables.
Lastly, while independence is a common assumption in many inequalities, concentration results such as Azuma's and McDiarmid's inequalities extend to dependent data through martingale-difference assumptions. These results are particularly useful in contextual bandit algorithms and other settings involving complex stochastic dependencies~\cite{fan2024policy}.

\subsection{Why Prefer Non-Asymptotic Confidence Intervals in Bandit Problems?}

The classical limit theory in probability \citep{petrov1975sums} provides powerful tools for large-sample analysis of estimators, often represented as sums of independent random variables. For~instance, in~bandit problems, asymptotic regret analysis for large  $T$, as~established in~\cite{lai1985asymptotically}. 

\vspace{3pt}
\noindent \textbf{{Asymptotic Confidence Intervals}} 
\vspace{3pt}

The law of large numbers (LLN) and the central limit theorem (CLT) form the cornerstone of classical asymptotic analysis.
Mathematically, let us consider
i.i.d.
r.v.s \( X_1, \ldots, X_n \) drawn from a distribution \( P \) on \( \mathbb{R} \), where both \( \mu = E[X_i] \) and \( \sigma^2 = \operatorname{Var}(X_i) \) are finite. 
LLN provides fundamental insights into the convergence behavior of the sample mean. The~\textit{{weak law of large numbers}}~(WLLN) states that \( \overline{X}_n \) converges in probability to \( \mu \) (\( \overline{X}_n\xrightarrow{p} \mu\)) as the sample size \( n \) approaches infinity, i.e.,
\[
\lim_{n \rightarrow \infty} {P}\left( \left| \overline{X}_n - \mu \right| < \epsilon \right) = 1,~\forall~\epsilon>0.
\]
The \textit{{strong law of large number (SLLN)}} strengthens WLLN by asserting that \( \overline{X}_n \) converges to \( \mu \) almost surely:
\[
{P}\left( \lim_{n \rightarrow \infty} \overline{X}_n = \mu \right) = 1.
\]
When \( \sigma^2<\infty \), the~\textit{{central limit theorem}} (CLT) describes the asymptotic distribution of the normalized sample mean:
\[
\sqrt{n} \left(\frac{\overline{X}_n - \mu}{\sigma} \right) \xrightarrow{d} {N}( 0, 1),
\]
where \( \xrightarrow{d} \) denotes convergence in distribution, i.e.,~$\lim_{n \rightarrow \infty}{P}\left(\sqrt{n} \left(\frac{\overline{X}_n - \mu}{\sigma} \right)\le u \right)=\Phi(u)$, where \( \Phi(u) = {P}(Z \leq u) \) is the cumulative distribution function of the standard normal distribution. 
The CLT implies that for sufficiently large \( n \), the~tail probabilities of the standardized sample mean can be approximated using the standard normal distribution
$${P}\left( \sqrt{n} \left| \frac{ \overline{X}_n - \mu }{ \sigma } \right| > u \right) \approx {P}\left( |Z| > u \right) = 2 \Phi(-u), \quad Z \sim {N}(0, 1).$$
By selecting \( u = 1.96 \), we obtain the familiar 95\% confidence interval
$$\mu \in\left[\overline{X}_n\pm {1.96\sigma}/{\sqrt{n}}\right].$$
Such intervals reflect the classical efficiency criterion established by Fisher in his early work~\citep{fisher1922mathematical}, which defines efficient statistics as ``\textit{{those which, when derived from large samples, tend to a normal distribution with the least possible standard deviation}}''. 

\vspace{3pt}
\noindent \textbf{{Challenges with Asymptotic Methods}} 
\vspace{3pt}


In the era of computer-age statistical inference \citep{efron2021computer}, there has been a resurgence of interest in analyzing rigorous error bounds with high probability for desired learning procedures \citep{devroye1997probabilistic,kearns1998large}. 
These methods are particularly relevant when the sample size is small due to measurement constraints or when computational resources limit the use of large samples. Such scenarios have motivated modern statisticians to shift their focus from asymptotic analysis to non-asymptotic analysis; see~\cite{arlot2010some,horowitz2020inference,rakhlin2020mathematical,zheng2021finite,bettache2021fast,kim2021valid,yu2021finite}.

To illustrate the strengths and limitations of classical methods, consider a sequence of i.i.d. 
r.v.s $\{X_i\}_{i=1}^n$ drawn from $N(\mu, \sigma^2)$. Owing to the additive properties of the normal distribution, the~sample mean $\overline{X}_n\sim N\left(\mu, \sigma^2/n\right)$ is itself normally distributed. 
This exact distribution allows us to construct a confidence interval for $\mu$ with a precise coverage probability for any sample size $n$ is $\left[\overline{X}_n \pm {1.96\sigma}/{\sqrt{n}}\right]$.
This result is a direct consequence of the properties of the Gaussian distribution and does not rely on asymptotic~approximations.

However, the~assumption of Gaussian data is often too restrictive in real-world applications. 
 Practical data may deviate from normality due to skewness, heavy tails, or~other distributional irregularities.
 When the data are non-Gaussian, and~only the mean $\mu$ and variance $\sigma^2$ are known (or $\sigma^2$ is unknown but we have the estimator $\widehat\sigma^2\xrightarrow{p}\sigma^2$), the~CLT can still be invoked to provide an approximate confidence interval for large $n$:
\[
\left[\overline{X}_n \pm 1.96 \frac{\sigma}{\sqrt{n}}\right] \quad \text{ or } \quad \left[\overline{X}_n \pm 1.96 \frac{\widehat\sigma}{\sqrt{n}}\right].
\]
The coverage probability of this interval approaches 95\% asymptotically as $n \to \infty$. However, for~finite $n$, the~exact coverage probability is no longer guaranteed, reflecting the limitations of asymptotic methods.

This result highlights the limitations of asymptotic confidence intervals for finite sample sizes. In~practical scenarios, data distributions are often unknown, and relying solely on asymptotic approximations may lead to inaccurate coverage probabilities. Non-asymptotic methods, especially concentration inequalities, provide an alternative by offering confidence bounds that hold for any sample size and under weaker distributional assumptions.
To illustrate this approach, we revisit Hoeffding's inequality, which offers a non-asymptotic confidence interval for bounded~r.v.s.


\begin{Example}[Hoeffding's inequality works for confidence intervals]
For i.i.d. $X_{i}$'s with ${a}\le {{X_i}} \le {b}$, Hoeffding inequality gives
$$P\left( {{\mu _0} \in [{{\overline X}_n} - \frac{{b - a}}{\sqrt 2}\sqrt {\frac{1}{n}\log (\frac{2}{\delta })},{{\overline X}_n} + \frac{{b - a}}{\sqrt 2}\sqrt {\frac{1}{n}\log (\frac{2}{\delta })} ]} \right) \ge 1 - \delta .$$
 Let us examine Bernoulli samples $\{ {X_i}\} _{i = 1}^n \stackrel{\rm i.i.d.}{\sim} \operatorname{Ber}\left(1/2\right)$, with~$0\le {{X_i}} \le 1$ and $\mathrm{Var}X_i=1/4$. Put $\delta  = 0.05$, {{for any sample size $n$}}, Hoeffding's inequality gives
$$P\left( {{\mu _0} \in [{{\overline X}_n} - \frac{{1.36}}{{\sqrt n }},{{\overline X}_n} + \frac{{1.36}}{{\sqrt n }}]} \right) {\ge 95} \%,$$
which is sharp in the rate but not the constant in comparison with the normal approximated CI:
$$\mathop {\lim }\limits_{n \to \infty } P\left( {{\mu _0} \in [{{\overline X}_n} - \frac{{0.98}}{{\sqrt n }},{{\overline X}_n} + \frac{{0.98}}{{\sqrt n }}]} \right) {=95} \%.$$
\end{Example}

This example raises crucial questions about the CIs under non-asymptotic settings:

\begin{itemize}
\item \textbf{{Q1}
}. For~a finite sample, what happens if the data are non-Gaussian and unbounded?
\end{itemize}

{We} 
 expect that concentration inequalities provide valid bounds
$$ P(\mu \in[\widehat L_n, \widehat U_n])\ge 1-\delta,$$
requiring no assumptions about {{densities}} but relying instead on a few {{moment conditions}}. Addressing this question is essential because practical scenarios often involve finite samples drawn from distributions that may exhibit significant deviations from~normality.

\begin{itemize}
\item \textbf{{Q2}}. What if {$n$ is extremely small}? How to obtain practical, robust, and~tight mean bounds
$$P(\mu \in[\widehat L_n, \widehat U_n])\ge 1-\delta$$
with minimal assumptions?

\end{itemize}


These questions are particularly critical in bandit problems, where decisions must be made in real-time under limited data. Concentration inequalities, such as those for sub-Gaussian or sub-exponential distributions, offer tools to construct confidence intervals that are valid for small samples. Using results from Theorem \ref{prop:Sub-Gaussian} or Theorem \ref{sub-exponentialConcentration}, these non-asymptotic intervals enable statistical inference even in challenging small-sample settings.
These intervals enable the exploration–exploitation trade-off fundamental to bandit problems, where uncertainty quantification directly impacts algorithmic~choices.

\section{Bandit~Algorithms}\label{sec:MAB_alg}



To achieve minimal regret defined in \eqref{def_reg}, the~agent must resolve the exploration–exploitation dilemma. Specifically, the~agent must decide whether to ``exploit'' the current information by pulling the arm with the highest known average reward to maximize immediate payoff or~to ``explore'' arms with greater uncertainty, which may lead to discovering a better strategy and securing higher returns in the long run. 
\par    	
Let $v$ be a stochastic bandit and define $\Delta_k(v)=\mu^*(v)-\mu_k(v)$ as the \textit{{suboptimality gap}} of action $k$. Let $S_k(t)=\sum_{s=1}^t{I}\{A_s=k\}$ represent the number of times arm $k$ has been selected up to round $t$. The~cumulative regret can be decomposed as follows, which is useful to derive the regret upper bounds.
\begin{Lemma}[Regret decomposition lemma]
  For any policy $\pi$ and stochastic bandit $v$, we have
$$\operatorname{Reg}_T(\pi,v)=\sum\nolimits_{a\in\mathcal{A}}\Delta_a(v){E}\left[S_a(T)\right]~for~finite~or~countable~\mathcal{A},~T\in\mathbb{N}.$$
\end{Lemma}
\begin{proof}[Proof]
For any fixed $t$, we have $\sum_{a \in \mathcal{A}} {I}\left\{A_t=a\right\}=1$. Hence, the~sum of rewards is
$S_n=\sum_{t=1}^T X_t=\sum_{t=1}^T\sum_{a \in \mathcal{A}} X_t {I}\left\{A_t=a\right\}, $
and thus
\begin{align*}
\operatorname{Reg}_T(\pi,v)&=T\mu_{k^{\star}}-{E}\sum\nolimits_{t=1}^{T}X_t=\sum\nolimits_{a \in \mathcal{A}} \sum\nolimits_{t=1}^T {E}\left[{{E}\left[\left(\mu^*-X_t\right) {I}\left\{A_t=a\right\} \mid A_t\right]} \right]\\
&=\sum\nolimits_{a \in \mathcal{A}} \sum\nolimits_{t=1}^T {E}\left[{I}\left\{A_t=a\right\} \Delta_a(v) \right]=\sum\nolimits_{a \in \mathcal{A}} \Delta_a(v){E} \sum\nolimits_{t=1}^T {I}\left\{A_t=a\right\}.
\end{align*}
where the  expected reward in round $t$ conditioned on $A_t$ is $\mu_{A_t}$,
$$
\begin{aligned}
&~~{{E}\left[\left(\mu^*-X_t\right) {I}\left\{A_t=a\right\} \mid A_t\right]} ={I}\left\{A_t=a\right\} {E}\left[\mu^*-X_t \mid A_t\right] \\
& ={I}\left\{A_t=a\right\}\left(\mu^*-\mu_{A_t}\right) ={I}\left\{A_t=a\right\}\left(\mu^*-\mu_a\right)={I}\left\{A_t=a\right\} \Delta_a(v).
\end{aligned}
$$
\end{proof}
A regret upper bound is referred to as problem-independent if it depends solely on the underlying distributional assumptions, without~explicitly involving the individual gap $\Delta_a(v)$ for each action. Conversely, a~bound is termed problem-dependent if it explicitly relies on the specific values of $\{\Delta_a(v)\}_{a \in \mathcal{A}}$. In~the following sections, we will introduce several widely used algorithms designed to address this issue~effectively.

\subsection{Explore-Then-Commit~Algorithm}

As discussed earlier, the MAB problem is the most classical and simplified framework of bandit problems. It serves as a foundation for understanding the exploration–exploitation trade-off under uncertainty. Leveraging concentration inequalities, we aim to construct robust confidence bounds for decision-making in MAB settings. These bounds not only quantify uncertainty but also guide the selection of actions to balance exploration and exploitation~effectively.

To recap, in~the MAB problem, at~each time step
$t \in [T]$, the~agent selects the arm $A_t \in [K]$ and receives a reward
$\{r_k(t)\}_{t \in [T]}$ drawn from an unknown distribution ${P_k}$~(assuming the $k$-th arm is selected). 
This ${r_k}(t)$ is characterized by conditional reward ${r_{A_t}}(t)$ on the random action $A_t=k$:
\begin{equation}
\begin{aligned}
    {r_k}(t) &={r_{A_t}}(t)\mid \{A_t=k\} \quad (\text{or denoted by }~X_t \mid \{A_t=k\}),
\end{aligned}
\end{equation}
with $E[ r_{A_t}(t) \mid A_t ] = \mu_{A_t}(v)$ and $X_t=r_{A_t}(t)$ with the RL notation in Section~\ref{se:RL}.

Assuming that the optimal arm is denoted by $k^*$, from~Section~\ref{se:RL},  
the criterion for the optimal sequence of actions $\{A_t\}_{t \in [T]}$ is to minimize the cumulative regret 
defined in \eqref{def_reg}, which could be rewritten as 
\[
    \begin{aligned}
        \operatorname{Reg}_{T}(r,v) := T \mu_{k^*} - E\left[\sum_{t=1}^T X_t\right] = T \mu_{k^*} - E\left[\sum_{t=1}^T E[X_t \mid A_t]\right]= E\left[\sum_{t=1}^T (\mu_{k^*}(v) - \mu_{A_t}(v))\right].
    \end{aligned}
\]

The basic idea of the Explore-Then-Commit (ETC) algorithm is to divide the search process for the optimal arm in the MAB problems into two distinct phases: the exploration phase and the exploitation~phase.
\begin{itemize}
\item
In the exploration phase, the~algorithm pulls each arm a fixed number of times to estimate its expected reward. 
\item
In the exploitation phase, the~algorithm selects the arm with the highest estimated reward based on exploration results and continues to select it.
\end{itemize}

\noindent {Specifically}
, the~ETC algorithm is described as follows: the algorithm conducts $m$ rounds of exploration for each arm during the exploration phase. When $t \leq mK$, that is, during~the first $mK$ selections, each of the $K$ arms is pulled once per round according to a certain rule. After~$t > mK$, the~algorithm will always select the arm that performed the best during the exploration phase. Let ${\widehat \mu _k}(t)$ be the average reward for selecting arm $k$ after $t$~rounds:
\[
{\widehat \mu _k}(t) = \frac{1}{{{S_k}(t)}}\sum\limits_{s = 1}^t {I \{ {A_s} = k\}{r_k}(s)}.
\]
The pseudocode of the ETC algorithm is shown in Algorithm \ref{alg:ETC}.

\begin{algorithm}[H]
\caption{Explore-then-Commit (ETC)}\label{alg:ETC}
\begin{algorithmic}[1]
\State \textbf{Input:} Total arms $K$, number of exploration steps $m$, horizon $T>mK$.
\State In round $t$ choose arm:
$$A_t = \begin{cases} 
(t \bmod K) + 1 & \text{if } t \leq mK \\
\arg\max_{k \in [K]} \widehat{\mu}_k(mK) & \text{if } t > mK
\end{cases}$$
\end{algorithmic}
\end{algorithm}
Regarding the regret of the ETC algorithm, we have the following Theorem \ref{thm:ETC}.
\begin{Theorem}\label{thm:ETC}
When ETC is interacting with any $v:={\rm{subG}}(1)$ bandit and $1 \leq m < T/K$, the~regret of ETC satisfies
\begin{equation}
\operatorname{Reg}_{T}(r,v) \le \underbrace{m \sum\nolimits_{i=1}^K \Delta_i}_{exploration}+\underbrace{(T-m K) \sum\nolimits_{i=1}^K \Delta_i e^{-{m \Delta_i^2}/{4}} }_{\text{exploitation}}.
\end{equation}
where ${\Delta _k} = {\mu _{{k^*}}} - {\mu _k}$ represents the expected reward gap between the optimal arm and the arm ${k}$.
\end{Theorem}
\begin{proof}[Proof]
Without loss of generality, $\mu_1=\mu_{k^{\star}}=\max _i \mu_i$. Regret decomposition gives
$$\operatorname{Reg}_T(\pi,v)=\sum\nolimits_{i=1}^K\Delta_i ES_i(T).$$

\noindent \textbf{{In the first} $mK$ {rounds}}: the policy is deterministic; choose each action exactly $m$ times.

\noindent \textbf{{In remaining} $T-mK$ {rounds}}: $A_t= \operatorname{argmax}_{i \in [K]} \widehat{\mu}_i(m K)$ for $T\ge t>m K$, then
$$\begin{aligned}
ES_i(T) &  = E\sum\nolimits_{s = 1}^T {I\{ {A_s} = i\}}=\sum\nolimits_{s = 1}^{mK} {I\{ {A_s} = i\}}+(T-mK) P\left(A_{t}=i\right) \\
~& \leq m+(T-mK) P(\widehat{\mu}_i(mK) \geq \max\nolimits_{j \neq i} \widehat{\mu}_j(mK)) .
\end{aligned}$$
Since $\{A_{t}=i\}\subset\{\widehat{\mu}_i(mK) \geq \max _{j \neq i} \widehat{\mu}_j(mK)\}$. The~probability on the right-hand side
$$
\begin{aligned}
P(\widehat{\mu}_i(mK) \geq \max _{j \neq i} \widehat{\mu}_j(mK)) & \leq P\left(\widehat{\mu}_i(mK) \geq \widehat{\mu}_1(mK)\right) \\
& =P\left(\widehat{\mu}_i(mK)-\mu_i-\left(\widehat{\mu}_1(mK)-\mu_1\right) \geq \Delta_i\right) .
\end{aligned}
$$
{
Since ${S_{i}(mK)}=m$, and~$\widehat{\mu}_i(mK):=\frac{1}{S_{i}(mK)} \sum_{\tau=1}^{mK} I_{\{ A_{\tau}=i\}}X_{\tau}=\frac{1}{m} \sum_{\tau=1}^{mK} I_{\{ A_{\tau}=i\}}X_{\tau}$, it gives
$$\widehat{\mu}_i(mK)-\mu_i-\left(\widehat{\mu}_1(mK)-\mu_1\right) \sim \operatorname{subG}\left(1/ m\right)+\operatorname{subG}\left(1/ m\right)\stackrel{d}{=}\operatorname{subG}\left(2 / m\right).$$

Since $P\left(\widehat{\mu}_i(mK)-\mu_i-\widehat{\mu}_1(mK)+\mu_1 \geq \Delta_i\right) \leq e^{-{m \Delta_i^2}/{4}}$ by sub-Gaussian concentration inequality \eqref{eq:Sub-Gaussian}, it gives
$$\operatorname{Reg}_T(\pi,v)\leq \sum\nolimits_{i=1}^K\Delta_i [m+(T-mK)e^{-{m \Delta_i^2}/{4}}]\leq {m \sum\nolimits_{i=1}^K \Delta_i}+{(T-mK) \sum\nolimits_{i=1}^K \Delta_i e^{-{m \Delta_i^2}/{4}} }.$$
}
\end{proof}
For fixed $m$, $\operatorname{Reg}_T(\pi,v)$ is linear in $T$. If~$K=2$ with ${k^{\star}}=1$ and $\Delta_1=0$ and $\Delta:=\Delta_2$, so $\operatorname{Reg}_T(\pi,v) \leq m \Delta+(T-2 m) \Delta \exp \left(-{m \Delta^2}/{4}\right) \leq m \Delta+T \Delta \exp \left(-{m \Delta^2}/{4}\right).$
The regret bounds are separated into exploration and exploitation terms by first conducting sufficient trials on all arms to gather information and then using these data to make decisions that optimize long-term rewards. An~optimal $m=\max \left\{1,\left\lceil\frac{4}{\Delta^2} \log \left(\frac{T \Delta^2}{4}\right)\right\rceil\right\}$ gives
$$\operatorname{Reg}_T(\pi,v)\le \Delta+ O(\sqrt{T}),$$
see Section~6.1 of~\cite{lattimore2020bandit}. The~$O(\sqrt{T})$ is the rate of CLT for the sum of $T$ independent~r.v.s. 

In the ETC framework, the~action $A_t$ is independent of the history $H_{t-1}$ and depends only on the history observed up to the exploration phase, $H_{mK}$. In~the following subsection, we consider a scenario where $A_t$ is closely tied to the updated estimates ${\widehat{\mu}_k}(t-1)$. This approach emphasizes exploitation more effectively compared with relying solely on the estimates ${\widehat{\mu}_k}(mK)$ obtained at the conclusion of the exploration~phase.

It is worth noting that Algorithm 1 includes the pre-determined horizon $T$ as an input, which might give the impression that the algorithm is \textit{{horizon-dependent}} rather than \textit{{anytime}}. While this may seem like a limitation, the~design of \textit{{anytime}} MAB algorithms does not require substantial modifications. The~well-established \textit{{doubling trick}} is a common technique to convert horizon-dependent algorithms into anytime algorithms with comparable guarantees ({see Section~6.2 in} 
 \cite{lattimore2020bandit}). Specifically, the~\textit{{Exponential Doubling Trick}} can preserve the problem-dependent regret bounds shown in all MAB algorithms discussed in this review ({see Theorems 7 and 9 in} \cite{besson2018doubling}).

Similarly, as~emphasized at the beginning of Section~8 in~\cite{lattimore2020bandit}, we maintain our focus on algorithms with pre-determined horizons for MAB problems, as~many studies continue to prioritize algorithms designed with horizon $T$ in their inputs rather than explicitly constructing \textit{{anytime}} versions.

\subsection{Upper Confidence Bound~Algorithm}\label{sec:UCB}
The Upper Confidence Bound (UCB) algorithm is a strategy that remains optimistic under uncertainty (see~\cite{lai1987adaptive,ren2024lai_ucb}).  The~core of the algorithm lies in using the data observed so far to assign a value to each arm, called the upper confidence bound, which is a high-probability upper estimate of the unknown~mean. 

At time $t$, the~estimate of ${\mu_k(v)}$ is based on information from previous steps\linebreak $s = 1, 2, \ldots, t-1$. Using probability techniques (concentration inequalities or Gaussian approximations), a~non-asymptotic $100(1 - \alpha )\%$ confidence interval is derived:
$${\mu _k}(v) \in [{\widehat \mu _k}(t - 1) - {c_k}(t - 1),{\widehat \mu _k}(t - 1) + {c_k}(t - 1)].$$
{$c_k(t-1)$ represents the confidence bound, measuring the uncertainty in ${\widehat \mu _k}(t-1)$.}
Statistically, this means estimating the potential reward of each option using CIs and quantifying the confidence in these estimates, for~example, by~using a $95\%$ CI. Based on this, the~algorithm selects the option with the largest upper bound of CI, defined as
\begin{equation}
{A_t} = \arg {\max _{k\in [K]}}\{{\widehat \mu _k}(t - 1)+ {\widehat c_k}(t - 1)\},
\end{equation}
where ${\widehat c_k}(t - 1)$ is the estimation of half width ${c_k}(t - 1)$ of CI using the information up to round $t - 1$ from the~data.

Therefore, the selected option is the one that maximizes the sum of the current estimated reward (exploitation by evaluating the empirical mean reward of different arms) and the half-width of the confidence interval (exploration by confidentially trying out different arms). As the number of trials increases, the confidence interval gradually narrows and shrinks to its true mean, making the selection decision more reliable. The~pseudocode for the UCB algorithm is presented in Algorithm \ref{alg:UCB}.

\begin{algorithm}[H]
\caption{Upper Confidence Bound (UCB)}\label{alg:UCB}
\begin{algorithmic}[1]
\State \textbf{Input:} $K$, $T$
\For{$t = 1, \dots, T$}
    \State choose ${A_t} = \arg\max_{k \in [K]}\{{\widehat \mu _k}(t - 1)+ {\widehat c_k}(t - 1)\}.$
    \State receive reward  and update the UCB.
\EndFor
\end{algorithmic}
\end{algorithm}


Assuming the independent rewards $\{X_i\}_{i=1}^n$ follow $\mathrm{subG}(1)$ distribution with mean $\mu$, using the sub-Gaussian concentration inequality \eqref{eq:Sub-Gaussian}, we can derive
\begin{equation}\label{eq:UCB}
    P\bigg(\mu  \le \overline{X}_n  + \sqrt {\frac{{2\log (1/\delta )}}{n}} \bigg) \ge 1-\delta,~\delta \in (0, 1).
\end{equation}

At decision round $t$, let the agent have obtained ${S_k}(t - 1)$ samples from arm $k$, with~the corresponding empirical mean reward denoted by ${\widehat{\mu}_k}(t - 1)$. Utilizing the expression in~\eqref{eq:UCB}, the~UCB for the reward associated with arm $k$ is given by
\begin{equation}\label{eq:UCB1}
{\rm{UC}}{{\rm{B}}_k}(t - 1,\delta ) = \left\{ {\begin{array}{*{20}{c}}
\infty & \text{if } {S_k}(t - 1) = 0\\
{{{\widehat \mu }_k}(t - 1) + \sqrt {\frac{{2\log (1/\delta )}}{{{S_k}(t - 1)}}}} & \text{otherwise.}
\end{array}} \right.
\end{equation}
The pseudocode of the algorithm is in Algorithm \ref{alg:SubG-UCB}.

\begin{algorithm}[H]
\caption{Sub-Gaussian~UCB}\label{alg:SubG-UCB}
\begin{algorithmic}[1]
\State \textbf{Input:} $K, T, \text{ and } \delta$.
\For{$t = 1, \dots, T$}
    \State choose $A_t = \arg\max_{k \in [K]} \text{UCB}_k(t - 1, \delta).$
    \State receive reward  and update the UCB.
\EndFor
\end{algorithmic}
\end{algorithm}

Regarding the regret of the UCB algorithm, we have the following Theorem \ref{thm:subG-UCB}.
\begin{Theorem}[Theorem 7.2 in~\cite{lattimore2020bandit}]\label{thm:subG-UCB}
 For $S_k(t-1)\ge 1$, pull
$A_t = \operatorname{argmax}_{k \in [K]} \mathrm{UCB}_i(t-1, \delta)$ if $v=\mathrm{subG}(1)$.
Let $\Delta_a:=\mu_{k^{\star}}-\mu_a$ be the suboptimality gap of action $a$. Let $\delta=1/T^2$, then $\operatorname{Reg}_T(\pi,v)\leq 
3\sum_{i=1}^K \Delta_i+\sum_{k: \Delta_i>0} \frac{16 \log T}{\Delta_k}$ (problem-dependent bound) and
$$\operatorname{Reg}_T(\pi,v)\leq \underbrace{3\sum\nolimits_{k=1}^K\Delta_k}_{exploration}+\underbrace{8\sqrt{TK\log T}}_{\text{exploitation}}~(problem-independent~bound).$$
\end{Theorem}

{The detailed proof can be found in Appendix~\ref{app:proof_regret_bound1}.} However, the proof lacks intuitive insight from the machine learning, potentially making it daunting for readers. To~provide more clarity, we present an alternative approach inspired by the excess risk bound decomposition (e.g., Theorem 36.1 in~\cite{lattimore2020bandit} and Theorem 1 in~\cite{lu2021bandit}). In~what follows, we summarize a general and practical regret decomposition for the UCB algorithm, offering a more structured and comprehensible~framework.

\begin{Lemma}[Regret decomposition lemma for UCB algorithms]\label{lem:UCB}
For any UCB~algorithms such that {$A_t = \operatorname{argmax}_{k\in[K]} \mathrm{UCB}_k(t-1, \delta),$} 
we have
$$\operatorname{Reg}_T(\pi,v) \leq {E}\sum\limits_{t=1}^T[ \mu_{k^*}-\mathrm{UCB}_{k^*}(t-1, \delta)+\mathrm{UCB}_{A_t}(t-1, \delta)-\mu_{A_t}].$$
\end{Lemma}
\begin{proof}
$-\operatorname{Reg}_T(\pi,v)$ is similar to the excess risk of the empirical risk minimization in machine learning~(see Remark \ref{re:SLT} below). Similarly, we have\vspace{3pt}

$$
\begin{aligned}
&\operatorname{Reg}_T(\pi,v)={E}\sum\nolimits_{t=1}^{T}(\mu_{k^*}-\mu_{A_{t}})\\
 & ={E}[\sum_{t=1}^T \mu_{k^*}-\mathrm{UCB}_{k^*}(t-1, \delta)+{\mathrm{UCB}_{k^*}(t-1, \delta)-\mathrm{UCB}_{A_t}(t-1, \delta)}+\mathrm{UCB}_{A_t}(t-1, \delta)-\mu_{A_t}]\\
& \leq {E}\sum\nolimits_{t=1}^T[ \mu_{k^*}-\mathrm{UCB}_{k^*}(t-1, \delta)+\mathrm{UCB}_{A_t}(t-1, \delta)-\mu_{A_t}],
\end{aligned}
$$
where the inequality is due to the optimal action $\mathrm{UCB}_{k^*}(t-1, \delta)-\mathrm{UCB}_{A_t}(t-1, \delta)\le 0$.
\end{proof}

The regret decomposition in Lemma \ref{lem:UCB} shares a conceptual parallel with the excess risk decomposition in statistical learning theory. In~the context of UCB, regret is decomposed into the difference between the optimal arm and the selected arm, reflecting the trade-off between exploration and exploitation. Similarly, in~statistical learning theory, the~excess risk is broken down into generalization error, optimization error, and~concentration error, as shown in the following~example.

\begin{Example}[Decomposition in {Statistical Learning Theory}]\label{re:SLT}
We assume that $\{(X_{i}, Y_{i})\}_{i=1}^n$ represents a sequence of i.i.d. r.v.s taking values in $\mathbb{R}^d \times \mathbb{R}$, with~each pair $(X_i, Y_i)$ being an independent copy of the r.v. $(X, Y)$. Let the loss function be denoted by $l(y, x, \theta)$, where $y \in \mathbb{R}$ represents the output variable, $x \in \mathbb{R}^d$ is the input variable, and~$\theta \in \Theta$, with~$\Theta \subset \mathbb{R}^d$ being the hypothesis space. We define the expected risk as $\mathcal{R}(\theta) := E [l(Y, X, \theta)]$, and~the empirical risk as $\widehat{\mathcal{R}}(\theta):= \frac{1}{n}\sum\nolimits_{i = 1}^n {l({Y_i}, X_i, \theta )}$. Let the true parameter be $\theta^* \in \arg\inf_{\theta' \in \Theta} \mathcal{R}(\theta')$, and~the empirical risk minimizer be $\widehat{\theta} \in \arg\min_{\theta \in \Theta} \widehat{\mathcal{R}}(\theta)$. The~excess risk is decomposed as
$$
\begin{aligned}
\mathcal{R}(\widehat{\theta})-\mathcal{R}({\theta^*}) & =\{\mathcal{R}({\widehat{\theta}})-\widehat{\mathcal{R}}({\widehat{\theta}})\}
+{\{\widehat{\mathcal{R}}({\widehat{\theta}})-\widehat{\mathcal{R}}({\theta^*})\}}
+\{\widehat{\mathcal{R}}({\theta^*})- \mathcal{R}({\theta^*})\}\\
&=:\text{generalization error}+ \text{optimization error}(\le 0)+\text{concentration error}\\
& \leq \{\mathcal{R}({\widehat{\theta}})-\widehat{\mathcal{R}}({\widehat{\theta}})\}
+\{\widehat{\mathcal{R}}({\theta^*})- \mathcal{R}({\theta^*})\}\leq 2 \sup\nolimits_{\theta \in \Theta}|\widehat{\mathcal{R}}(f_\theta)-\mathcal{R}(f_\theta)|.
\end{aligned}
$$
\end{Example}

{In Appendix~\ref{app:proof_regret_bound2}, we present an alternative proof of the $O(K + \sqrt{KT\log T})$-regret bound for UCB, utilizing the method outlined in Lemma~\ref{lem:UCB} and Example~\ref{re:SLT}.
}


\subsection{The Minimax Lower Bound in Instance-Dependent MAB~Problems}

For the UCB algorithm of MAB problems, a~fundamental question arises: Is the obtained regret rate of \( O(\sqrt{K T \log T}) \) for the exploitation term in the upper bound of regret, as~stated in Theorem \ref{thm:subG-UCB}, truly optimal? 
The answer to this question has profound implications for both theory and practice in statistical learning and decision-making under uncertainty. To~answer it, we turn to the establishment of a \textit{{minimax lower bound}} on the regret, a~cornerstone concept from non-parametric statistical theory \citep{Tsybakov2009Int}.
\begin{Definition}[Minimax regret]
    Let $\mathcal{E}$ be a set of stochastic bandits and $\pi$ be a policy. The~\textit{worst-case regret }is
$\operatorname{Reg}_T(\pi, \mathcal{E})=\sup _{\nu \in \mathcal{E}} \operatorname{Reg}_T(\pi, \nu)$.
Let $\Pi$ be the set of all policies. The~minimax regret is
$$\operatorname{Reg}_T^*(\mathcal{E})=\inf \nolimits_{\pi \in \Pi} \operatorname{Reg}_T(\pi, \mathcal{E})=\inf \nolimits_{\pi \in \Pi} \sup _{\nu \in \mathcal{E}} \operatorname{Reg}_T(\pi, \nu).$$
A policy is called \textit{minimax optimal} for $\mathcal{E}$ if $\operatorname{Reg}_T\left(\pi, \mathcal{E}\right)=\operatorname{Reg}_T^*(\mathcal{E})$.
\end{Definition}

\noindent Understanding the minimax lower bound serves several critical~purposes:
\begin{itemize}
    \item \textit{{Optimality.}} Establishing a minimax lower bound allows us to rigorously demonstrate that no algorithm can achieve a better regret rate in the worst-case setting. This is crucial for confirming that the sub-Gaussian UCB algorithm is not just efficient within the class of all possible algorithms.
    \item \textit{{Informative lower bounds.}} Lower bounds often provide deeper insights than upper bounds because they highlight the intrinsic difficulty of the problem itself, independent of any specific algorithm. They serve as a benchmark for assessing the performance of existing and future algorithms.
    \item 
    \textit{{Problem complexity.}} By identifying the fundamental challenges and limitations in the problem through lower bounds, we gain valuable insight into what makes the problem hard. This understanding is essential for designing new algorithms that can effectively address these challenges and to advance the theoretical foundations of machine learning.
\end{itemize}

\subsubsection{A Lower Bound on the Minimax Regret for Sub-Gaussian~Bandits}

For specific environment classes $\mathcal{E}$, explicit minimax lower bounds provide critical insights into the limitations of any algorithm. The~minimax lower bound in Theorem \ref{thm:lower} offers a theoretical benchmark on the best possible performance achievable in the worst-case scenario for sub-Gaussian rewards.
\begin{Theorem}[Theorem 15.1 in~\cite{lattimore2020bandit}]\label{thm:lower}
Let $K>1, T \geq K-1$. Given $\mu:=(\mu_1,\cdots,\mu_K)^T\in \mathbb{R}^K$, let $\nu_\mu$ be the Gaussian bandit for which the $i$-th arm has reward ${N}\left(\mu_i, 1\right)$. For~any $\pi$, there exists a $\mu \in[0,1]^K$ such that
$$\operatorname{Reg}_T(\pi, \mathcal{E})=\sup\nolimits _{\nu \in \mathcal{E}} \operatorname{Reg}_T(\pi, \nu) \ge \operatorname{Reg}_T\left(\pi, \nu_\mu\right) \geq 27^{-1}\sqrt{(K-1)T},$$
\textls[-10]{for the sub-Gaussian environment $\mathcal{E}$. Further, $\operatorname{Reg}_T^*(\mathcal{E})=\inf _{\pi \in \Pi} \operatorname{Reg}_T(\pi, \mathcal{E}) \geq 27^{-1}\sqrt{(K-1)T}$.}
\end{Theorem}
For the UCB algorithm in Algorithm \ref{alg:SubG-UCB}, the~regret rate with factor $\sqrt{\log T}$ does not meet this minimax lower bound, indicating that there is room for improvement in its design. 
Specifically, while Algorithm \ref{alg:SubG-UCB} is widely used, it is not minimax optimal, which motivates the need for algorithms that can close this gap and achieve performance closer to the theoretical lower limit.
To address this, we explore an extended version of the UCB algorithm in the next subsection. This extension is designed to match the minimax lower bound in Theorem \ref{thm:lower}, up~to a constant factor, providing a step toward optimal algorithm design in bandit problems.

\subsubsection{Minimax Optimal Strategy in the Stochastic~Case}
MOSS (Minimax Optimal Strategy in the Stochastic case,~\cite{audibert2009minimax}) is an algorithm designed to minimize regret in the worst-case scenario, specifically tailored for MAB problems in stochastic environments. The~core idea of MOSS lies in constructing confidence intervals, where the confidence level depends on each arm's historical number of pulls, the~time horizon $T$, and~the number of arms $K$. This approach avoids over-exploration or premature exploitation, thus maintaining a balance between exploration and exploitation throughout the process to achieve minimax regret under $\operatorname{subG}(1)$ rewards.~\cite{audibert2009minimax} replaced the factor $\log (1/\delta )$ in \eqref{eq:UCB1} with $\delta=T^{-2}$ by an adaptive factor $\log^+ \left(\frac{T}{KS_k(t-1)}\right)$, and~proved that the MOSS algorithm attains the minimax lower bound in Theorem \ref{thm:lower}; see~\cite{ren2024lai_ucb} for more~details.

This advancement underscores the importance of carefully designing the exploration component in bandit algorithms. By~tuning the exploration function to more precisely reflect the uncertainty and potential of underexplored arms, MOSS effectively balances exploration and exploitation. This balance is crucial to minimize regret and achieve optimal performance over the time~horizon.

Using Doob's submartingale inequality, Theorem 9.1 in~\cite{lattimore2020bandit} obtained the following regret bound of the MOSS~algorithm. 

\begin{Theorem}\label{thm:MOSS}
For any $v=\operatorname{subG}(1)$ bandit, the~regret of the MOSS algorithm \ref{alg:MOSS}  satisfies
\begin{equation}\label{eq:MOSS}
\operatorname{Reg}_T(\pi,v) \le 39\sqrt {KT}  + \sum\limits_{k = 1}^K {{\Delta _k}}.
\end{equation}
\end{Theorem}
\noindent According to Theorem \ref{thm:lower}, Algorithm \ref{alg:MOSS} with the regret bound \eqref{eq:MOSS} achieves the minimax~\mbox{optimality}.

\begin{algorithm}[H]
\caption{{Minimax} 
 Optimal Strategy in the Stochastic case (MOSS)}
\label{alg:MOSS}
\begin{algorithmic}[1]
\State \textbf{Input:}  $K$,$T$.
\State Choose each arm once.
\State Subsequently choose:
$$A_t = \arg\max_{k \in [K]} \left( \widehat{\mu}_k(t-1) + \sqrt{\frac{4}{S_k(t-1)} \log^+ \left(\frac{T}{KS_k(t-1)}\right)} \right) $$

where ${\log ^ + }(x) = \log {\rm{ max\{ 1,x\} }}.$
\end{algorithmic}
\end{algorithm}

\subsection{Thompson Sampling~Algorithm}

Thompson Sampling (TS,~\cite{russo2018tutorial}) is an algorithm based on Bayesian inference originating from~\cite{thompson1933likelihood}. The~core idea is to continuously update the posterior distribution of each arm's reward using historical data and~to sample from this distribution to decide the next~action.

During the first $K$ steps, the~algorithm plays each arm $k \in [K]$ once and updates the estimated average reward $\widehat{\mu}_k(K+1)$ for each $k$. In~subsequent steps $t = K+1, \dots, T$, the~algorithm samples instances $\widetilde\theta_k(t)$  for all $k \in [K]$ from a certain distribution $F(\widehat{\mu}_k(t-1), \widehat{\sigma}_k^2(t-1)$ with empirical mean $\widehat{\mu}_k(t-1)$ and empirical variance $\widehat{\sigma}_k^2(t-1)$ under information at time step $t - 1$. The~algorithm then selects the arm that maximizes $\theta_k(t)$. The~average reward $\widehat{\mu}_k(t)$, reward variance $\widehat{\sigma}_k^2(t)$, and the number of pulls $S_k(t)$ for arm $k \in [K]$ are subsequently updated. 
The pseudocode of the TS algorithm is presented as Algorithm \ref{alg:TS}.
\begin{algorithm}[H]
\caption{Thompson Sampling (TS)}\label{alg:TS}
\begin{algorithmic}[1]
\State \textbf{Input:} $K$, $T$.
\State \textbf{Initialization:} Play arm once and set $S_k(K+1) = 1$; let $\widehat{\mu}_k(K+1)$ be the average reward estimation of arm $k$.
\For {$t = K+1, K+2, \dots, T$}
    \For {all $k \in [K]$}
        \State Sample $\widetilde\theta_k(t)\sim F(\widehat{\mu}_k(t-1),\widehat{\sigma}_k^2(t-1))$.
    \EndFor
    \State Choose arm $A_t =\arg\max_{k \in [K]} \theta_k(t)$ and observe the reward $r_t$.
    \For {all $k \in [K]$}
        \State $\widehat{\mu}_k(t) = \frac{S_k(t-1) \cdot \widehat{\mu}_k(t-1) + r_t I\{A_t=k\}}{S_k(t-1) + I\{A_t=k\}},~
        \widehat{\sigma}_k^2(t) = \frac{(S_k(t-1) - 1) \cdot \widehat{\sigma}_k^2(t-1) + \left(r_t - \widehat{\mu}_k(t-1)\right)^2 \cdot I\{A_t = k\}}{S_k(t-1) + I\{A_t = k\} - 1}.$
        \State $S_k(t) = S_k(t-1) + I\{A_t=k\}$.
    \EndFor
\EndFor
\end{algorithmic}
\end{algorithm}

When \(F\) is a Gaussian distribution, consider that the prior distribution of the reward for arm \(k\) is
$\mu_k \sim N(0, 1)$ with known variance. The~reward 
\(r_k(t)\) of arm \(k\) at time \(t\) 
\[
r_k(t) \sim N(\mu_k, 1).
\]

Then the posterior for arm k after time step \(t-1\) is given by the normal distribution $N(\hat{\mu}_k(t-1), \hat{\sigma}_k^2(t-1)$, where
\begin{equation}
\hat{\mu}_k(t-1) = \frac{\sum_{s=1}^{t-1} I(A_s = k) r_k(s)}{1 + \sum_{s=1}^{t-1} I(A_s = k)}, \quad
\hat{\sigma}_k^2(t-1) = \frac{1}{1+ \sum_{s=1}^{t-1} I(A_s = k)}.
\label{eq:TS-gaussian}
\end{equation}

This leads to Algorithm  \ref{alg:TS-Gaussian}.

\begin{algorithm}[H]
\caption{Gaussian~TS}\label{alg:TS-Gaussian}
\begin{algorithmic}[1]
\State \textbf{Input:} $K,T$.
\For {$t = 1, 2, \dots, T$}
    \For {$k = 1, 2, \dots, K$}
        \State Sample $\widetilde{\mu}_k(t) \sim N(\hat{\mu}_{t-1}(a), \hat{\sigma}_k^2(t-1))$ according to \eqref{eq:TS-gaussian}.
    \EndFor
    \State Let $A_t = \arg\max_{k \in [K]} \widetilde{\mu}_k(t)$.
    \State Pull arm $A_t$ and observe reward $r_k(t)$.
\EndFor
\end{algorithmic}
\end{algorithm}

Specifically, for~the case of binary rewards, it can be assumed that the prior distribution of the rewards follows a Beta distribution, while the reward distribution for each arm follows a Bernoulli distribution. We have
\[
\mu_k \sim \mathrm{Beta}(1, 1), \quad r_k(t) \sim \mathrm{Bernoulli}(\mu_k).
\]

After each pull of arm $k$, the~observed reward $\widetilde{\mu}_t(k)$ is used to update its posterior distribution. The~posterior distribution of arm $k$ at time $t - 1$ is given by
$${\rm{Beta}}(1 +  {S_k^1}(t-1),1 +{S_k^0}(t-1)),$$
where ${S_k^y}(t-1) = \sum\nolimits_{s = 1}^{t-1} I\{ {A_s} = k\} I\{r_k(s) = y\}$ denotes the number of times arm $k$ has received reward $y \in \{ 0,1\}$ by time step $t - 1$. The~pseudocode for the Beta TS algorithm is presented as Algorithm \ref{alg:betaTS}.

\begin{algorithm}[H]
\caption{Beta~TS}\label{alg:betaTS}
\begin{algorithmic}[1]
\State \textbf{Input:} $K, T$.
\For {$t = 1, 2, \dots, T$}
    \For {$k = 1, 2, \dots, K$}
        \State Sample $\widetilde{\mu}_k(t) \sim {\rm{Beta}}(1 +  {S_k^1}(t-1),1 +{S_k^0}(t-1))$.
    \EndFor
    \State Let $A_t = \arg\max_{k \in [K]} \widetilde{\mu}_k(t)$.
    \State Pull arm $A_t$ and observe reward $r_k(t) \in \{0, 1\}$.
\EndFor
\end{algorithmic}
\end{algorithm}
Regarding the regret of the TS algorithm, we have the following Theorem \ref{thm:betaTS}.

\begin{Theorem}[Theorem 36.1 in~\cite{lattimore2020bandit}]\label{thm:betaTS}
When TS algorithm is interacting with any $v=\operatorname{subG}(1)$ bandit and mean in $[0,1]$, the~regret of TS satisfies
\begin{equation}
\operatorname{Reg}_T(\pi,v) = O(\sqrt{KT \log T}).
\end{equation}
\end{Theorem}

By employing risk decomposition from statistical learning theory, the~proof of \mbox{Theorem \ref{thm:betaTS}} closely parallels that of Appendix~\ref{app:proof_regret_bound2}.

\subsection{Minimax Optimal Thompson Sampling~Algorithm}

Minimax Optimal Thompson Sampling (MOTS,~\cite{jin2021mots}) algorithm is an improvement of the classical TS algorithm by introducing a truncation mechanism for the arm reward distribution. The~core idea is that, at~each time step, the~algorithm samples from the posterior distribution of each arm but~performs a truncation on the sampling results to avoid overestimating suboptimal arms and underestimating the optimal arm. Specifically, MOTS uses truncated normal distributions to adjust the estimation of arm rewards. This mechanism effectively enhances the robustness of the algorithm when dealing with suboptimal arms and reduces the probability of selecting suboptimal arms~incorrectly.

The theoretical analysis of the MOTS algorithm shows that, within~a finite time horizon $T$, the~algorithm can achieve a minimax regret upper bound of ${\rm O}(\sqrt {KT})$, which is problem-independent. This addresses the limitation of the traditional TS algorithm, which is unable to reach this optimal regret bound. This improvement allows MOTS to demonstrate more robust performance in complex decision environments, significantly reducing the growth rate of cumulative regret. The~pseudocode of the MOTS algorithm is in Algorithm \ref{alg:MOTS}.

\begin{algorithm}[H]
\caption{Minimax Optimal Thompson Sampling (MOTS)}\label{alg:MOTS}
\begin{algorithmic}[1]
\State \textbf{Input:} $K, T$.
\State \textbf{Initialization:} Choose arm once and set $S_k(K+1) = 1$; let $\widehat{\mu}_k(K+1)$ be the observed reward of arm $k$.
\For {$t = K+1, K+2, \dots, T$}
    \For {all $k \in [K]$}
        \State Sample $\widehat\theta_k(t)\sim D_k(t-1)$.
    \EndFor
    \State Choose arm $A_t = \arg\max_{k \in [K]} \widehat\theta_k(t)$ and observe the reward $r_t$.
    \For {all $k \in [K]$}
        \State $\widehat{\mu}_k(t) = \frac{S_k(t-1) \cdot \widehat{\mu}_k(t-1) + r_t I\{A_t=k\}}{S_k(t-1) + I\{A_t=k\}}$.
        \State $S_k(t) = S_k(t-1) + I\{ A_t=k\}$.
    \EndFor
\EndFor
\end{algorithmic}
\end{algorithm}

The ${\widehat\theta_k}(t) \sim D_k(t-1)$ generated in line 5 of the algorithm satisfies
$$
{\widehat\theta _k}(t) = \min \{ {\widetilde \theta _k}(t),\;{\tau _k}(t)\},
$$
where ${\widetilde \theta _k}(t)\sim N({\widehat \mu _k}(t-1),1/(\rho {S_k}(t-1)))$, with~$\rho  \in (1/2,1)$ being a tuning parameter. 
$
{\tau _k}(t) = {\widehat \mu _k}(t-1) + \sqrt {\frac{\alpha }{{{S_k}(t-1)}}{{\log }^ + }(\frac{T}{{K{S_k}(t-1)}})} 
$
and $\alpha  > 0$ is a~constant.

In other words, ${\widehat\theta _k}(t)$ follows a truncated Gaussian distribution with the density
$$
f(x) = \left\{ 
\begin{array}{ll}
{\varphi _t}(x) + \left( {1 - {\Phi _t}(x)} \right)\delta \left( {x - {\tau _k}(t)} \right) & \text{if } x \le {\tau _k}(t) \\
0 & \text{otherwise.}
\end{array}
\right.
$$
where ${\varphi _t}(x)$ and ${\Phi_t}(x)$ respectively represent the probability density function (PDF) and cumulative density function (CDF) of $N({\widehat \mu _k}(t-1),1/(\rho {S_k}(t-1)))$, and~$\delta ( \cdot )$ is the Dirac delta~function.

Regarding the regret of the MOTS algorithm,~\cite{jin2021mots} derived Theorem \ref{thm:MOTS} that shows that the MOTS achieves minimax optimality as established by Theorem 15.2 in~\cite{lattimore2020bandit}.

\begin{Theorem}[Theorem 1 in~\cite{jin2021mots}]\label{thm:MOTS}
Assume that the reward of arm $k \in [K]$ is $\mathrm{subG}(1)$  with mean ${\mu _k}$. For~any fixed $\rho \in (1/2,1)$ and $\alpha \geq 4$, the~regret of MOTS satisfies
\begin{equation}
\operatorname{Reg}_T(\pi,v)= O\left( \sqrt{KT} + \sum_{i=2}^K \Delta_i \right).
\end{equation}
\end{Theorem}

We compared cumulative regrets for ETC ($m=210$ from Theorem \ref{thm:ETC}), UCB, MOSS, TS with a Gaussian prior, and~MOTS over 2000 rounds, averaging results across 100~simulations. The~three-arm bandit rewards followed $N(\mu_k, 1)$ with $\mu_1 = 0.5, \mu_2 = 0.6$, and \mbox{$\mu_3 = 0.8$}. UCB used a sub-Gaussian variance proxy of $1$, while TS initialized Gaussian priors as $N(0,1)$ and updated posteriors iteratively. MOTS employed $\rho=0.8$ for exploration–exploitation balance and $\alpha=1.5$ for a practical confidence bound, despite $\alpha \geq 4$ being theoretically required (Theorem \ref{thm:MOTS}). The~results, plotted over 2000 rounds, show that MOSS and MOTS achieved lower regret compared with non-minimax optimal algorithms, demonstrating their effectiveness under Gaussian~rewards.

\begin{figure}[H] 
   
   \includegraphics[width=0.9\textwidth]{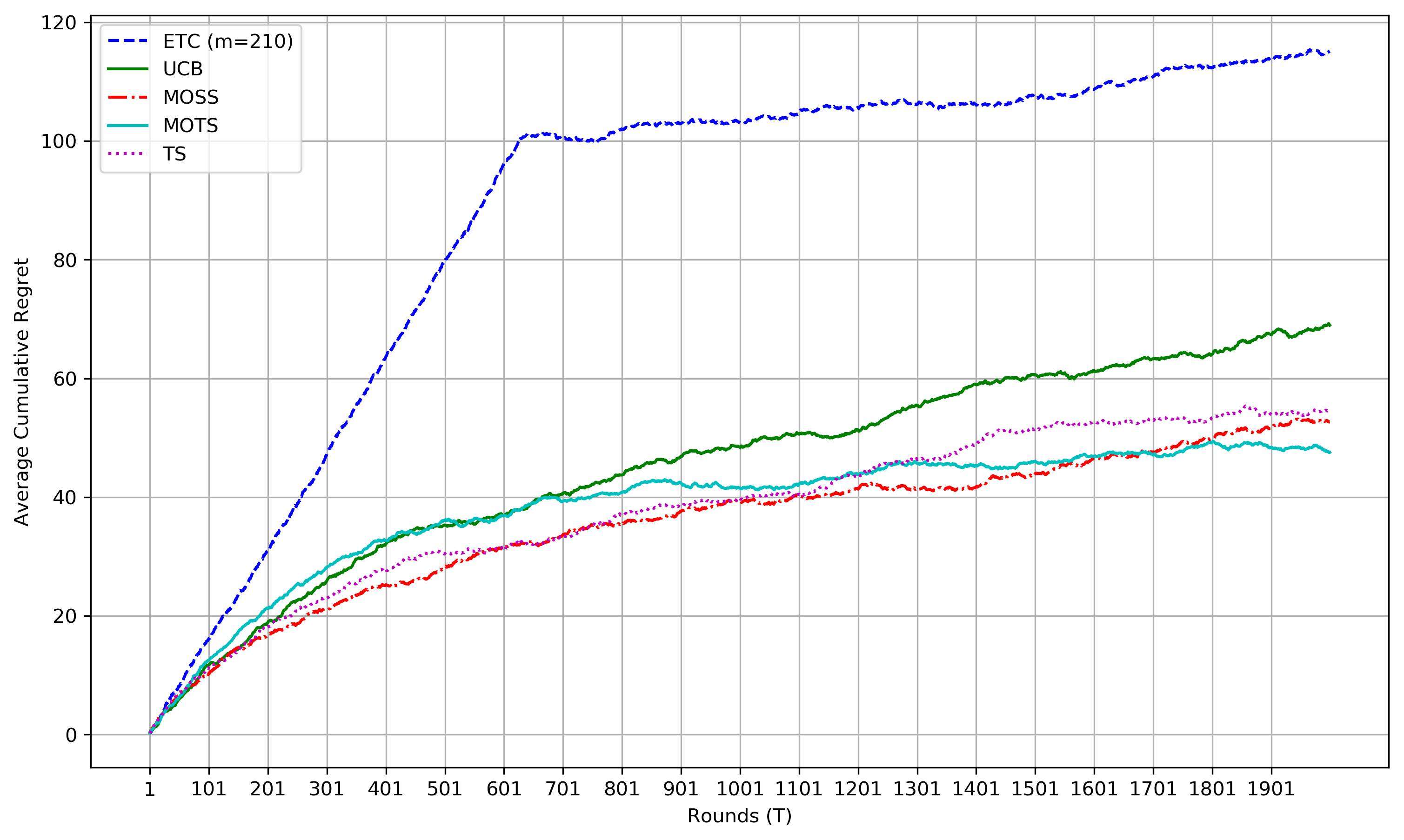} 
   \caption{{Cumulative} 
 regret comparisons of ETC, UCB, MOSS, TS, and MOTS~algorithms.}  
 \label{fig:label} 
\end{figure}

{
The advantages and disadvantages of the ETC, UCB, MOSS, TS, and~MOTS algorithms are summarized in Table~\ref{tab:algorithm_comparison} {and Figure} 
 \ref{fig:label}.
}

\vspace{-9pt}

\begin{table}[H]\small
    \caption{{{Summary} 
 of advantages and disadvantages of ETC, UCB, MOSS, TS, and~MOTS algorithms.}}
    \label{tab:algorithm_comparison} 
    \begin{tabularx}{\textwidth}{>{\centering\arraybackslash}p{1.5cm}XX}
        \toprule
        \textbf{Algorithm} & \multicolumn{1}{c}{\textbf{Advantages}} & \multicolumn{1}{c}{\textbf{Disadvantages}} \\
        \midrule
        {{ETC}
} & 
        1. Simple and easy to understand. Clear separation between exploration and exploitation, easy to implement. \newline
        2. Clear theoretical performance guarantees, suitable for simple scenarios. & 
        1. Not suitable for dynamic environments. \newline
        2. The~fixed exploration phase can lead to~inefficiency.\\
        \midrule
        {{UCB}} & 
        1. Strong theoretical performance guarantees with asymptotic convergence to the optimal solution. \newline
        2. Effectively balances exploration and exploitation by confidence intervals. & 
        1. Assumes reward independence, which may not be suitable for scenarios with dependent rewards. \newline
        2. Sensitive to the construction of confidence intervals; poor choices may result in higher~regret.\\
        \midrule
        {{MOSS}} & 
        1. Minimizes cumulative regret in the worst-case scenario, ensuring robustness in complex environments. \newline
        2. Provides theoretical guarantees for worst-case performance, making it suitable for applications requiring strong assurances. & 
        1. Can be arbitrarily worse than UCB in an infinite arms setting, despite being nearly asymptotically optimal; see \mbox{Section~9.2} in~\cite{lattimore2020bandit}. \newline
        2. Exhibits instability in regret distribution, making it less well-behaved in adaptive~settings. \\
        \midrule
        {{TS}} & 
        1. Efficiently balances exploration and exploitation by sampling from the posterior distribution of each arm's reward. \newline
        2. Asymptotically optimal with strong theoretical guarantees for the regret. & 
        1. Computationally expensive due to posterior sampling. \newline
        2. Sensitive to model assumptions; poor priors can degrade performance when the true distribution deviates from the assumed~model. \\
        \midrule
        {{MOTS}} & 
        1. Minimizes cumulative regret in the worst-case scenario, providing increased robustness in complex environments than TS. \newline
        2. The truncation mechanism prevents overestimation of suboptimal arms, reducing excessive exploration. & 
        1. More complex than standard TS due to the need for truncation mechanisms. \newline
        2. Relies on prior assumptions; improper priors can lead to instability in the algorithm's~performance. \\
        \bottomrule
    \end{tabularx}
\end{table}
\clearpage

\section{\boldmath{$K$}-Armed Contextual~Bandit}\label{sec:clb}

{
A limitation of standard MAB is that the environment remains constant for every round. In~practical applications, decision-making often relies on covariate information to improve the precision and effectiveness of decisions. For~example, in~healthcare, an~individual's treatment decision may depend on patient-specific characteristics such as genetic background, lifestyle, biomarkers, and~environmental factors. Ignoring these covariates could result in suboptimal or even incorrect treatment~plans.

Contextual information, such as in advertising recommendation systems and medical diagnosis, is crucial for making accurate decisions. Unlike traditional MAB problems that consider only mean rewards, multi-armed contextual bandits incorporate both contextual information (or features, covariates, inputs in statistics and machine learning) and the independent reward distributions of each arm. This allows the algorithm to adapt better to decision-making requirements across varying environments. In~such problems, the~reward distribution depends not only on the chosen arm but also on the current context, enabling the algorithm to respond flexibly to changing~environments.

For a vector $x \in \mathbb{R}^d$, we use $\|x\|$ to denote its $\ell_2$-norm. The~weighted $\ell_2$-norm associated with a positive-definite matrix $A$ is defined by $\|x\|_A:=\sqrt{x^{T} A x}$. 
}

\subsection{Linear Upper Confidence Bound with Disjoint Linear~Models}\label{sec:LinUCB}
A common approach to solving this is the stochastic $K$-armed contextual bandit problem~\cite{li2010contextual}. At~each time step $t$, the~algorithm receives the covariate vector $x_{t,k} \in \mathbb{R}^d$ for each arm $k$. We assume that the expected reward for the arm $k$ is a linear function of its $d$-dimensional feature vector $x_{t,k}$, with~an unknown coefficient vector $\theta_k \in \mathbb{R}^d$; specifically, for~all $t$, along with a noise term $\eta_{t,k}  \sim \mathrm{subG}(\sigma^2)$, i.e.,~disjoint linear models,
\begin{equation}\label{eq:noisy}
r_k(t) = x_{t,k}^T \theta_k + \eta_{t,k}.
\end{equation}
When \( d = 1 \) and \( x_{t,k} \) is fixed as \(1\), it reduces to the standard MAB setting since $E r_k(t) = \theta_k$.

For each arm \( k \), assuming the parameter \( \theta_k \) is bounded, the~loss function at time \( t-1\) is defined as the ridge-penalized least square:
$$\sum\limits_{s = 1}^{t-1} {{{({r_k(s)} - x_{s,k}^T\theta_k )}^2} + \lambda {{\left\| \theta_k  \right\|}^2}},$$
where $\lambda>0$ is a tuning parameter. The~estimate of the parameter $\theta_k$ is obtained through
$${\widehat \theta _k}(t-1) = \Sigma _{t-1,k}^{-1}\sum\limits_{s = 1}^{t-1} {{r_k(s)}} {x_{s,k}},~\text{where}~\Sigma _{t-1,k} = \lambda I + \sum_{s = 1}^{t-1} {{x_{s,k}}x_{s,k}^T}.$$

In each round \( t \) of the experiment, the~algorithm selects an arm $A_t \in [K]$, where the optimal arm is denoted by
\[
k_t^* = \arg \max_{k \in [K]} x_{t,k}^T \theta_k.
\]

Having observed a new context $x_{t,k}$ for arm $k$, it is suggested in~\cite{li2010contextual} that the UCB is
\begin{equation}
x_{t,k}^T{\widehat \theta _k(t-1)} + \alpha \sqrt {x_{t,k}^T\Sigma _{t-1,k}^{-1}{x_{t,k}}},
\end{equation}
where $\alpha>0$ is a parameter that controls the exploration–exploitation trade-off (or the mean-variance trade-off). 

Following a principle similar to the UCB algorithm, LinUCB selects the arm with the highest UCB. This approach enables LinUCB to effectively balance the exploitation of known rewards with the exploration of new information, progressively improving decision-making accuracy. The~pseudocode for the LinUCB algorithm is in Algorithm \ref{alg:LinUCB}.

\begin{algorithm}[H]
\caption{LinUCB with disjoint linear 
 models}\label{alg:LinUCB}
\begin{algorithmic}[1]
\State \textbf{Inputs:} $K$, $T$, $\lambda$, $\alpha \in \mathbb{R}^+$
    \For{all $k \in [K]$}
        \State $\Sigma_{0,k} = \lambda I$ ($d$-dimensional identity matrix).
        \State $b_{0,k} = 0$ ($d$-dimensional zero vector).
        \State $\widehat{\theta}_k(0) = 0$ ($d$-dimensional zero vector).
    \EndFor
\For{$t \in [T]$}
    \State Observe features of all arms $k \in [K]$: $x_{t,k} \in \mathbb{R}^d$.
    \State Choose arm 
    $$A_t = \arg\max_{k \in [K]} \left(x_{t,k}^T{\widehat \theta _k(t-1)} + \alpha \sqrt {x_{t,k}^T\Sigma _{t-1,k}^{-1}{x_{t,k}}}\right),$$
    with ties broken arbitrarily, and~observe a real-valued reward $r_{A_t}(t)$.
    \State $\Sigma_{t,k} = \Sigma_{t-1,k} + x_{t,A_t} x_{t,A_t}^T.$
    \State $b_{t,k} = b_{t-1,k} + r_{A_t}(t) x_{t,A_t}.$
    \State $\widehat{\theta}_k(t) = \Sigma_{t,k}^{-1} b_{t,k}.$
\EndFor
\end{algorithmic}
\end{algorithm}

\subsection{Linear Upper Confidence Bound with Common Linear~Models}

Below, we introduce the general LinUCB algorithm for linear models with a common regression parameter vector. At~each time step $t$, the~MAB receives $K$ feature vectors ${x_{t,1}}, {x_{t,2}}, \cdots, {x_{t,K}} \in \mathcal{X}_t
$, where $\mathcal{X}_t \subseteq \mathbb{R}^d$. The~reward obtained by each arm is assumed to be a linear function of its respective feature vector, where the parameter vector $\theta$ is fixed and identical for all arms. For~fixed $k$, assume that the noises $\{\eta _{t,k}\}\sim \mathrm{subG}(\sigma^2)$ are independent; we have
\begin{equation}\label{eq:reward_}
{r_k}(t) = x_{t,k}^T\theta  + {\eta _{t,k}},
\end{equation}

At each round $t$, the~algorithm selects an arm $A_t \in [K]$, where the optimal arm is denoted as
\[
k_t^* = \arg \max_{k \in [K]} x_{t,k}^T \theta.
\]
The \textit{{suboptimality gap}} of the chosen arm $A_t$ at time $t$ is then given by 
$$\Delta_t = x_{t,k_t^*}^T \theta - x_{t,A_t}^T \theta.$$
The agent's goal is to minimize the cumulative regret over the time horizon $T$
\begin{equation}
\operatorname{Reg}_{T}(\pi,\eta) = \sum_{t=1}^T \Delta_t = \sum_{t=1}^T \langle x_{t,k_t^*} - x_{t,A_t}, \theta \rangle.
\end{equation}

By the property of the estimator in ridge regression, one has the following result on the confidence set of $\widehat{\theta}$, and the confidence radius is determined in the next lemma.
\begin{Lemma}[Lemma 17.8 in~\cite{zhang2023mathematical}]\label{thm:bound}
Let $u$ be the new observation vector (context information), and~assume there exists a constant $B$ such that $\|{\theta} \| \le B$. Furthermore, let $\{ \beta_t \}$ be a sequence so~that
$$P \left( \forall 0 \leq t \leq T : \beta_t \geq \sqrt{\lambda B} + \left\| \sum_{s=1}^t \eta_{s,k} x_{s,k} \right\|_{\Sigma_t^{-1}} \right) \geq 1 - \delta.$$
Then with probability at least $1 - \delta$, for~all $t=0 , \dots , T$ we have
\begin{equation}
\left| {{u^T}(\widehat{\theta}_t - \theta )} \right| \le {\beta _t}\sqrt {{u^T}\Sigma _t^{ - 1}u}.
\end{equation}
\end{Lemma}
\begin{Lemma}[Example 17.9 in~\cite{zhang2023mathematical}]\label{thm:betazhi}
Assume $d$ is finite-dimensional with $B' = \sup_{t,k} \|x_{t,k}\|$ and noise terms $\eta_{t,k}  \sim \mathrm{subG}(\sigma^2)$. Then, in Lemma \ref{thm:bound}, we can set
\begin{equation}
\beta_t = \sqrt{\lambda B} + \sigma \sqrt{2 \log(1/\delta) + d \log\left(1 + \frac{T(B')^2}{d\lambda}\right)}.
\end{equation}
\end{Lemma}

Similarly to Algorithm \ref{alg:LinUCB}, a~pseudocode is given for the general LinUCB Algorithm \ref{alg:genlinearUCB}.
\begin{algorithm}[H]
\caption{General Linear UCB~Algorithm}
\label{alg:genlinearUCB}
\begin{algorithmic}[1]
\State \textbf{Input:} $\lambda$, $K$, $T$, $\{\beta_t\}$.
\State $\Sigma_0 = \lambda I$($d$-dimensional identity matrix).
\State $\widehat{\theta}_0 = 0$ ($d$-dimensional zero vector).
\State $b_0 = 0$($d$-dimensional zero vector).
\For{$t = 1, 2, \dots, T$}
    \State Observe  ${x_{t,1}}, {x_{t,2}}, \cdots, {x_{t,K}}$.
    \State  Choose arm $A_t = \arg\max_{k \in [K]} \left( \widehat{\theta}_{t-1}^T x_{t,k} + \beta_{t-1} \sqrt{x_{t,k}^T \Sigma_{t-1}^{-1} x_{t,k}} \right)$.
    \State Observe reward $r_{A_t}(t)$.
    \State  $b_t = b_{t-1} + r_{A_t}(t) x_{t,A_t}$.
    \State  $\Sigma_t = \Sigma_{t-1} + x_{t,A_t} x_{t,A_t}^T$.
    \State  $\widehat{\theta}_t = \Sigma_t^{-1} b_t$.
\EndFor
\end{algorithmic}
\end{algorithm}

\begin{Theorem}[Example 17.12 in~\cite{zhang2023mathematical}]\label{thm:liner_UCB}
Assume ${r_k}(t) \in [0,1]$ and $\{ \beta_t \}$ satisfies the conditions of Lemma \ref{thm:betazhi} with $\sigma=0.5$ 
(the sub-Gaussian parameter for a $[0,1]$-valued r.v. is $0.5$),
then the regret of LinUCB satisfies
\begin{equation}
\operatorname{Reg}_T(\pi,v) = \widetilde O(d\sqrt T + \sqrt {\lambda dT} B ),
\end{equation}
where $\widetilde O$ hides logarithmic factors.
\end{Theorem}

Theorem \ref{thm:liner_UCB} shows an $\widetilde O(d \sqrt{T})$ regret bound that is independent of the number of arm $K$. This rate matches the minimax lower bound up to a logarithm factor for the contextual bandit problems of infinite actions~\cite{dani2008stochastic}.

\subsection{Thompson Sampling for Linear
Contextual~Bandits}
Using the notation defined above, suppose the rewards satisfy the condition in \eqref{eq:reward_}. Following the idea of the TS algorithm, we design the algorithm using a Gaussian likelihood function and a Gaussian prior. More precisely, suppose that at time \(t\), given feature vectors \(x_{t,k}\) and parameter \(\theta\), the~reward \(r_k(t)\) satisfies
\[
r_k(t)\sim N\left( x_{t,k}^T \theta, v^2 \right),
\]
where \(v\) is a constant used to parametrize the algorithm. Then, the~posterior distribution of the parameter \(\theta\) at time \(t\) follows:
\[
N(\widehat{\theta }_{t-1}, v^2 \Sigma_{t-1}^{-1}).
\]
At each time step \(t\), a~sample \(\widetilde{\theta}_{t-1}\) is simply drawn from this distribution, and~the arm is selected to maximize \(x_{t,k}^T \widetilde{\theta }_{t-1}\).

The pseudocode of the TS for Linear Contextual Bandits (LinTS) algorithm is as Algorithm \ref{alg:LinTS}.

\begin{algorithm}[H]
\caption{LinTS}
\label{alg:LinTS}
\begin{algorithmic}[1]
\State \textbf{Input:} $K$, $T$, $v$.
\State $\Sigma_0 = I$(d-dimensional identity matrix).
\State $\widehat{\theta}_0 = 0$ (d-dimensional zero vector).
\State $b_0 = 0$(d-dimensional zero vector).
\For{$t = 1, 2, \dots, T$}
    \State Observe  ${x_{t,1}}, {x_{t,2}}, \cdots, {x_{t,K}}$.
    \State Sample \(\widetilde{\theta}_{t-1}\) from  distribution $N(\widehat{\theta}_{t-1}, v^2 \Sigma_{t-1}^{-1})$.
    \State  Choose arm $A_t = \arg\max_{k \in [K]}  \widetilde{\theta}_{t-1}^T x_{t,k} $.
    \State Observe reward $r_{A_t}(t)$.
    \State  $b_t = b_{t-1} + r_{A_t}(t) x_{t,A_t}$.
    \State  $\Sigma_t = \Sigma_{t-1} + x_{t,A_t} x_{t,A_t}^T$.
    \State  $\widehat{\theta}_t = \Sigma_t^{-1} b_t$.
\EndFor
\end{algorithmic}
\end{algorithm}

\begin{Theorem}[Theorem 1 in~\cite{agrawal2014thompsonsamplingcontextualbandits}]\label{thm:linTS}
Assume that \(\|x_{t,k}\| < 1\), \(\|\theta\| < 1\), and~\(\Delta_t  < 1\). For~the stochastic contextual bandit problem with linear payoff functions, with~probability $1 - \delta$, the~total regret in time $T$ for LinTS (Algorithm \ref{alg:LinTS}) is bounded by
\[
    \operatorname{Reg}_T(\pi,v)=O\left( d^{3/2} \sqrt{T} \left( \log(T) + \sqrt{\log(T) \log\left(\frac{1}{\delta}\right)} \right) \right),
\]
for any $0 < \delta < 1$, where $\delta$ is a parameter used by the LinTS algorithm.
\end{Theorem}
It is worthy to note that the regret bound in Theorems \ref{thm:liner_UCB} and \ref{thm:linTS} does not depend on $K$ and~is applicable to the case of infinite~arms.

Next, we introduce the \textit{general} contextual bandit TS algorithm to finish this part. The~set of observations 
${\mathcal D}_{t} := \bigcup\nolimits_{i=1}^{t} (x_{i,A_i}, r_{A_i}(i))$
are modeled using a parametric likelihood function 
\( P(r |\theta; x_A) \) depending on some parameters \( \theta \). Given the prior distribution \( P(\theta) \), the~posterior distribution is given by the 
Bayes rule:
$$
P(\widehat{\theta}_t \mid {\mathcal D}_{t} ) \propto \prod P(r_{A_t}(t)  \mid  x_{t,A_t},\widehat{\theta}_{t-1}) P(\theta).
$$

In general, the~expected reward is a non-linear function of the action, context, and the unknown 
true parameter \( \theta \). Ideally, we aim to maximize the expected 
reward:
$$
E \big[ r_k(t) \mid x_{t,k}, \widetilde{\theta}_{t-1} \big] = g(x_{t,k}, \widetilde{\theta}_{t-1}), \quad \text{($g$ is known or unknown)}
$$
where $\widetilde{\theta}_{t-1}$ is drawn from the posterior distribution $P(\widehat{\theta}_{t-1} \mid \mathcal  D_{t-1})$. When $g(x_{t,k}, \widetilde{\theta}_{t-1}) = \mu(x_{t,k}^T \widetilde{\theta}_{t-1})$ with a known mean function $\mu(\cdot)$, this defines generalized linear contextual bandits~\cite{russo2014learning,li2017provably}. More generally, $g$ can be a deep neural network, as~in Neural UCB~\cite{zhou2020neural} and Neural TS~\cite{zhang2021neural}. The~pseudocode for the contextual bandit TS algorithm is in Algorithm \ref{alg:GenLinTS}.

\begin{figure}[H] 
   
   \includegraphics[width=0.7\textwidth]{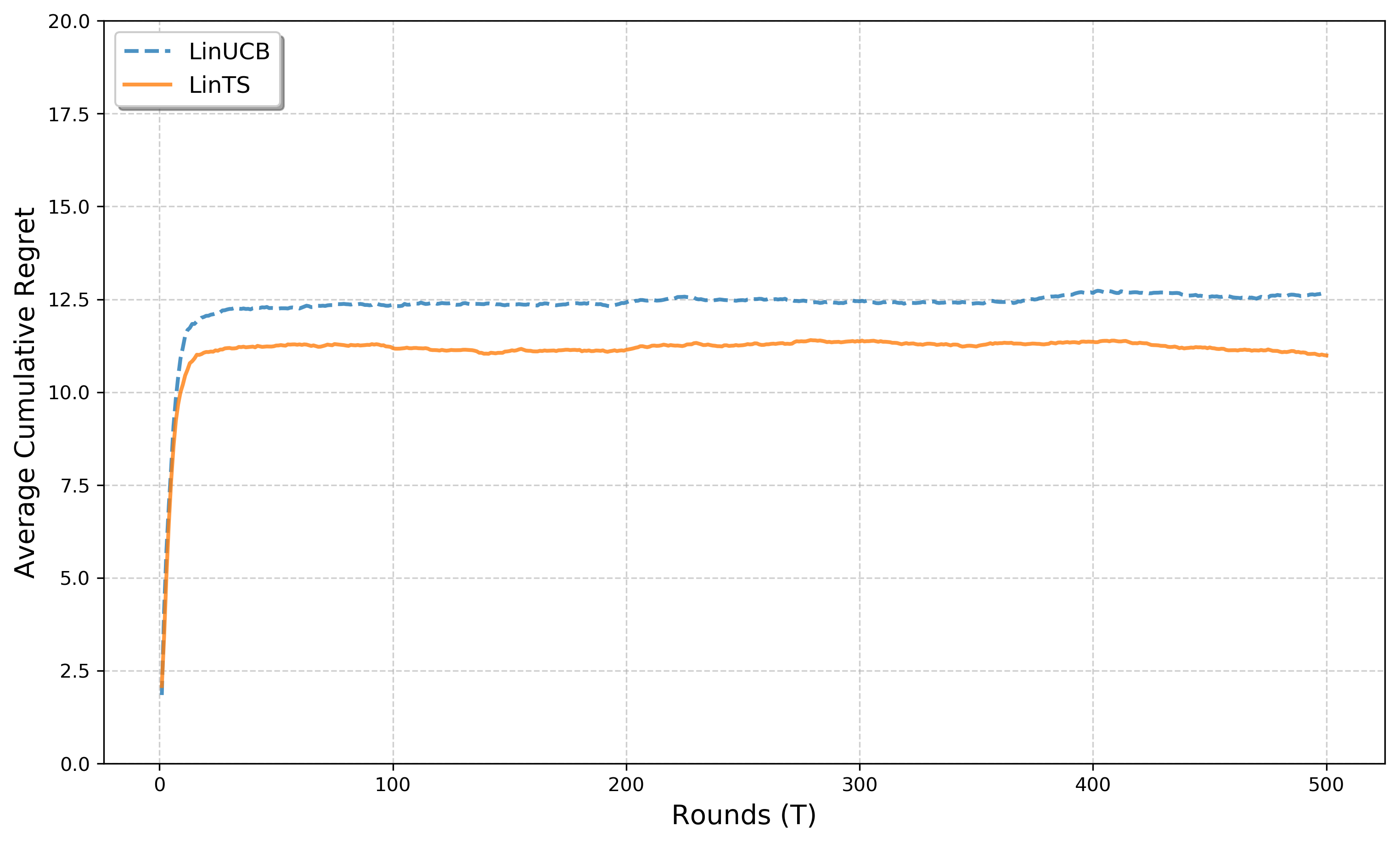} 
   \caption{Cumulative regret comparison of LinUCB and LinTS~algorithms}  
 \label{fig:Linsumulate} 
\end{figure}

Finally, we evaluate the performance of the LinUCB and LinTS algorithms through a simulation experiment. 
In the experiment, we set the number of arms to \( K = 5 \). The~feature vectors are drawn from \( N(0, 1) \), while the true parameter vector \( \theta \) is sampled from $U(0, 1)$ , with~a dimensionality of \( d = 10 \).
The true reward $r_k(t)$ at each round is assumed to have a linear relationship  with the corresponding feature vector as \eqref{eq:reward_}, 
incorporating additive noise sampled from $N(0,0.1)$ .
In Figure~\ref{fig:Linsumulate}, the~simulation results confirm that both the LinUCB and LinTS algorithms can efficiently capture the relationship between rewards and feature vectors, achieving  convergence~quickly.

\begin{algorithm}[H]
\caption{{Contextual} 
 bandit TS~algorithm}
\label{alg:GenLinTS}
\begin{algorithmic}[1]
\State \textbf{Input:} $K$, $T$.
\State $\mathcal D_0 = \emptyset$.
\For{$t = 1, \ldots, T$}
    \State Observe  ${x_{t,1}}, {x_{t,2}}, \cdots, {x_{t,K}}$.
    \State Sample \(\widetilde{\theta}_{t-1}\) from   $P(\widehat{\theta}_{t-1} \mid \mathcal D_{t-1})$.
    \State Choose arm $A_t = \arg\max_{k \in [K]} g(x_{t,k}, \widetilde{\theta}_{t-1})$.
    \State Observe reward $r_{A_t}(t)$.
    \State $\mathcal  D_t = \mathcal  D_{t-1} \cup (x_{t,A_t}, r_{A_t}(t))$.
    \State Update $P(\widehat{\theta}_t\mid \mathcal   D_t)$ by Bayes rule.
    
\EndFor
\end{algorithmic}
\end{algorithm}

\vspace{-6pt}

{
The advantages and disadvantages of the LinUCB and LinTS algorithms are summarized in Table~\ref{tab:linucb_lin_ts_comparison}.
}

\begin{table}[H]\small
    
    \caption{{Summary of advantages and disadvantages of LinUCB and LinTS algorithms.}}
    \label{tab:linucb_lin_ts_comparison}
    \begin{tabularx}{\textwidth}{>{\centering\arraybackslash}p{1.5cm}XX}
        \toprule
        \textbf{Algorithm} & \multicolumn{1}{c}{\textbf{Advantages}} & \multicolumn{1}{c}{\textbf{Disadvantages}} \\
        \midrule
        {{LinUCB}
} & 
        1. Simple and computationally efficient with a closed-form solution for UCBs in linear models. \newline
        2. Provides strong theoretical guarantees for sub-linear regret in static environments with linear reward models. & 
        1. The~linearity of the reward model can result in suboptimal performance in non-linear environments. \newline
        2. Exploration strategy based on upper confidence bounds may not fully capture the exploration–exploitation trade-off in complex~scenarios. \\
        \midrule
        {{LinTS}} & 
        1. With~uncertainty by Bayesian sampling, leading to a more balanced exploration–exploitation trade-off. \newline
        2. Performs well in non-stationary environments due to quick adaptation to reward changes. & 
        1. More computationally intensive than LinUCB due to the need for sampling from posterior distributions. \newline
        2. Performance can degrade if the posterior distribution is inaccurate or poorly~initialized. \\
        \bottomrule
    \end{tabularx}
\end{table}
\unskip

\section{Stochastic Continuum-Armed Bandits~Algorithms}

A stochastic continuum-armed bandit policy \( \pi= \{ \mathcal{A}_1, \ldots, \mathcal{A}_T \} \), \eqref{eq:mp} is defined as a sequence of possibly randomized maps:
$$
\mathcal{A}_t: D^{t-1} \times \mathbb{R}^{t-1} \rightarrow D, \quad t = 2, \ldots, T,
$$
with initial action \( \mathcal{A}_1 \in D \), which is maybe a random~number.

The algorithm generates a sequence of arms \( \{ x_1, \ldots, x_T \} \in D^T \) and corresponding observations \( \{ y_1(x_1), \ldots, y_T(x_T) \} \in \mathbb{R}^T \), where \( x_1 = \mathcal{A}_1 \) and 
$$
x_t = \mathcal{A}_t\left(x_1, \ldots, x_{t-1}, y_1(x_1), \ldots, y_{t-1}(x_{t-1})\right).
$$
The goal of the decision maker is to minimize the cumulative regret $\operatorname{Reg}_T(\pi;\mathcal{F},v)$ in Section~\ref{se:SCB}  over the time horizon \( T \). 
\subsection{Gaussian Process-Upper Confidence Bound~Algorithm}

A state-of-the-art setting for continuum-armed bandit algorithms was first introduced by~\cite{srinivas2009gaussian}, where they proposed the GP-UCB algorithm. The~reward function \( f \) is modeled as being sampled from a Gaussian process~\cite{williams1995gaussian},
$$
f(x) \sim \mathcal{GP}\left(\mu({x}), k\left({x}, {x}^{\prime}\right)\right), \quad x, x' \in D ,
$$
which is  a collection of dependent r.v.s (an extension of the multivariate Gaussian distribution to an infinite-dimensional Gaussian distribution 
$\{f(x)\}_{x \in D}$
). Here, the~mean function \( \mu(x) = {E}[f(x)] \) and the covariance function \( k(x, x') \) are defined as
$$
k\left (x, {x}^{\prime}\right) = {E}[(f(x) - \mu({x}))(f(x^{\prime}) - \mu(x^{\prime}))].
$$

Since the infinite-dimensional parameter $f$ is random and unknown, the~resulting bandit algorithm is a Bayesian optimization~\cite{garnett2023bayesian}. The~$k(x, {x}^{\prime})$, also known as a kernel function, is a positive semidefinite function. It generalizes the concept of a positive semidefinite matrix to an infinite-dimensional space and encodes the dependencies between function values at different points $x$ and $x^{\prime}$.

\begin{Definition}[Positive semidefinite kernel function]
A function $k: \mathcal{X} \times \mathcal{X} \rightarrow \mathbb{R}$ is positive semidefinite (PSD) if it is symmetric and for all $n \in \mathbb{N}$ and all $x_{1}, \ldots, x_{n} \in \mathcal{X},$ {the} 
$$
\mathbf K=\left[\begin{array}{ccc}
k\left(x_{1}, x_{1}\right) & \cdots & k\left(x_{1}, x_{n}\right) \\
\vdots & \ddots & \vdots \\
k\left(x_{n}, x_{1}\right) & \cdots & k\left(x_{n}, x_{n}\right)
\end{array}\right]
$$
is positive semidefinite, i.e.,~for all $\alpha \in \mathbb{R}^{n}$:~~$\alpha^{T} \mathbf K \alpha \geq 0$.
\end{Definition}

Common choices of covariance functions~include the following:
\begin{itemize}
    \item Finite-dimensional linear kernel: 
    $
    k\left({x}, {x}^{\prime}\right) = {x}^T {x}^{\prime}.
    $
    
    \item Squared exponential kernel: 
    $
    k\left({x}, {x}^{\prime}\right) = \exp\left\{-\frac{1}{2 l^2}\left\|{x}-{x}^{\prime}\right\|^2\right\}
    $, where $l>0$
    
    \item Matern kernel: 
    $
    k\left({x}, {x}^{\prime}\right) = \left(\frac{2^{1-\nu}}{\Gamma(\nu)}\right) r^\nu B_\nu(r),~\text{where}~r = \frac{\sqrt{2 \nu}}{l}\left\|{x}-{x}^{\prime}\right\|.
    $, which represents the similarity of ${x},{x}^{\prime}\in D$; $B_\nu(r)$ is a modified Bessel function.
\end{itemize}

Gaussian processes allow for smoothness assumptions about the reward function \( f \) via the kernel choice in a flexible, non-parametric manner. Given noisy observations \( {\mathbf Y}_t = (y_1, \ldots, y_t)^T \) at fixed points \( {\mathbf x}_t =(x_1, \ldots, x_t)^T \), we can express the \textit{Gaussian processes regression} as
\begin{equation}\label{eq:GPMODEL}
    y_t = f(x_t) + \epsilon_t, \quad \epsilon_t \overset{\text{i.i.d.}}{\sim} N(0, \sigma^2),
\end{equation}
for \(i = 1, \ldots, t-1\). Here, \(\sigma^2\) is the variance of the observation~noise. 
   
We consider a Gaussian process prior over the function \( f \). By~definition, a~GP prior states that for any finite collection of input points, the~function values are jointly Gaussian distributed. Specifically, for~a set of points \( \{x_1, \ldots, x_{t-1}, x_t\} \), the~function values \(\mathbf{f}_{1:t-1} = (f(x_1), \ldots, f(x_{t-1}))^T\) and \(f_t = f(x_t)\) follow a joint Gaussian distribution:
\begin{equation}\label{eq:JOINT}
  \begin{bmatrix}
   \mathbf{f}_{1:t-1} \\
   f_t
   \end{bmatrix} \sim {N}\left(\mathbf{0}, 
   \begin{bmatrix}
   \mathbf{K}_{t-1} & \mathbf{k}_{t-1}(x_t) \\
   \mathbf{k}_{t-1}(x_t)^T & k(x_t, x_t)
   \end{bmatrix}\right),   
\end{equation}
where \(\mathbf{K}_{t-1}\) is the \((t-1) \times (t-1)\) kernel matrix with \([\mathbf{K}_{t-1}]_{ij} = k(x_i, x_j)\) for \(i,j=1,\ldots,t-1\), and~\(\mathbf{k}_{t-1}(x) = [k(x_1, x), \ldots, k(x_{t-1}, x)]^T\). Next, we detailedly derive the GP-UCB~algorithm.

{\textbf{{Likelihood model}}: Assume the observations in \(\mathbf{y}_{t-1}\) are generated from the model~\eqref{eq:GPMODEL}. Thus, the~likelihood of observing \(\mathbf{y}_{t-1}\) given the function values \(\mathbf{f}_{1:t-1}\) is
   $$
   P(\mathbf{Y}_{t-1} \mid \mathbf{f}_{1:t-1}) = {N}(\mathbf{f}_{1:t-1}, \sigma^2 \mathbf{I}).
   $$

\textbf{{Bayesian updating}}: By Bayes’ rule, the~posterior distribution over $f_t = f(x_t)$ after observing \(\mathcal{H}_{t-1} := \{(x_1, y_1), \ldots, (x_{t-1}, y_{t-1})\}\) is
$$P\left(f_t \mid \mathbf{Y}_{t-1};{\mathbf x}_{t-1}\right) \propto P\left(\mathbf{Y}_{t-1};{\mathbf x}_{t-1} \mid f_t\right) P(f_t)$$
where ${\mathbf x}_{t-1}$ is fixed and $\mathbf{Y}_{t-1}$ is random. Since both $P\left(f_t \mid \mathbf{Y}_{t-1};{\mathbf x}_{t-1}\right)$ and \(P\left(f_t\right)\) are Gaussian, the~posterior is also Gaussian:
$$f_t(\cdot) \mid \mathbf{Y}_{t-1} \sim N(\mu_t(\cdot), \sigma_t^2(\cdot)), $$
   where the posterior mean function $\mu_t(\cdot)$ and covariance function $\sigma_t^2(\cdot)$ can be derived by standard Gaussian conditioning~rules.

\textbf{{Predictive distribution}}: We now want the distribution over \(f_t\) given the data \(\mathcal{H}_{t-1}\) and the new input \(x_t\). Given the observed data \(\mathcal{H}_{t-1}\), the~posterior predictive distribution for \(f_t\) is obtained by conditioning this joint Gaussian on the noisy values observed \(\mathbf{y}_{t-1}\). 
Thus, to~obtain \(f_t \mid (\mathcal{H}_{t-1}, x_t)\), we consider the joint distribution \citep{brochu2010tutorial,schulz2018tutorial}
   $
   \begin{bmatrix}
   \mathbf{y}_{t-1} \\
   f_t
   \end{bmatrix}
   $
   and use the conditioning property of the joint Gaussian~variables.

Applying the conditioning formula: For Gaussian random variables, if~we partition
   $$
   \begin{bmatrix} \mathbf{a} \\ \mathbf{b} \end{bmatrix} \sim {N}\left( \begin{bmatrix} \boldsymbol{\mu}_a \\ \boldsymbol{\mu}_b \end{bmatrix}, \begin{bmatrix} \mathbf{A} & \mathbf{C} \\ \mathbf{C}^T & \mathbf{B} \end{bmatrix}\right),
   $$
by the Sherman–Morrison–Woodbury formula, the~conditional distribution \(\mathbf{b} \mid \mathbf{a}\) is
\begin{equation}\label{eq:conditional}
     \mathbf{b} \mid \mathbf{a} \sim {N}(\boldsymbol{\mu}_b + \mathbf{C}^T\mathbf{A}^{-1}(\mathbf{a}-\boldsymbol{\mu}_a), \; \mathbf{B}-\mathbf{C}^T\mathbf{A}^{-1}\mathbf{C}).
\end{equation}
   We apply this to \(\mathbf{a} = \mathbf{y}_{t-1}\)
   and \(\mathbf{b} = f_t\). Before~conditioning on \(\mathbf{y}_{t-1}\), we must rewrite the joint distribution in terms of \(\mathbf{y}_{t-1}\) rather than \(\mathbf{f}_{1:t-1}\). From~the GP prior and the noise model:
   $$
   \mathbf{y}_{t-1} \sim {N}(0, \mathbf{K}_{t-1} + \sigma^2\mathbf{I}),
   $$
and so
   $\mathrm{cov}(f_t, \mathbf{y}_{t-1}) = \mathrm{cov}(f_t, \mathbf{f}_{1:t-1}) = \mathbf{k}_{t-1}(x_t)^T$ by \eqref{eq:JOINT} and the property of the Gaussian process. Thus, the~joint distribution of \(\mathbf{y}_{t-1}\) and \(f_t\) is
   $$
   \begin{bmatrix}
   \mathbf{y}_{t-1} \\[6pt]
   f_t
   \end{bmatrix}
   \sim {N}\left(
   \begin{bmatrix}
   0 \\[6pt] 
   0
   \end{bmatrix},
   \begin{bmatrix}
   \mathbf{K}_{t-1} + \sigma^2 \mathbf{I} & \mathbf{k}_{t-1}(x_t) \\[6pt]
   \mathbf{k}_{t-1}(x_t)^T & k(x_t, x_t)
   \end{bmatrix}
   \right).
   $$
   
   Now we apply the conditional formula \eqref{eq:conditional} to obtain 
$$f_t \mid (\mathbf{Y}_{t-1}, \mathbf{x}_t) \sim N(\mu_t(x_t), \sigma_t^2(x_t)), $$
where the posterior mean is 
$\mu_t(x) := \mathbf{k}_{t-1}(x)^T({\bf{K}}_{t-1} + \sigma^2 \mathbf{I})^{-1} {\mathbf y}_{t-1}$ and the posterior variance is
$
\sigma_t^2(x) :=k_{t-1}(x, x) := k(x, x) - \mathbf{k}_{t-1}(x)^T({\bf{K}}_{t-1} + \sigma^2 \mathbf{I})^{-1} \mathbf{k}_{t-1}(x)$, and~the posterior distribution over $f$ is
$$f_{t}(\cdot)\sim \mathcal{GP}(\mu_i(\cdot), k_{t-1}(\cdot,\cdot)),$$
with $k_t(x, x') = k(x, x') -\mathbf{k}_{t-1}(x)^T({\bf{K}}_{t-1} + \sigma^2 \mathbf{I})^{-1} \mathbf{k}_{t-1} (x')$ according to the covariance function in \eqref{eq:JOINT}.

At time \( t \), suppose that we have already evaluated $x^{\star}$ in \eqref{eq:mp} at the points \(\mathbf{x}_t\) and obtained \(\mathbf{Y}_{t-1}\). An~exploitation algorithm selects the next domain point
$$
{\tilde x_t}= \underset{x \in D}{\operatorname{argmax}} \, \mu_{t-1}(x),
$$
by maximizing the posterior mean. However, this approach is too greedy and may get stuck in local optima. To~address this, if~\( \beta_t \) are appropriate constants, the~GP-UCB chooses
$$
x_t = \underset{x \in D}{\operatorname{argmax}} \left\{ \mu_{t-1}(x) + \beta_t^{1/2} \sigma_{t-1}(x) \right\},
$$
which implicitly balances the exploration–exploitation tradeoff. The~GP-UCB algorithm prefers points \( x \) where \( f \) is sufficiently uncertain (large \( \sigma_{t-1}(\cdot) \)) and where the empirical rewards are high (large \( \mu_{t-1}(\cdot) \)). 

In general, to~determine the next point \( x_t \) in the domain \( D \), we define an \textit{acquisition function} \( f_t(x): D \rightarrow \mathbb{R} \) that quantifies the utility of evaluating \( f \) at any point \( x \in D \). We then select the next point by minimizing this acquisition function:
\[ x_t = \operatorname{argmin}_{x \in D} f_{t}(x), \]
where we proceed to evaluate $x^{\star}$ at time \( t \). For~example, in~the GP-UCB algorithm, we have $f_{t}(x)=\mu_{t-1}(x) + \beta_t^{1/2} \sigma_{t-1}(x)$. This approach guides the selection of evaluation points to efficiently explore the domain based on the information gathered so~far.

The pseudocode of the GP-UCB algorithm is given in Algorithm \ref{alg:GP-UCB}.

\begin{algorithm}[H]
\caption{GP-UCB}
\label{alg:GP-UCB}
\begin{algorithmic}[1]
\State \textbf{Input:} Input space $D$; GP Prior $\mu_0 = 0$, $\sigma_0 >0$.
\For{$t = 1, \ldots, T$}
    \State Choose $x_t = \arg\max_{x \in D} \left( \mu_{t-1}(x) + \sqrt{\beta_t} \sigma_{t-1}(x) \right)$.
    \State Sample $y_t = f(x_t) + \epsilon_t$.
    \State Perform a Bayesian update to obtain $\mu_t(\cdot)$ and $\sigma_t(\cdot)$.
\EndFor
\end{algorithmic}
\end{algorithm}

Srinivas et al.
 \cite{srinivas2009gaussian} determined an exploration level $\{\beta_t\}$ in the GP-UCB algorithm, and~uncertainty is quantified by defining $\beta_t = 2 \log(|D| \omega_t / \delta)$ , where  $\omega_t> 0,~\sum_{t \geq 1} \omega_t^{-1} = 1$, and~$\delta \in (0,1)$. From~Lemma 5.1 in~\cite{srinivas2009gaussian}, the~following event holds with probability at least $1-\delta$
$$  
\mathcal{A} := \left\{\left|f(\boldsymbol{x}) - \mu_{t-1}(\boldsymbol{x})\right| \leq \sqrt{\beta_t} \sigma_{t-1}(\boldsymbol{x}), ~\forall \boldsymbol{x} \in D,~\forall t \geq 1\right\}  
$$  
Conditioned on $\mathcal{A}$, the~ choice $x_t = \arg\max_{x \in D} \left(\mu_{t-1}(x) + \sqrt{\beta_t} \sigma_{t-1}(x)\right)$ then implies  
$$  
f\left(x^{\star}\right) \leq \mu_{t-1}\left(x^{\star}\right) + \sqrt{\beta_t} \sigma_{t-1}\left(x^{\star}\right) \leq \mu_{t-1}\left(x_t\right) + \sqrt{\beta_t} \sigma_{t-1}\left(x_t\right) \leq f\left(x_t\right) + 2 \sqrt{\beta_t} \sigma_{t-1}\left(x_t\right).  
 $$  
From this, we derive the inequality:  
 $
 f\left(x^{\star}\right) - f\left(x_t\right) \leq 2 \sqrt{\beta_t} \sigma_{t-1}\left(x_t\right)$. The~cumulative regret over  $T$  rounds is bounded as
$$  
\operatorname{Reg}_T(\pi; \mathcal{F}, v) := \sum_{t=1}^T \left(f(x^{\star}) - f(x_t)\right) \leq 2 \sum_{t=1}^T \sqrt{\beta_t} \sigma_{t-1}\left(x_t\right) \leq 2 \sqrt{T \beta_T \sum_{t=1}^T \sigma_{t-1}^2\left(x_t\right)},  
$$  
with probability at least $1-\delta$, 
where the last inequality follows from the Cauchy–Schwarz inequality. Lemma 5.4 in~\cite{srinivas2009gaussian} shows $\sum_{t=1}^T \sigma_{t-1}^2(x_t) = O(\gamma_T)$, where the maximum information gain $\gamma_T$ quantifies entropy-based complexity:
\[
    \gamma_T := \max_{A \subset D: |A|=T} I(y_A ; f_A), \qquad I(y_A ; f) := I(y_A ; \mathbf{f}_A) = \frac{1}{2} \log \left| I + \sigma^{-2} K_A \right|.
\]
Here  \(\mathbf{f}_A = [f(x)]_{x \in A}\). \( K_A = \left[k(x, x')\right]_{x, x' \in A} \) for $A\subset D$, and~the function \(I(\cdot; \cdot)\) is measured by the \textit{{mutual information}} (or \textit{{information gain}}). Here, {the} 
 entropy \( H(X) \) of a r.v. \( X \) with density \( p \) is defined as
$H(X) = -\sum_{x \in \mathcal{X}} p(x) \log p(x) = E[-\log p(X)].$
The conditional entropy \( H(Y \mid X) \) is defined as $H(Y \mid X) = E[f(X, Y)]$, where \( f(x, y) = -\log(p(y \mid x)) \) for r.v.s \( X \) and \( Y \) with conditional density \( p(x|y) \). The~mutual information between \( f \) and observations \( y_A = f_A + \epsilon_A \) is $I(y_A ; f) = H(y_A) - H(y_A \mid f)$ for $A \subset D.$ For a multivariate Gaussian distribution, we have
$H(N(\mu, \Sigma)) = \frac{1}{2} \log |2 \pi e \Sigma|,$
leading to $ I(y_A ; f) = \frac{1}{2} \log \left| I + \sigma^{-2} K_A \right|.$), see~\cite{thomas2006elements}. By~letting weights $\omega_t = t^2 \pi^2 / 6$, one obtains the following result.}

\begin{Theorem}[Information regret bounds, Theorem 1 in~\cite{srinivas2009gaussian}]
Let $\delta \in (0,1)$ and
$$\beta_t = 2 \log(|D| t^2 \pi^2 / (6 \delta)). $$
Running \textit{GP-UCB} with $\beta_t$ for a sample $f$ of a \textit{GP} with mean function zero and covariance function $k(x, x')$, we obtain an information-theoretic regret bound with high probability,
$$P \left\{ \operatorname{Reg}_T(\pi;\mathcal{F},v) \leq \sqrt{C_1 T \beta_T \gamma_T}, \forall T \geq 1 \right\} \geq 1 - \delta,~where~C_1 = 8 / \log(1 + \sigma^{-2}).$$
\end{Theorem}

Further developments, such as the improved GP-UCB (IGP-UCB) algorithm proposed by~\cite{chowdhury2017kernelized}, extend this framework by relaxing the noise assumption to sub-Gaussian and focusing on cases where the expected reward function lies within the reproducing kernel Hilbert space of the Gaussian process kernel. Moreover, if~contextual information about experimental conditions is available, the~GP-UCB algorithm can be adapted to handle contextual bandit problems effectively~\cite{krause2011contextual}. Beyond~regret minimization, Gaussian process bandits are also applied to level set estimation problems, where the goal is to classify points in the domain $D$ into superlevel ($H = \{ \boldsymbol{x} \in D \mid f(\boldsymbol{x}) > h \}$) or sublevel ($L = \{ \boldsymbol{x} \in D \mid f(\boldsymbol{x}) \leq h \}$) sets with high probability, as~explored in~\cite{gotovos2013active,hayashi2024gaussian}. These developments highlight the versatility and theoretical foundations of Gaussian process bandits in tackling both classical and contextual decision-making~problems.



\subsection{Thompson Sampling Algorithm of~SCAB}

{Refs.} 
 \cite{mathieugaussian,chowdhury2017kernelized} proposed a generalization of the TS algorithm for the SCAB problem, where the reward functions are sampled from a Gaussian process. Specifically, early work~\cite{russo2014learning} studied the TS algorithm of discretized SCAB. Given observations ${\mathcal D}_{t-1}$, sampling
$f_{t}(\cdot)\sim \mathcal{GP}(\mu_{t-1}(\cdot), k_{t-1}(\cdot,\cdot))$, then one selects the next arm $x_t$ according to
$$
x_t = \underset{x \in D}{\operatorname{argmax}}\{ f_{t-1}(x) \}.
$$
The pseudocode of the GP-TS algorithm is summarized in Algorithm \ref{alg:GP-TS}.

\begin{algorithm}[H]
\caption{GP-TS}
\label{alg:GP-TS}
\begin{algorithmic}[1]
\State \textbf{Input:} $K$,$T$. {GP Prior $\mu_0 = 0$, $k_0(\cdot, \cdot)$, $\sigma^2$.}
\For{$t = 1, \ldots, T$}
    \State Sampling $f_{t}(\cdot)\sim \mathcal{GP}(\mu_{t-1}(\cdot), k_{t-1}(\cdot,\cdot))$.
    \State Choose $x_t = \arg\max_{x\in D} f_{t-1}(x)$.
    \State Observe $y_t = f(x_t) + \epsilon_t$.
    \State $\mu_t(x) = \mathbf{k}_{t-1}(x)^T \left(\mathbf{K}_{t-1} + \sigma^2 I\right)^{-1} y_t$.
    \State $k_t(x, x') = k(x, x') - \mathbf{k}_{t-1}(x)^T \left(\mathbf{K}_{t-1} + \sigma^2 I\right)^{-1} \mathbf{k}_{t-1}(x')$.
    \State Perform update to obtain $\mu_t(\cdot)$ and $k_t(\cdot,\cdot)$.
\EndFor
\end{algorithmic}
\end{algorithm}
Moreover, 
Chowdhury and Gopalan~\cite{chowdhury2017kernelized} suggested that $f_{t}(\cdot)\sim \mathcal{GP}(\mu_t(\cdot), v_t k_{t-1}(\cdot,\cdot))$ in the GP-TS algorithm for a given scaling factor $v_t$.

In practice, accurately specifying the Gaussian process prior can be challenging. Misspecifications may arise due to several factors~\cite{neiswanger2021uncertainty}, including:
\begin{itemize}
\item Incorrect kernel (e.g., using a squared exponential kernel instead of a Matérn kernel);
\item Poor estimates of kernel parameters (e.g., variance parameter in Gaussian kernel);
\item Heterogeneous smoothness of the function \( f \) over the domain \( \mathcal{X} \).
\end{itemize}

{To} 
 obtain robust estimations under such uncertainties, 
 Clare et al.
 \cite{clare2020confidence} applied confidence bound minimization to SCAB using Student's t-processes. They proposed an alternative robust TS algorithm that addresses known weaknesses in Gaussian processes. Furthermore, %
 Neiswanger and Ramdas
 \cite{neiswanger2021uncertainty} utilized the GP framework as a working model without assuming the correctness of the Gaussian prior. Instead, they constructed a confidence sequence for the unknown function using martingale~techniques.

\subsection{Comparison of GP-UCB and GP-TS~Methods}
In GP optimization, both GP-UCB and GP-TS methods aim to efficiently explore and exploit the function \( f\) to identify the optimal points $x^{\star}$ in the domain \( D \). The~key distinction lies in how they define the acquisition function used to select the next evaluation point (this is also true for the UCB and TS methods in MAB problems for discrete and finite $D$).

Although both GP-UCB and GP-TS require a prior, the~role of the prior is different. For~GP-UCB, the~prior is used to model the randomness in $f$ at $t = 0$, while exploration arises from $\{\beta_t\}$. In~contrast, for~GP-TS, the~prior serves the usual role in TS algorithms, with~exploration stemming from posterior~sampling.

The comparison between GP-UCB and GP-TS is presented in Table~\ref{tab:gp_comparison}.

In essence, while both methods aim to select the next evaluation point \( x_t \) that is most informative for learning \( f \), GP-UCB does so by constructing a deterministic acquisition function that upper-bounds the true function with high probability, whereas GP-TS uses stochastic sampling to guide its selection, capturing the uncertainty in a probabilistic manner. The~choice between these methods may depend on the specific context of the problem, computational resources, and~the desired balance between exploration and~exploitation.

\begin{figure}[H] 
   \includegraphics[width=0.7\textwidth]{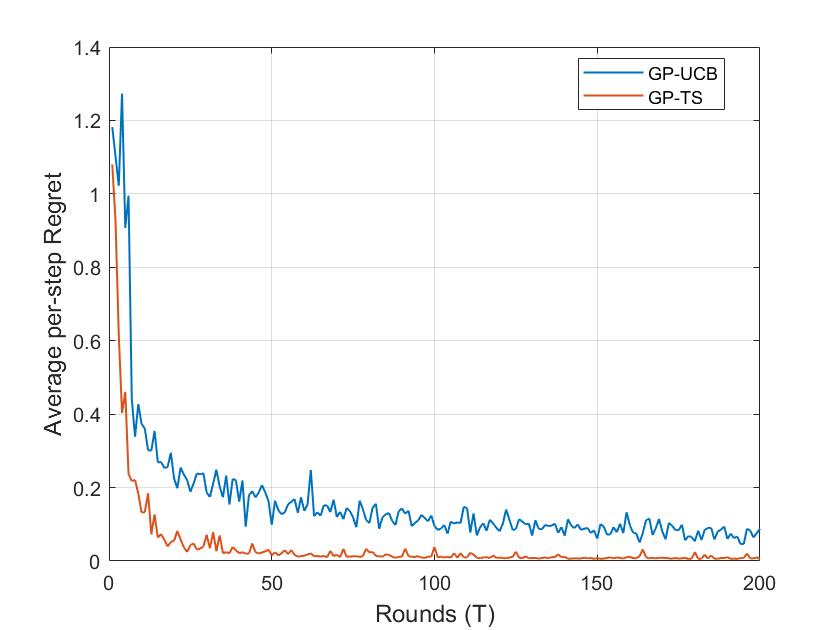} 
   \caption{{Average regret comparisons of GP-UCB and GP-TS algorithms.}}  
 \label{fig:GPsumulate} 
\end{figure}

{We evaluate the GP-UCB and GP-TS algorithms through a simulation experiment. 
First, we generate 500 independent realizations, each consisting of 100 points uniformly sampled from the interval \([0, 1]\). Given these sample points, we draw the corresponding response values from a squared exponential kernel $k\left({x}, {x}^{\prime}\right) = \exp\left\{-\frac{1}{2 l^2}\left\|{x}-{x}^{\prime}\right\|^2\right\}$
with a hyperparameter \(l=0.25\).

\begin{table}[H]\small
        \caption{{Summary} 
 of differences between GP-UCB and~GP-TS.}
    \label{tab:gp_comparison}
    \begin{tabularx}{\textwidth}{>{\centering\arraybackslash}p{4cm}XX}
        \toprule
        \multicolumn{1}{c}{\textbf{Aspect}} & \multicolumn{1}{c}{\textbf{GP-UCB}} & \multicolumn{1}{c}{\textbf{GP-TS}} \\
        \midrule
        Acquisition Function Definition & Combines the posterior mean and variance with a confidence parameter to form an upper confidence bound. & Directly uses a function sampled from the GP posterior as the acquisition~function. \\
        \midrule
        Exploration vs. Exploitation & Explicitly balances exploration and exploitation through the parameter \( \beta_t \), which scales the influence of the uncertainty term \( \sigma_{t-1}(x) \). & Implicitly balances exploration and exploitation through the randomness of the sampled functions, capturing the posterior~uncertainty. \\
        \midrule
        Parameter Tuning & Requires careful selection of \( \beta_t \) to ensure optimal performance and convergence guarantees. & Generally requires less parameter tuning since the balance is managed through posterior~sampling. \\
        \midrule
        Computational Considerations & Involve optimization over the acquisition function that includes mean and variance. & Optimization is performed over a single sampled function, which can be computationally efficient but may require multiple samples to stabilize~performance. \\
        \bottomrule
    \end{tabularx}
\end{table}

When $l$ is unknown in real data, we estimate the hyperparameter \(l\) following the methodology outlined in Section~2.2 of~\cite{williams2006gaussian}, which employs maximum likelihood estimation~(MLE). For~the squared exponential kernel, the~log-likelihood function is given~by
\[
N(y \mid 0, \mathbf{K}) = (2\pi)^{-n/2} |\mathbf{K} + \sigma^2I|^{-1/2} 
\exp\left(-\frac{1}{2} y^\top (\mathbf{K} + \sigma^2I)^{-1} y\right).
\]
The MLE problem is then reformulated as the following optimization task:
\[
l^\ast \in \arg \min_{l \in [0, 0.5]} \left\{ \frac{1}{2} \log |\mathbf{K} + \sigma^2I| + \frac{1}{2} y^\top (\mathbf{K} + \sigma^2I)^{-1} y \right\}.
\]
Solving this optimization problem yields an optimal hyperparameter value of \(l = 0.2575\). This estimate is subsequently employed to simulate and evaluate the performance of the GP-UCB and GP-TS~algorithms.

Following the procedure described above, we generated an additional 150 samples and subsequently conducted 200 rounds of evaluations, with~the exploration–exploitation trade-off parameter set to \(\beta = 2.0\). We then compute the average per-step regret over these 200 rounds, which served as the primary performance metric for each algorithm. As~illustrated in Figure~\ref{fig:GPsumulate}, GP-TS consistently outperforms GP-UCB, achieving smaller regret under our experimental conditions.}

\unskip

\subsection{SCAB in Reproducing Kernel Hilbert~Space}
In practical scenarios, the~objective function exhibits considerable complexity, necessitating strategies that optimize cumulative rewards or minimize regret over time. This is achieved by systematically balancing exploration and exploitation within the action space to acquire information about, which resides in structured function spaces such as Hölder, Besov, or~Reproducing Kernel Hilbert Spaces (RKHS; see~\cite{singh2021continuum}). Our analysis centers on RKHS, where the connection between GPs and their associated covariance functions is rigorously formalized through the framework of RKHS~theory.

\begin{Definition}[Definition of RK and RKHS]
Let \( \mathcal{H} = \mathcal{H}(\mathcal{X}) \) be a Hilbert space of functions \( f: \mathcal{X} \rightarrow \mathbb{R} \) defined on a non-empty set \( \mathcal{X} \). A~function \( k: \mathcal{X} \times \mathcal{X} \rightarrow \mathbb{R} \) is termed a reproducing kernel of \( \mathcal{H} \) if it~satisfies the following:
\begin{itemize}
    \item The reproducing kernel (RK): \( \forall x \in \mathcal{X}, k(\cdot, x) \in \mathcal{H} \);
    \item The reproducing property: \( \forall x \in \mathcal{X}, \forall f \in \mathcal{H},~\langle f, k(\cdot, x)\rangle_k = f(x) \).
\end{itemize}

\noindent {When} 
 a Hilbert space \( \mathcal{H}\) possesses an RK \( k \), it is referred to as an RKHS. The~RKHS norm \( \|f\|_k := \sqrt{\langle f, f\rangle_k} \) measures the smoothness of the function \( f \). 
\end{Definition}
Instead of assuming an unknown function sampled from a Gaussian process prior, we consider a more agnostic scenario where the function has low complexity, measured by its norm in an RKHS with kernel $k({x}, {x}^{\prime})$. We also consider noise variables \(\varepsilon_t\), forming an arbitrary martingale difference sequence with \(E[\varepsilon_t \mid \varepsilon_{<t}] = 0\) and uniformly bounded by a constant \(\sigma\). Despite prior and noise model misspecification, the~GP-UCB algorithm is still employed, and~it is shown to achieve sublinear regret in this agnostic~setting.

\begin{Theorem}[RKHS information regret bounds, Theorem 3 in~\cite{srinivas2009gaussian}]
Let $\delta \in(0,1)$. Assume that the true underlying $f$ lies in the RKHS $\mathcal{H}_k(D)$ corresponding to the kernel $k\left(\boldsymbol{x}, \boldsymbol{x}^{\prime}\right)$, and~that the noise $\varepsilon_t$ has zero mean conditioned on the history and is bounded by $\sigma$ almost surely. In~particular, assume $\|f\|_k^2 \leq$ $B$ and let $\beta_t=2 B+300 \gamma_t \log ^3(t / \delta)$. Running GP-UCB with $\beta_t$, prior $\mathrm{GP}\left(0, k\left(\boldsymbol{x}, \boldsymbol{x}^{\prime}\right)\right)$, and~noise model $N\left(0, \sigma^2\right)$, we obtain an anytime regret bound with high probability (over the noise).
$$
P\left\{ \operatorname{Reg}_T(\pi;\mathcal{F},v) \leq \sqrt{C_1 T \beta_T \gamma_T},~\forall T \geq 1\right\} \geq 1-\delta~\text{where}~C_1=8 / \log (1+\sigma^{-2}).
$$
\end{Theorem}

\subsection{Relationship Between SCAB and Functional Data~Analysis}

SCAB and functional data analysis (FDA,~\cite{hsing2015theoretical}) are two areas in statistics and machine learning that, at~first glance, may seem distinct due to their differing primary objectives and contexts. However, there exists a profound connection between them, primarily through the lens of function estimation and analysis over continuous domains~\cite{shi2011gaussian,zhang2023growing}.

In SCAB, an~agent sequentially selects actions from a continuous action space \( \mathcal{X} \subset \mathbb{R}^d \) as a decision-making problem. At~each time \( t \), the~agent selects an action \( x_t \in \mathcal{X} \) and observes
where \( f: \mathcal{X} \rightarrow \mathbb{R} \) is an unknown reward function, and~\( \epsilon_t \) represents random noise. 
In contrast, in~FDA, it deals with statistical analysis where the primary data units are functions or curves rather than scalar or vector observations. The~data are assumed to be realizations of random functions defined over a continuous domain \( \mathcal{T} \) (e.g., time, space). The~goal in FDA is to analyze these functional observations to understand underlying patterns, make predictions, or~perform~classifications.

The connection between SCAB and FDA is evident as both focus on learning and analyzing functions over continuous~domains.

\begin{enumerate}
\item[1.] \textbf{{Function estimation:}
}
\end{enumerate}

 In SCAB, the~agent seeks to estimate the unknown reward function \( f\) to make informed decisions. This involves constructing estimators \( \widehat{f}_t(x) \) based on past observations \( \mathcal{H}_{t-1}\). FDA focuses on estimating the underlying functional form from observed data, often involving smoothing techniques, basis expansions (e.g., Fourier, wavelets), or~functional principal component~analysis.

\begin{enumerate}
\item[2.] \textbf{{Continuous domains:}}
\end{enumerate}

SCAB and FDA operate over continuous domains. In~SCAB, actions are selected from a continuous space \( \mathcal{X} \), and~in FDA, functions are defined over a continuous domain \( \mathcal{T} \).

\begin{enumerate}
\item[3.] \textbf{{Handling noise and uncertainty:}}
\end{enumerate}

SCAB: observed rewards are noisy evaluations of \( f \), and~the agent must account for this uncertainty in its estimates and decisions. FDA: observations are often contaminated with noise, and~FDA aims to recover the true underlying functional relationships~\cite{yao2005functional,zhou2023functional}.

\begin{enumerate}
\item[4.] \textbf{{Methodological parallels:}}
\end{enumerate}

Both fields frequently employ nonparametric methods. In~SCAB, nonparametric regression techniques like GPs are used to model \( f \) without strong parametric assumptions. Similarly, FDA relies on nonparametric smoothing and functional regression methods. Bayesian methods are prevalent in both areas. In~SCAB, Bayesian optimization and TS utilize posterior distributions over functions. In~FDA, Bayesian functional models incorporate prior beliefs about functional~forms.

\begin{enumerate}
\item[5.] \textbf{{Dimensionality considerations:}}
\end{enumerate}

High-dimensional function estimation is a challenge in both SCAB and FDA. Techniques to mitigate the curse of dimensionality, such as dimension reduction and exploiting smoothness or sparsity, are common to both fields. Developing scalable algorithms, such as distributed computing and online learning methods, is~crucial.

\begin{enumerate}
\item[6.] \textbf{{Transfer of techniques:}}
\end{enumerate}

 Methods developed in FDA for function smoothing and estimation can be adapted for use in SCAB. For~instance, functional basis expansions or additive models could enhance the representation of the reward function in SCAB~\cite{cai2022stochastic}.  In~contrast, SCAB's exploration–exploitation strategies and sequential decision-making frameworks can inspire new FDA approaches to data collection and experimental design~\cite{ji2017optimal}, especially when observations are expensive or time-consuming to obtain. SCAB inherently involves sequential data acquisition, which aligns with the emerging area of sequential FDA, where data arrive over time, and~analyses need to adapt~accordingly.

The relationship between SCAB and FDA is rooted in their mutual focus on function estimation over continuous domains under uncertainty. Recognizing this connection opens avenues for methodological advancements by taking advantage of the strengths of both fields. Incorporating FDA techniques into SCAB can improve function estimation and uncertainty quantification, while applying SCAB principles in FDA can enhance adaptive sampling and experimental design. By~recognizing and leveraging the connections between these fields, researchers can develop more robust methodologies, improve computational efficiency, and~address complex real-world problems (such as healthcare, environmental science, and~industrial technology) that require both sequential decision-making and sophisticated functional data~analysis.

\section{Advanced~Topics}

In Section~\ref{sec:clb}, we introduced $K$-armed contextual bandits. However, as~the context space grows (e.g., factors like user type, age, location, or~time since last active), running separate $K$-armed bandits for each unique context becomes infeasible. The~true power of contextual bandits lies in leveraging models that capture the relationship between experimental conditions and outcomes (e.g., conversion rates). Instead of treating each context independently, these models allow researchers to share information across contexts, enabling efficient handling of large and complex context~spaces.

This section selectively reviews advanced bandit methodologies, starting with contextual bandits and extending the discussion to non-contextual bandits (e.g., MAB) from a broader perspective, contrasting with the approaches covered in Section~\ref{sec:MAB_alg}. Finally, we explore applied bandit problems and bandits with unknown variance proxies, emphasizing recent developments mostly from top statistical journals and~conferences.

\subsection{Contextual~Bandits}


Sarkar
 \cite{sarkar1991one} extended Woodroofe's Bayesian sequential allocation work by incorporating covariates and demonstrating that the myopic rule remains asymptotically optimal under geometric~discounting.

Yang and Zhu 
\cite{yang2002randomized} introduced a nonparametric approach to estimate the relationship between rewards and covariates. Their randomized allocation rule effectively balances exploration and exploitation, demonstrating the long-term benefits of integrating covariate data into decision-making. Similarly, 
Perchet and Rigollet 
\cite{perchet2013multi} developed the Adaptively Binned Successive Elimination (ABSE) policy, which dynamically partitions the covariate space to maximize cumulative rewards. 
Baransi et al.
\cite{baransi2014sub}
introduced a novel sampling algorithm called BESA (Best Empirical Sampled Average) for the MAB problems, which is a fully non-parametric genalization of TS. 
Chan
\cite{chan2020multi} extended BESA and proposed a non-parametric solution using subsample mean comparison (SSMC) for unknown reward distributions; then 
Ai et al.
\cite{Ai2021} studied the binned subsample mean comparisons (BSCM) policy for allocating arms by decomposing the covariates region and comparing the subsample~means.

Cai et al.
\cite{cai2024transfer} extended the use of covariates in MAB problems by addressing transfer learning under covariate shifts. Their nonparametric contextual bandit model leverages data from source bandits to improve decision-making in new environments, achieving minimax regret by adapting to changes in covariates. In~the context of high-dimensional covariates, 
Qian et al.
\cite{qian2024adaptive} proposed a multi-stage arm allocation algorithm that integrates arm elimination and randomized assignment strategies, demonstrating robustness across various real-world~applications.

Covariates also play a crucial role in dynamic pricing models. 
Liu et al. 
\cite{liu2024contextual} proposed a dynamic strategic pricing policy in which buyers manipulate their observable features to influence prices. By~accounting for these covariates, the~model achieves optimal regret bounds while adapting to buyer behavior. By~the idea of the ETC algorithm, 
Fan et al. 
\cite{fan2024policy} introduced a semi-parametric model for contextual dynamic pricing, which integrates both parametric and non-parametric components to optimize pricing decisions based on market~conditions.

Battiston et al. 
\cite{battiston2018multi} used the Hierarchical Pitman-Yor (HPY) process for Bayesian nonparametric modeling, which accounts for shared species across different populations, akin to incorporating covariates to manage exploration–exploitation trade-offs. Their TS strategy effectively balances species discovery across multiple contexts. 
Chen et al. 
\cite{chen2021statistical} expanded the application of contextual MAB by focusing on statistical inference in online decision-making. By~exploring the $\epsilon$-greedy policy in a linear reward model, they developed an inverse propensity weighted (IPW) estimator and proposed an online weighted least squares (WLS) estimator to correct for sampling bias, enhancing decision-making accuracy in news recommendation~systems.

 Wang et al. 
\cite{wang2020nearly} addressed the challenge of high-dimensional linear bandit problems by proposing a best subset selection method for parameter estimation. This approach leverages covariates to address the complexity of small action spaces in high-dimensional settings, demonstrating its applicability in personalized recommendations and online advertising. 
Zhou et al. 
\cite{zhou2024stochastic} extended the application of covariates to multi-dimensional tensor bandits, proposing low-rank tensor structures to optimize decisions in multi-dimensional~environments.

Zhu et al. 
\cite{zhu2023principled} investigated principled reinforcement learning with human feedback (RLHF,~\cite{ouyang2022training}) by employing pairwise comparisons within the contextual bandit framework. Building upon this foundation, 
Scheid et al.
\cite{scheid2024optimal} aimed to formalize the reward training model in RLHF. They framed the selection of an effective dataset as a simple regret minimization task and addressed it using a linear contextual dueling bandit approach. {Recently, DeepSeek-R1's technical contribution~\cite{guo2025deepseek} diverges from conventional RL frameworks employing human/AI feedback through PPO, instead introducing a novel open-source reasoning model alongside a comprehensive training methodology for LLMs. The~architecture implements Group Relative Policy Optimization (GRPO), a~rule-driven RL framework that eliminates dependency on external value models while preserving PPO's training stability. This methodological innovation establishes GRPO as a computationally efficient paradigm for LLM alignment, marking substantial progress in RL implementations for LLMs.}

{Another key perspective in contextual bandits is evaluating policy quality, typically measured by the average reward obtained under the policy. While simulatable environments like video games allow low-cost policy evaluation, real-world applications such as autonomous driving or medical treatment face challenges due to the high cost, risk, or~ethical concerns of deploying new policies. Simulated environments offer an alternative, but~designing high-fidelity simulators, such as one capturing all medical conditions, is often more complex than policy optimization itself. For~example, it is difficult to simulate a patient that includes all medical conditions~\cite{li2019perspective}. Off-policy evaluation (OPE) assesses a policy (the ``target policy'') using only historical data from a different policy (the ``behavior policy''), without~executing the target policy \citep{richard2012barto}. This is particularly relevant in scenarios resembling contextual bandits, where actions affect immediate rewards but not future states~\cite{li2019perspective}. The~observed data only reveal the reward and context (state) of the chosen actions. 
Uehara et al.
\cite{uehara2022review} provided a comprehensive review of the (OPE) problem, adopting a rigorous statistical lens and centering their attention on the Markovian framework that encompasses scenarios with either long or infinite horizons.}

\subsection{Non-Contextual~Bandits}

Many algorithms that do not incorporate covariates have demonstrated remarkable effectiveness. These algorithms encompass a range of strategies, from~asymptotic to non-asymptotic approaches, addressing diverse applications such as continuous value problems and high-dimensional challenges. These studies have provided crucial insights into balancing exploration and exploitation, significantly contributing to both theoretical innovation and practical~applications.

The foundational work by 
Lai
introduced an adaptive allocation rule based on UCB and demonstrated its asymptotic optimality. This approach is effective under both Bayesian and frequentist frameworks, and~is applicable to various distributions, such as exponential families. Building upon Lai's work,
Berry et al. 
\cite{berry1997bandit} investigated the MAB problem with an infinite number of arms, proposing strategies to minimize long-term failure rates, which have since been applied in areas like clinical trials and resource exploration. In~the context of finite arms,
Clayton and Berry 
\cite{clayton1985bayesian} introduced a Bayesian ``stay-on-the-winner'' rule, showing that it approaches optimality within a finite time horizon. Concurrently, 
Kelly 
\cite{kelly1981multi} proposed the ``least failures rule,'' optimizing rewards as the discount factor nears one. These contributions laid the foundation for further research in finite arm~scenarios.

Gittins 
\cite{gittins1979bandit} introduced the concept of dynamic allocation indices (DAI), which simplifies MAB problems by providing efficient computation and guiding optimal decision-making. This approach makes previously challenging problems more tractable, particularly in clinical trials and stochastic scheduling.
Whittle 
\cite{whittle1980multi} expanded on this by introducing the Gittins index, which assigns an index to each arm to simplify decision-making and provided a proof of its~optimality.

Further research on asymptotic methods includes 
\cite{glazebrook1980randomized}, who investigated randomized dynamic allocation indices (RDAI) for Bernoulli populations, and
\cite{fuh2000asymptotically}, who optimized sequential job processing under stochastic conditions. These studies advanced the understanding of asymptotic optimization in MAB problems. Asymptotic methods have also been applied to continuous rewards. 
Gittins and Wang 
\cite{gittins1992learning} developed a dynamic allocation index that adjusts for the uncertainty of each arm's reward potential, achieving an optimal balance between immediate rewards and long-term learning. 
Chen et al. 
\cite{chen2023strategic} extended traditional statistical methods by incorporating the two-armed bandit model into hypothesis testing. This approach challenges the conventional assumption of exchangeability in i.i.d. data and introduces a strategy-specific test statistic, termed the ``strategy statistic,'' which utilizes the decision-making process of the two-armed bandit to enhance testing power.
Han et al.
\cite{han2024ucb} presented a rigorous regret analysis and adaptive statistical inference framework for UCB algorithms by leveraging a deterministic characterization of the number of arm pulls. This approach offers deeper insights into the algorithm's asymptotic behavior and~performance.

In finite-time settings, 
Fox 
\cite{fox1974finite} highlighted the limitations of the ``play-the-winner'' strategy for two-armed bandit problems. Through Monte Carlo experiments, alternative policies were proposed that demonstrate superior performance with limited sample sizes.
Li and Zhang 
\cite{li1992asymptotically} extended this line of research by developing an asymptotically efficient allocation rule for two Bernoulli populations, minimizing regret in sequential sampling. Non-asymptotic methods are increasingly important for finite-time applications.
Ren and Zhang
\cite{ren2024lai_ucb} derived precise non-asymptotic regret bounds for UCB algorithms with a fixed exploration parameter in the context of Gaussian rewards. While their analysis focuses on the Gaussian assumption, it can be extended to accommodate sub-Gaussian rewards in future research, broadening its~applicability.

The MAB problem has also been studied in continuous value settings. 
Cappe et al. 
\cite{cappe2013kullback} introduced the KL-UCB algorithm, which uses the Kullback–Leibler divergence to compute confidence bounds for one-dimensional exponential families, proving its asymptotic optimality in scenarios involving continuous reward distributions. 
Kaufmann
\cite{kaufmann2018bayesian} extended this work with the Bayes-UCB algorithm, which selects arms based on posterior quantiles, further enhancing the effectiveness of decision-making in continuous settings by utilizing posterior information to better balance exploration and~exploitation.

Addressing continuous value rewards presents additional challenges.
Ginebra and Clayton  
\cite{ginebra1995response} introduced the concept of response surface bandits, which optimize continuous rewards by leveraging controllable variables to navigate complex environments.
Cai and Pu
\cite{cai2022stochastic}  further advanced this area by tackling multi-dimensional SCAB, presenting an adaptive algorithm that mitigates the curse of dimensionality and enhances performance in high-dimensional scenarios. 
Wang et al.
\cite{wang2024pre} proposed HyperBO, which improves Bayesian optimization by automating the construction of pre-trained  GP priors. This method enhances efficiency in optimizing complex black-box functions, achieving up to improvements across~\mbox{benchmarks}.

These studies provide comprehensive solutions to the MAB problem, from~asymptotic approaches for infinite-horizon settings to non-asymptotic methods for finite-time scenarios. They address the challenges of continuous rewards and high-dimensional data, offering valuable tools and strategies for managing uncertainty and optimizing decisions across diverse~applications.

\subsection{Applied~Bandits}

The MAB problem has shown significant potential in clinical trial design and other applied black-box optimization problems, particularly in areas ranging from treatment allocation optimization to false discovery rate (FDR) control, selection of critical image features, and~drug design~problems. 

The earliest contribution by 
\cite{berry1978modified}  proposed a strategy for two-armed clinical trials, aiming to maximize patient success rates. This approach balanced learning the efficacy of different treatments with optimizing patient outcomes, providing a foundation for subsequent optimization designs and proving versatile for both known and unknown trial lengths. Building on this, 
Bather
\cite{bather1981randomized} expanded the concept by introducing randomized allocation rules, which prioritize selecting the current best treatment while maintaining exploration of suboptimal treatments. This strategy proved to converge to the optimal allocation in the long run, reinforcing the balance between exploration and exploitation, particularly in sequential experimental settings.
Cheng and Berry 
\cite{cheng2007optimal} further advanced this work by proposing the r-optimal design, a~compromise between deterministic and randomized designs. This design ensures that each treatment arm has a minimum selection probability \(r\), reducing bias and achieving asymptotic optimality in large-scale trials. By~minimizing allocations to inferior treatments, the~method enhances trial efficiency and fairness. 
Mozgunov and Jaki 
\cite{mozgunov2020information} introduced an information-theoretic response-adaptive design, leveraging Shannon's differential entropy to dynamically adjust the probability of selecting each treatment arm. This approach maximizes the likelihood of assigning patients to superior treatments and is particularly effective in trials with complex multinomial endpoints and ethically constrained, high-cost~settings.

Beyond clinical trial design, 
Wang and Ramdas
\cite{wang2022false} applied MAB principles to FDR control, proposing the Benjamini–Hochberg (BH) procedure for false discovery rate (FDR) control that utilizes e-values, a~modification of the classical Benjamini–Hochberg method. Their approach is more robust in the presence of complex dependence structures, showing effectiveness not only in finance but also in statistical control within MAB problems. By~establishing a novel connection between combinatorial binary bandits and spike-and-slab variable selection, 
Liu and Ro{\v{c}}kov{\'a}
\cite{liu2023variable} proposed a stochastic optimization approach to subset selection known as Thompson Variable Selection (TVS). This method leverages the principles of TS within the variable selection framework, providing an efficient probabilistic algorithm that balances exploration and exploitation when identifying optimal subsets of~variables.

In addition, the~MAB model is extensively applied in the pricing strategy. 
Misra, Schwartz, and Abernethy 
\cite{misra2019dynamic} innovated real-time pricing for online retailers lacking complete demand data, employing MAB algorithms combined with microeconomic choice theory. Validated by Monte Carlo simulations, their method significantly reduces revenue losses and enhances profit potential compared with traditional methods. Following this,
Jain et al. 
\cite{jain2024effective} developed a novel algorithm that integrates discrete choice modeling with TS to address limited demand information in retail. This approach not only minimizes losses from suboptimal pricing but~also integrates pricing and promotional strategies within a unified demand model, significantly improving retail~effectiveness.

In computer vision, 
Duan et al. 
\cite{duan2023bandit} introduced Bandit Interpretability via Confidence Selection (BICS), a~model-agnostic framework that leverages the MAB paradigm and the UCB algorithm to identify critical image regions. This approach delivers precise and compact explanations for deep neural networks, significantly improving interpretability across various~applications.


Digital twins, virtual replicas of physical systems, have also benefited from applied bandits. {Binois et al.}~\cite{binois2019replication} introduced a sequential learning algorithm for heteroskedastic GP regressions, optimizing design points using a mean-squared prediction error (MSPE) criterion. Similarly, {Makarova et al.}~\cite{makarova2021risk} developed the Risk-Averse Heteroskedastic Bayesian Optimization (RAHBO) algorithm, balancing mean and variance trade-offs during query selection. {Zhang et al.}~\cite{zhang2023digital} showed the effectiveness of GP regression in constructing digital triplets through extensive simulations, while {Karkaria et al.}~\cite{karkaria2024towards} highlighted digital twin applications in additive manufacturing.

For more reviews of applied bandits,
{Burtini et al.}~\cite{burtini2015survey} provided a review of MAB algorithms in the context of experimental design, with~particular emphasis on their application in sequential and adaptive approaches to optimize online experiments.
{Zhou} \cite{zhou2015survey} provided a comprehensive survey on MAB, categorizing algorithms based on stochastic and adversarial settings.
{Refs.} \cite{elena2021survey,letard2024bandit} reviewed the application of MAB algorithms in recommendation systems, examining the effectiveness of various types of algorithms to improve recommendation performance.
In the medical field, 
{Lu et al.}~\cite{lu2021bandit} presented a comprehensive review of the potential of MAB algorithms to improve medical decision-making and improve patient outcomes, underscoring their growing relevance in precision medicine.
{Shah} \cite{shah2020survey} provided a novel review incorporating causal inference into MAB~algorithms. 

\subsection{Unknown Variance~Proxy}

In the main context, as~in much of the literature, the~assumption is that rewards are sub-Gaussian with a known variance proxy, often set to $1$ (or known $\sigma^2$). However, in~practical scenarios where this parameter is unknown, standardizing the rewards or using an estimated variance as a substitute for the unknown $\sigma^2$ are common approaches~\mbox{\citep{wu2016conservative,lattimore2020bandit,wu2022residual}}. Unfortunately, these methods can sometimes be invalid or lack adequate exploration, particularly when the rewards exhibit specific structures~\cite{zhang2023tight,wei2023zero}.

Although this is a relatively new field, an~increasing number of researchers have recognized this issue and proposed solutions to address it. For~general sub-Gaussian rewards, {refs.}~\cite{lieber2022estimating,zhang2023tight} have explored methods for estimating such parameters, designing valid concentration inequalities, and corresponding algorithms. For~scenarios involving specific structures, 
{Wei et al.}~\cite{wei2023zero} proposed avoiding parameter estimation altogether, leveraging existing information to construct valid concentration inequalities and develop appropriate algorithms.

To keep our discussion concise, we present only the simulation results in this section. The~simulation focuses on mixed Gaussian rewards with an unknown sub-Gaussian variance proxy. 
The methods compared include ``TS'' and ``UCB (asymptotic)'', which assume that the rewards are Gaussian and use the estimated variance as the true variance. ``UCB (wrong use Hoeffding)'' treats the rewards as bounded and constructs a Hoeffding-type concentration using the maximal and minimal observed values. 
It is worth noting that these three methods are based on invalid concentration and may lack validity for the algorithm.
``UCB (estimated sub-G norm)'' follows the method proposed by~\cite{zhang2023tight}, which estimates the sub-Gaussian norm, while
``UCB (estimated variance proxy)'' adopts the approach from~\cite{lieber2022estimating} to estimate the variance~proxy.

\begin{figure}[H] 
   
   \includegraphics[width=0.9\textwidth]{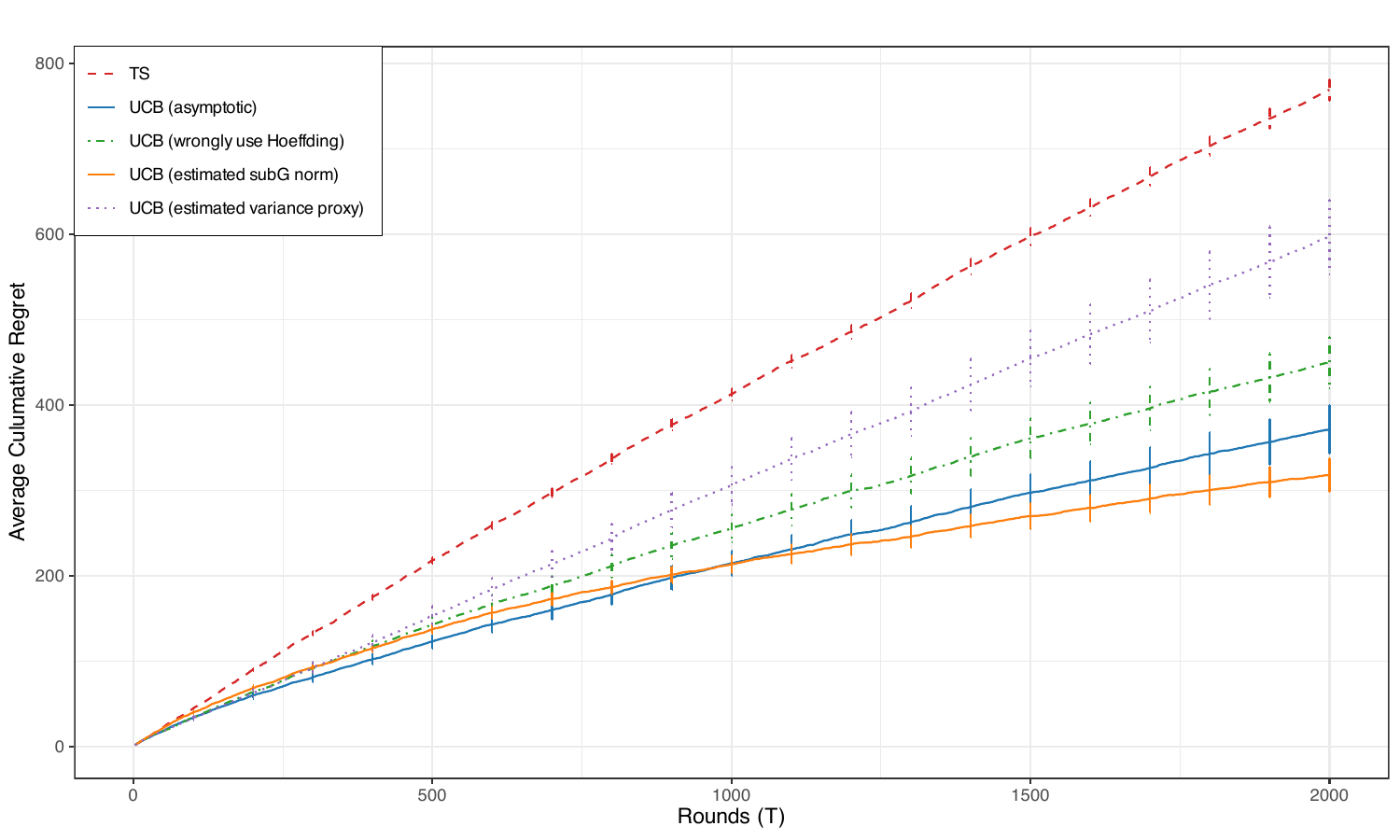} 
   \caption{Cumulative regret comparisons under mixed Gaussian rewards with an unknown sub-Gaussian variance~proxy.}
 \label{fig:new} 
\end{figure}

To evaluate algorithm performance, we conduct 100 simulations with Gaussian mixture rewards drawn from $\frac{2}{5} N \left( \frac{5}{4} \mu_k, 1\right) + \frac{3}{5} N \left( \frac{5}{6} \mu_k, 4\right)$, where $\mu_k = 1 - 0.05k$ for $k = 1, \ldots, 20$. The~algorithm parameters were configured as recommended in their respective references. 
Figure~\ref{fig:new} presents the outcomes of these simulations. 
As indicated, the~UCB method based on the estimated sub-Gaussian norm \citep{zhang2023tight} achieves the lowest regret, as~this method uses a valid concentration bound, reduces computational cost, and~avoids computational errors compared with the approach in~\cite{lieber2022estimating,zhang2023tight}.

\vspace{-9pt}

\section{Concluding Remarks and Future~Directions}

Bandit algorithms have gained significant attention and widespread applications across various fields. Accurate uncertainty quantification remains crucial for addressing the exploration–exploitation tradeoff inherent in these algorithms. In~this paper, we review two of the most commonly used methods: the UCB and TS~approaches.

Recent advancements, such as the multiplier bootstrap method \citep{wan2023multiplier} and perturbation-based methods \citep{kveton2019perturbed,kveton2020perturbed}, have also shown promise in effectively managing the exploration–exploitation tradeoff in both multi-armed and linear bandit settings. Such novel approaches are becoming increasingly relevant as researchers explore new~scenarios.

Structured and unstructured bandit problems represent two main categories within bandit frameworks, each suited to specific applications and needs. As~outlined in Table~\ref{tab:bandits}, these two types offer distinct advantages and limitations. Unstructured bandits are simple in design, easy to implement, and~intuitive, making them ideal for straightforward applications. However, they may lack the efficiency of structured bandits, which leverage additional information to enhance decision-making, especially in data-rich environments. Structured bandits, on~the other hand, can quickly adapt to complex scenarios but may suffer from over- or under-exploration when the models are incorrectly~specified.

\begin{table}[H]\small
     \caption{{Comparison} 
 of unstructured and structured~bandits.}
    \label{tab:bandits}
    \begin{tabularx}{\textwidth}{>{\raggedright\arraybackslash}p{3cm}XX}
        \toprule
        \textbf{Aspect} & \textbf{Unstructured Bandits} & \textbf{Structured Bandits} \\
        \midrule
        Arm Independence & Arms are independent for MBA and dependent for CAB & Arms are related through features or~context \\
        \midrule
        Side Information & No additional information about arms & Additional features or context are~available \\
        \midrule
        Algorithms Used & Classical algorithms: UCB and TS & Algorithms that exploit structure: Contextual Bandits and~LinUCB \\
        \midrule
        Applications & Scenarios with no contextual data (e.g., traditional slot machines) & Personalized systems, adaptive treatments, and~recommendations \\
        \midrule
        Regret & TS usually has smaller empirical regrets than UCB in MAB and CAB & TS usually has smaller empirical regrets than UCB for~LinUCB \\
        \bottomrule
    \end{tabularx}
\end{table}

{In practical applications, selecting an appropriate bandit model should be informed by the specific requirements of the environment, such as data availability, response time, and~system complexity. For~data-rich environments, structured bandits are advantageous due to their ability to incorporate additional features. Conversely, in~data-scarce or less complex scenarios, unstructured bandits may be more suitable. It’s important to note that structured bandits often require prior knowledge and careful feature engineering, which can increase both the complexity and cost of~implementation.

A promising research direction is the development of adaptive methods that can dynamically select between structured and unstructured bandit models based on the environment. Recent studies have explored model selection in contextual bandits, aiming to balance exploration and exploitation effectively. For~instance, {Muthukumar and Krishnamurthy}~\cite{muthukumar2022universal} propose data-adaptive algorithms for model selection in linear contextual bandits.

which adaptively choose the appropriate model complexity based on observed data. Similarly, {Pacchiano Camacho}~\cite{pacchiano2021model} introduces a regret balancing approach for model selection in contextual bandits and reinforcement learning,which dynamically adjusts the learning strategy to optimize performance. These approaches aim to develop algorithms that can adaptively select the most appropriate model structure in an online learning setting, thereby improving decision-making in varying environments.
}

\vspace{6pt}

\funding{{H. Zhang is supported in part by the National Natural Science Foundation of China (No. 12101630) and the Beihang University under Youth Talent Start up Funding Project (No.~KG16329201).}} 

\acknowledgments{The~authors thank Guangqiang Teng, Yanpeng Li, and Jin Liu for comments on the early versions of this~paper.}

\conflictsofinterest{{The authors declare no conflicts of interest.}}

\appendixtitles{no} 
\appendixstart
\appendix
\section[\appendixname~\thesection]{\label{app:proof_regret_bound1}}
{

\begin{proof}[Proof of Theorem~\ref{thm:subG-UCB}]
{Without} 
 loss of generality (WLOG), assume that $\mu_1=\mu_{k^{\star}}$; the regret decomposition lemma shows that
$\operatorname{Reg}_T(\pi,v)=\sum_{i=1}^K \Delta_i {E}\left[S_i(T)\right].$ Next, we bound ${E}S_i(T)$ for suboptimal arm $i\not=1$ that is not too~large.

A key observation is that after the initial period where the algorithm chooses each action
once, {{suboptimal arm} $i$} can only be chosen if its index $\mathrm{UCB}_i(t-1, \delta)$ is higher than that of an optimal arm 1. To~avoid suboptimal arms, let $G_i$ be the ‘good’ event defined by
		$$
		G_i=\left\{\frac{1}{u_i} \sum_{\tau \leq t, A_{\tau}=i} Y_{i}(\tau)+\sqrt{\frac{2}{u_i} \log \left(\frac{1}{\delta}\right)}<\mu_1\right\}\cap \left\{\mu_1<\min _{t \in [T]} \mathrm{UCB}_1(t, \delta)\right\} .
		$$
\begin{itemize}
\item For arm $i$, the~UCB for $\mu_i$ after $u_i$ observations are taken from this arm is below $\mu_1$.

\item $\mu_1$ is never underestimated by the UCB of the first arm.
\end{itemize}\vspace{-6pt}

$$		We~firstly~prove~that~if~G_i~holds~for~i\not=1,~then~S_i(T)\le u_i.$$
If $G_i$ holds and $S_i(T)>u_i$~(arm $i$ is played at least $u_i+1$ times over
the $T$ rounds), then $\exists t\in[T]$ with
$$over~t-1~rounds~we~have~S_i(t-1)=u_i,~and~play~A_t= \operatorname{argmax}_j \mathrm{UCB}_j(t-1, \delta)=i~at~t,$$
{so arm $i$ is played at least $u_i+1$ times at round $t$}. Using the definition of $G_i$,
		$$
		\begin{aligned}
			\mathrm{UCB}_i(t-1, \delta) & =\widehat{\mu}_i(t-1)+\sqrt{\frac{2 \log (1 / \delta)}{S_i(t-1)}}~~[\text{Definition of } \operatorname{UCB}_i(t-1, \delta)]\\
	[\text{Since } S_i(t-1)=u_i]~& =\frac{1}{u_i} \sum_{\tau \leq t, A_{\tau}=i} Y_{i}(\tau)+\sqrt{\frac{2 \log (1 / \delta)}{u_i}} \\
[\text{Definition of } G_i]~& <\mu_1 <\mathrm{UCB}_1(t-1, \delta).
		\end{aligned}
		$$
{Hence, $A_t=1\not=i$, which is a contradiction.} Therefore,
$$if~G_i~holds,~{E}S_i(T)\le u_i.$$
Then,
$${E}\left[S_i(T)\right]={E}\left[{I}\left\{G_i\right\} S_i(T)\right]+{E}\left[{I}\left\{G_i^c\right\} S_i(T)\right] \leq u_i+{P}\left(G_i^c\right)T,$$
where ${E}\left[{I}\left\{G_i^c\right\} S_i(T)\right]\le {P}\left(G_i^c\right) T$ is obvious since $S_i(T)\le T$. For~    	$$
    	G_i^c=\left\{\mu_1 \geq \min _{t \in[T]} \mathrm{UCB}_1(t, \delta)\right\} \cup\left\{\frac{1}{u_i} \sum_{\tau \leq t, A_{\tau}=i} Y_{i}(\tau)+\sqrt{\frac{2 \log (1 / \delta)}{u_i}} \geq \mu_1\right\},
    	$$
    	it holds that
    	$$
    	\begin{aligned}
    		&P\left(\mu_1 \geq \min _{t \in[T]} \mathrm{UCB}_1(t, \delta)\right)  \leq P\left(\bigcup_{s \in[T]}\left\{\mu_1 \geq \frac{1}{s} \sum_{\tau \leq t, A_{\tau}=1} Y_{i}(\tau)+\sqrt{\frac{2 \log (1 / \delta)}{s}}\right\}\right) \\
    		& \leq \sum_{s=1}^T P\left(\mu_1 \geq \frac{1}{s} \sum_{\tau \leq t, A_{\tau}=1} Y_{i}(\tau)+\sqrt{\frac{2 \log (1 / \delta)}{s}}\right) \leq T \delta,
    	\end{aligned}
    	$$
    	the last inequality is by \eqref{eq:UCB}.

Then, since $\mu_1=\mu_i+\Delta_i$, and~using sub-G inequality for the second term in $G_i^c$,
\[
    	\begin{aligned}\small
    		P\left(\sum_{\tau \leq t, A_{\tau}=i} \frac{Y_{i}(\tau)}{u_i} \right. &+ \left. \sqrt{\frac{2 \log (1 / \delta)}{u_i}} \geq \mu_1\right) 
     =P\left( \sum_{\tau \leq t, A_{\tau}=i}\frac{Y_{i}(\tau)}{u_i} -\mu_i \geq \Delta_i-\sqrt{\frac{2 \log (1 / \delta)}{u_i}}\right) \\
    [\text{By}~\eqref{equation1}]		& \leq P\left( \sum_{\tau \leq t, A_{\tau}=i} \frac{Y_{i}(\tau)}{u_i}-\mu_i \geq c \Delta_i\right) \leq \exp \left(-\frac{u_i c^2 \Delta_i^2}{2}\right),
    	\end{aligned}
\]
where $u_i$ is chosen large enough that \textit{{signal-to-noise condition}}
\begin{equation}
    		\label{equation1}
    		\Delta_i-\sqrt{\frac{2 \log (1 / \delta)}{u_i}} \geq c \Delta_i\Leftrightarrow (1-c)\Delta_i \geq \sqrt{\frac{2 \log (1 / \delta)}{u_i}}
    	\end{equation}
    	for some $c \in(0,1)$ to be chosen later. It remains to choose a proper $u_i \in[T]$. 
        
        A best choice is the smallest integer s.t. $u_i\ge \frac{2 \log (1 / \delta)}{(1-c)^2 \Delta_i^2}$ holds, which is
$u_i=\left\lceil\frac{2 \log (1 / \delta)}{(1-c)^2 \Delta_i^2}\right\rceil$.
Since
$P\left(G_i^c\right) \leq T \delta+\exp \left(-\frac{u_i c^2 \Delta_i^2}{2}\right)$. Let $\delta=1/T^2$ and $c=1/2$,  we have
\begin{equation}\label{equation2}
    		\begin{aligned}
    			{E}\left[S_i(T)\right]&\leq u_i+{P}\left(G_i^c\right)T \leq u_i+T\left(T \delta+\exp \left(-\frac{u_i c^2 \Delta_i^2}{2}\right)\right)\\
    	[\text{By}~u_i\ge \frac{2 \log (T^2)}{(1-c)^2 \Delta_i^2}]~		& \leq\left\lceil\frac{2 \log \left(T^2\right)}{(1-c)^2 \Delta_i^2}\right\rceil+1+T^{1-2 c^2 /(1-c)^2} \leq \frac{16 \log T}{\Delta_i^2}+3,
    		\end{aligned}
    	\end{equation}
where we use $\left\lceil x \right\rceil < 1+x$ in the last inequality. We obtain the problem-dependent bound
$$\operatorname{Reg}_T(\pi,v)=\sum_{i=1}^K \Delta_i {E}\left[S_i(T)\right]\le \sum_{i=1}^K \Delta_i[\frac{16 \log T}{\Delta_i^2}+3]=3\sum_{i=1}^K \Delta_i+\sum_{i: \Delta_i>0} \frac{16 \log T }{\Delta_i}.$$
    
For problem-dependent bound, let $\Delta>0$ be a threshold value (to be tuned) for $\Delta_k$.
    	$$
    	\begin{aligned}
    		\operatorname{Reg}_T(\pi,v) & =\sum_{i=1}^K\Delta_i {E}\left[S_i(T)\right]=\sum_{i: \Delta_i<\Delta} \Delta_i {E}\left[S_i(T)\right]+\sum_{i: \Delta_i \geq \Delta} \Delta_i {E}\left[S_i(T)\right] \\
  [\text{By~\eqref{equation2}~and}~{E}[S_i(T)] & \leq T \Delta+\sum_{i: \Delta_i \geq \Delta}\left(3 \Delta_i+\frac{16 \log T}{\Delta_i}\right) \leq T\Delta+\frac{16 K \log T}{\Delta}+3 \sum\nolimits_{i=1}^K \Delta_i \\
    		& \leq 8 \sqrt{T K\log T}+3 \sum\nolimits_{i=1}^K \Delta_i,
    	\end{aligned}
    	$$
    	where the first inequality is by $\sum_{i: \Delta_i<\Delta} S_i(T) \leq T$ and the last line is by
    $\Delta:=\sqrt{(16 K \log T) /T}$.
\end{proof}

\section{}
\label{app:proof_regret_bound2}
\begin{proof}[Proof of $O(K + \sqrt{KT\log T})$-regret bound for UCB by Lemma~\ref{lem:UCB}]
Define a good event:
$$G:=\left\{|\widehat{\mu}_k(t-1)-\mu_k| \leq \sqrt{\frac{2 \log (1 / \delta)}{S_k(t-1) \vee 1}}, \forall t \in[T], \forall k \in [K]\right\} .$$
By \eqref{eq:Sub-Gaussian} and the union bound of $TK$ events above, it implies
\begin{align}\label{eq:good0}
{P}\left(G^c\right) \leq 2 T K \delta~\text{with}~\mu_{k^*}-\mathrm{UCB}_{k^*}(t-1, \delta) \leq 0,~\forall~t \in[T].
\end{align}
where $\text{UCB}_{k^*}(t-1,\delta)=\widehat{\mu}_{k^*}(t-1)+\sqrt{2\log(1/\delta)/(S_{k^*}(t-1)\vee 1)}$, and~under the $G$,
\begin{align}\label{eq:good}\scriptsize
\mathrm{UCB}_{A_t}(t-1, \delta)-\mu_{A_t}&=\widehat{\mu}_{A_t}(t-1)-\mu_{A_t}+\sqrt{2\log(1/\delta)/(S_{k^*}(t-1)\vee 1)}\nonumber\\
& \le 2\sqrt{2\log(1/\delta)/(S_{k^*}(t-1)\vee 1)}.
\end{align}
The regret decomposition lemma for UCB algorithms shows

$$
\begin{aligned}
&\operatorname{Reg}_T(\pi,v)  \leq {E}\sum_{t=1}^T[\mathrm{UCB}_{A_t}(t-1, \delta)-\mu_{A_t}]\\
&={E}\sum_{t=1}^T\left[\frac{\sum_{\tau=1}^t [Y_{A_t}(\tau)-\mu_{A_t}]}{S_{A_t}(t-1)\vee 1} +\sqrt{\frac{2\log({\delta^{-1}})}{S_{A_t}(t-1)\vee 1} }\right]\\
 &\leq {E}\{1_{G^c}\sum\nolimits_{t=1}^T[\mathrm{UCB}_{A_t}(t-1, \delta)-\mu_{A_t}]\}+{E}\{1_{G}\sum\nolimits_{t=1}^T[\mathrm{UCB}_{A_t}(t-1, \delta)-\mu_{A_t}]\}\\
&\leq2|\mu_{k^*}|T {P}\left(G^c\right)+\sum_{t=1}^T {E}\sqrt{\frac{2\log({\delta^{-1}})}{S_{A_t}(t-1)\vee 1} }+{P}(G){E}\left[\sum_{t=1}^T [\mathrm{UCB}_{A_t}(t-1, \delta)-\mu_{A_t}] \mid G\right] \\
& \leq 4|\mu_{k^*}|T^2 K \delta+3 {E}\left[\sum_{t=1}^T \sqrt{\frac{2 \log (1 / \delta)}{S_{A_t}(t-1) \vee 1}}\right].
\end{aligned}
$$
where the last inequality is by \eqref{eq:good0} and \eqref{eq:good}, and~the second inequality is from
$E\{{\sum_{\tau=1}^T [Y_{A_t}(\tau)-\mu_{A_t}]}/({S_{A_t}(t-1)\vee 1})\}\le \max_tE[Y_{A_t}(\tau)-\mu_{A_t}]\le 2|\mu_{k^*}|.$
Estimating the last summation by the integral, we have
$$
\begin{aligned}
\operatorname{Reg}_T(\pi,v)&  \leq 4|\mu_{k^*}|T^2 K \delta+3\sum_{k=1}^K {E}\left[\sum_{t=1}^T \sqrt{\frac{2\log (1 / \delta)}{S_{k}(t-1) \vee 1}}I({A_t}=k)\right] \\
&\leq 4|\mu_{k^*}|T^2 K \delta+3 \sum_{k=1}^K E\int_0^{S_k(T)} \sqrt{\frac{2\log (1 / \delta)}{s}} \mathrm{~d} s\\
 &= 4|\mu_{k^*}|T^2 K \delta+3E \sum_{k=1}^K [ 2\sqrt{{2{S_k(T)}\log (1 / \delta)}}]\\
~[\text{Put}~\delta=1 / T^2]~&=4|\mu_{k^*}| K +6\sqrt{2\log (1 / \delta)}E \sum_{k=1}^K  \sqrt{{{S_k(T)}}} \\
[\text{Cauchy's inequality}]~& \leq 4|\mu_{k^*}| K +6\sqrt{2\log (1 / \delta)}E \sqrt{{\sum_{k=1}^K 1^2 \cdot\sum_{k=1}^K{S_k(T)}}} \\
[\sum\nolimits_{k=1}^K{S_k(T)}= T]~& \leq 4|\mu_{k^*}| K+12\sqrt{\log T}\sqrt{K T},
\end{aligned}
$$
where the second inequality is from the fact that $S_{k}(t)$ is a random and non-decreasing step function with the possible jump $1$.
\end{proof}
}
\begin{adjustwidth}{-\extralength}{0cm}

\reftitle{References}
\PublishersNote{}
\end{adjustwidth}

\end{document}